\documentclass{article}

\PassOptionsToPackage{numbers, compress}{natbib}

\usepackage[final]{neurips_2024}

\usepackage[utf8]{inputenc}
\usepackage[T1]{fontenc}
\usepackage{hyperref}
\usepackage{url}
\usepackage{booktabs}
\usepackage{amsmath,amssymb,amsthm}
\usepackage{bbm}
\usepackage{nicefrac}
\usepackage{microtype}
\usepackage{xcolor}
\usepackage{graphicx}
\usepackage{subcaption}
\usepackage{multirow}
\usepackage{siunitx}
\usepackage{placeins}
\usepackage{fancyhdr}

\fancypagestyle{firstpage}{%
  \fancyhf{}%
  \fancyfoot[C]{\footnotesize Preprint. Last updated \today.}%
}

\graphicspath{{figures/}}

\newcommand{\Nstar}{N^{\star}}
\newcommand{\Dstar}{D^{\star}}

\newcommand{\Kstar}{K^{\star}}
\newcommand{\Bcrit}{B_{\mathrm{crit}}}
\newcommand{\Lseq}{L_{\mathrm{seq}}}
\newcommand{\ind}{\mathbbm{1}}

\makeatletter
\newcommand{\appendixsection}[1]{%
  \begingroup
  \renewcommand{\addcontentsline}[3]{}%
  \section{#1}%
  \endgroup
}
\makeatother

\title{Scaling Laws for Behavioral Foundation Models over\\ User Event Sequences}

\author{%
  Rickard Br\"uel Gabrielsson\\
  Unbox AI\\
  \texttt{rickard@unboxai.com}
}

\begin{document}

\maketitle
\thispagestyle{firstpage}

\begin{abstract}
Foundation models are increasingly trained on sequences of user actions in
recommendation, payments, fraud, and commerce, but these models still lack the
kind of compute calibration that scaling laws provide for language models.  We
study a common two-part behavioral-model architecture: a feature-based event embedder
maps each multi-modal item to a vector, and a decoder-only transformer predicts
the next event from the resulting sequence.  Across roughly 600 runs on real
interaction data, spanning $10^{15}$--$10^{19}$ training FLOPs, we jointly vary
four deployment-relevant axes: the two-part parameter split,
critical batch size, model/data allocation, and the number of sampled negatives
used after freezing the embedder.  A small embedder
($s^{\star}\!\approx\!2\%$ of parameters) is compute-optimal at every budget we
test because embedder parameters are both more expensive per step and exposed
to far more repeated items than contextualizer parameters.  Compute-optimal
training is data-heavy relative to text at low compute, but its $D/N$ ratio
moves toward the Chinchilla heuristic as compute increases.  The sampled
training objective and deployed ranking metrics disagree in ways that themselves
scale: critical batch size, optimal negative count after freezing, and the agreement
between loss and ranking quality all shift with compute and with the chosen
evaluation metric.  For negative sampling, larger budgets increasingly prefer
more negatives; by $10^{19}$ FLOPs the active constraint is candidate-axis
memory rather than FLOPs.  In behavioral foundation models, the evaluation
metric is therefore part of the scaling law: changing it can change the
compute-optimal recipe.
\end{abstract}

\setcounter{tocdepth}{2}% Sections and subsections only (no \paragraph lines in the TOC).
\tableofcontents
\FloatBarrier

% =====================================================================
\section{Introduction}
\label{sec:intro}

Scaling laws turned large language model development from a sequence of ad hoc
training runs into a quantitative allocation problem: given a compute budget,
how many parameters should be trained on how many tokens, at what batch size,
and with which optimizer recipe?  The same question is now appearing outside
text.  Industrial systems increasingly train foundation models over sequences
of human actions: recommendations, purchases, payments, financial events,
workforce activity, and other behavioral traces
\citep{zhai2024hstu,zhang2024wukong,netflix2025fm,visa2025treasure,
stripe2025pfm,revolut2026pragma,jpmorgan2025tradefm,unboxai2026lbm}.  These
models are large enough that scaling-law mistakes are expensive, yet the field
has little Chinchilla-style guidance for them.

Behavioral foundation models differ from text LMs in a way that matters for
scaling.  A token in a language model is usually an opaque vocabulary id.  An
event in a behavioral model is feature-rich: a product or transaction can have
text, category metadata, visual features, price, timestamp, payment channel, and
other structured fields.  Catalogues also change over time.  Modern systems
therefore tend to use a two-part architecture: a feature-based \emph{event
embedder} maps raw item features into a dense representation, and a
\emph{contextualizer}, typically a decoder-only transformer, models sequences of
those event embeddings.

This architecture creates scaling questions that do not appear in ordinary text
LMs.  The embedder is suitably trained end-to-end~\citep{unboxai2025consumption} on the sequential
objective, but re-embedding every candidate item during a full softmax is
prohibitively expensive for million-item catalogues.  Prior work addresses this with a two-stage recipe: train the
embedder and contextualizer jointly with an in-batch sampled softmax, then
freeze and cache the embedder and continue training only the contextualizer with
a larger sampled candidate pool~\citep{unboxai2025consumption,unboxai2026lbm}.
The resulting system has at least four coupled knobs: how much capacity belongs
in the embedder, which batch size is data-efficient, how compute should be split
between model size and data, and how many negatives should be sampled after the
embedder is frozen.

We calibrate those knobs on a single behavioral-model stack and a single real
interaction corpus.  The study spans approximately 600 runs and
$C\!\in\![10^{15},10^{19}]$ FLOPs, with a shared evaluation pipeline across all
experiments.  The goal is not to propose a new architecture, but to answer a
more basic question: if a practitioner is already training this now-standard
event-embedder $\rightarrow$ transformer stack, what scaling laws should guide
the next run?

Our main findings are:
\begin{itemize}
  \item \textbf{The compute-optimal event embedder is small.}  Across four
  decades of compute, two-term iso-FLOP fits place the optimal embedder share in
  a narrow band around $s^{\star}\!\approx\!2\%$ of parameters.  The optimum is
  explained by two asymmetries: embedder parameters are touched many more times
  per item, and popular items are repeated hundreds to thousands of times while
  contextualizer windows rarely repeat.

  \item \textbf{Behavioral scaling is initially data-heavy but moves toward the
  Chinchilla heuristic.}  The compute-optimal $D/N$ ratio decreases from roughly
  $340$ at $10^{15}$ FLOPs to roughly $36$ at $10^{19}$ FLOPs, approaching the
  text-LM rule of thumb at larger budgets.  The fitted model-size exponent is
  $\Nstar\!\propto\!C^{0.617\pm0.025}$ under validation loss, with similar
  exponents under the headline ranking metrics.

  \item \textbf{The evaluation metric is part of the scaling law.}
  Cross-budget exponents are relatively stable across metrics, but the
  actionable recipe is not.  Critical batch size, optimal negative count after
  freezing, and the agreement between loss and ranking quality all depend on
  compute, evaluation regime, and target metric.  In particular, the
  sampled-softmax loss used during training is not always a reliable proxy for
  full-catalogue ranking quality.

  \item \textbf{Negative sampling shifts from a compute question to a memory
  question at scale.}  At smaller budgets, smooth fits place useful negative
  counts in the low hundreds of thousands and the optimum is metric-dependent.
  At $C\!=\!10^{19}$, every headline metric is still improving at the largest
  $K$ we train, so the active constraint becomes candidate-axis memory rather
  than available FLOPs.
\end{itemize}

Table~\ref{tab:study_map} summarizes the experimental map.  The remainder of
the paper follows the table: Section~\ref{sec:setup} defines the model,
training, compute accounting, and evaluation protocol; Sections~\ref{sec:phase1}--\ref{sec:phase4}
calibrate the four knobs; Section~\ref{sec:metrics} explains why metric choice
changes the recipe; and Section~\ref{sec:discussion} distills the practical
schedule and limitations.

\begin{table}[!t]
  \centering
  \footnotesize
  \setlength{\tabcolsep}{3pt}
  \caption{\textbf{Experimental map.}  Each axis is swept on the same
  two-part behavioral-model stack.  Labels retain the original phase numbers
  for cross-reference with the appendix, but the main text treats them as four
  scaling-law questions.}
  \label{tab:study_map}
  \begin{tabular}{p{0.18\linewidth}p{0.31\linewidth}p{0.22\linewidth}p{0.21\linewidth}}
    \toprule
    Axis & Main question & Primary sweep & Headline result \\
    \midrule
    Architecture & How large should the embedder be? & $s\!=\!0$--$50\%$ & $s^{\star}\!\approx\!2\%$ \\
    Batch size & Where is the data-efficiency knee? & $B\!=\!64$--$2048$ & $\Bcrit$ depends on metric \\
    Model/data allocation & How should $C$ split into $(N,D)$? & $10^{15}$--$10^{19}$ FLOPs & $D/N$ falls $344\!\to\!36$ \\
    Negative sampling & How many negatives after freezing? & $K\!=\!0$--$2$M & Metric-dependent; memory-bound at high $C$ \\
    Metric/eval regime & Do metrics and candidate pools rank cells consistently? & Cross-metric and local-vs-global correlations & High within-regime correlation; local loss is not global loss \\
    \bottomrule
  \end{tabular}
\end{table}
% =====================================================================
\section{Experimental Setup}
\label{sec:setup}

\paragraph{Model.}
All experiments use the same two-part next-event prediction architecture.  The
feature-based embedder consumes the raw fields of each catalogue item and
produces an event embedding at hidden size $h$.  The contextualizer is a
decoder-only transformer that consumes a sequence of $\Lseq\!=\!256$ event
embeddings and predicts the next event with a sampled softmax.  The parameter
count $N$ excludes vocabulary parameters and is decomposed into embedder
parameters $p_e$ and contextualizer parameters $p$; the embedder share is
$s\!=\!p_e/N$.

\paragraph{Two-stage training.}
Following the deployed recipe~\citep{unboxai2025consumption,unboxai2026lbm}, all
runs use two stages.  In \emph{Stage~1} the embedder and contextualizer are
trained jointly on the next-event objective with an \emph{in-batch} sampled
softmax: the positive is the true next item and the negatives are the other
targets in the same global batch, so the candidate pool is the
$\le\!B\Lseq$ unique items present in the batch.  In \emph{Stage~2} the trained
embedder is frozen and its event embeddings are cached, and only the
contextualizer continues training, now scoring each position against a larger
pool of $K$ uniformly sampled extra negatives on top of the in-batch
candidates.  Stage~1 sets architecture, batch size, and the $(N,D)$
allocation; Stage~2 isolates the negative-pool size $K$ at a fixed Stage~1
backbone.  This split is what makes re-embedding a million-item catalogue
tractable at serving time and defines the two evaluation regimes below.

\paragraph{Data.}
All sweeps train on an anonymized real-world retail interaction
corpus that combines offline and online consumer activity (product
searches, views, clicks and purchases), chunked into sequences of
$\Lseq\!=\!256$ events per training example.  Each event is multi-modal:
a free-text description, categorical fields (e.g. event
type, merchant, device, etc), numerical fields (e.g. price, timestamp, etc), and
optional visual features.  The catalogue contains on the order of $10^8$ unique
actions, and training consumes on the order of $10^9$ event tokens.  Item
popularity is strongly heavy-tailed: a small head of frequent actions is
observed many times within a single training run while much of the long tail is
observed only once or twice.  This asymmetry drives the
embedder/contextualizer compute tradeoff (\S\ref{sec:phase1}).

\paragraph{Training recipe.}
We train with AdamW, weight decay $0.1$, \texttt{bf16} mixed precision, and
fully sharded data parallelism.  The architecture, allocation, and sampling
experiments use cosine learning-rate decay with $5$--$10\%$ linear warm-up.  The
batch-size experiment uses a constant learning rate so that ``updates to target''
measures optimization efficiency rather than schedule shape.  Learning rate is
selected per cell from the training loss, not from validation metrics.

\paragraph{Compute accounting.}
We report training compute in standardized buckets and target
$C\!\approx\!6ND$ following the language-model scaling-law convention, where
$D\!=\!T B \Lseq$ is the number of event tokens consumed by $T$ optimizer steps
at global batch size $B$.  The experiment generator uses the finer Kaplan-style
per-step formula
\begin{equation}
  F_\text{step} \;=\; 6\,B\,\Lseq\,\bigl(t\,p_e + p + 3\,B\,\Lseq\,h\bigr),
  \label{eq:flops}
\end{equation}
where $t\!=\!24$ is the embedder context length in events.  The hidden size
$h$ is not a global constant: it is set per architectural cell and determines
both the event-embedding dimension used by the contextualizer and the dimension
at which in-batch contrastive scoring is performed.  In the embedder-share
sweep, changing the target share $s\!=\!p_e/N$ changes both $h$ and the
embedder parameter count $p_e$; the contextualizer parameter count $p$ then
follows from the target total $N$ at the fixed $D/N$ ratio.  In the depth sweep,
$h$ and $p_e$ are held fixed while contextualizer depth changes $p$, so the
embedder contributes the same $6B\Lseq t p_e$ compute tax across depth cells.
The term $3B\Lseq h$ is the in-batch contrastive scoring term after candidate
embeddings are all-gathered across data-parallel ranks.  We also use
$N_\text{eff}\!\equiv\!p+t p_e$ when interpreting the embedder/contextualizer
tradeoff.

\paragraph{Evaluation.}
The evaluation protocol intentionally separates the two regimes used by the
training system.  Before the embedder is frozen, checkpoints are evaluated
against the unique target items in each validation batch, matching the in-batch
negative distribution used during training.  After freezing the embedder,
Stage~2 checkpoints are evaluated against the full cached deployed product
catalogue, matching the deployed retrieval setting.  We report
cross-entropy, perplexity, recall@$k$, NDCG@$k$~\citep{jarvelin2002ndcg},
MRR@10~\citep{voorhees1999mrr}, coverage@$k$~\citep{herlocker2004eval}, and
predictive entropy; the standard ranking metrics follow~\citet{manning2008ir},
and Appendix~\ref{app:metricdefs} gives the exact formulas as we compute them.
Throughout, ``training loss'' means the sampled-softmax objective optimized by
the model, while ``validation loss'' means the corresponding held-out
cross-entropy under the evaluation regime being used.
% =====================================================================
\section{Scaling the Event Embedder}
\label{sec:phase1}

\subsection{Embedder Share}
\label{sec:phase1w}

The first question is architectural: at fixed compute, how much capacity should
belong to the feature-based event embedder rather than the transformer
contextualizer?  The answer is stable and surprisingly small.  Across four
compute budgets, the best share is about $2\%$ of parameters.  This is not just
an empirical accident: the embedder is more expensive per parameter and sees a
more repetitive effective data distribution than the contextualizer.  The scale
is also a useful point of comparison to multimodal foundation models, where a
comparatively small modality encoder feeds a larger language model: LLaVA,
Flamingo and BLIP-2 use vision encoders on the order of a few to tens of
percent of total parameters~\citep{liu2023llava,alayrac2022flamingo,li2023blip2}.
Those systems have not, to our knowledge, been calibrated with the same
encoder-share scaling-law sweep; checking whether similar share laws hold for
multimodal encoders is natural future work.

\textbf{Setup.}  We hold the data-to-parameter ratio $D/N\!\approx\!15$
(Chinchilla) fixed across the sweep, so the width study asks \emph{where
on the embedder/contextualizer split} the loss bottoms out given that we
are Chinchilla-matched.  This complements the model/data allocation sweep of \S\ref{sec:phase3},
which instead asks where along the iso-FLOP frontier to sit at each~$C$.  Each $(C, s)$ cell
solves the Kaplan formula~\eqref{eq:flops} for $(N, D)$ at that ratio
and sizes the joint embedding dim $h$ (which also sets $p_e$ via the
fixed embedder architecture) to hit the target share
$s\!=\!p_e/N$; the realized $N/D$ proxy stays essentially flat
across~$s$ at each budget (Appendix~\ref{app:phase1w}), so the sweep
slides along the fixed-ratio line rather than drifting off it.  We sweep embedder
share $s$ from $0\%$ to $50\%$ (finer spacing below $6\%$) at four compute
budgets and three learning rates per cell, select the best LR on
training loss without using the validation set for that choice, and fit
each per-budget share-loss curve
with the two-term iso-FLOP form
\begin{equation}
  \mathrm{sign}\cdot y(s) \;=\;
  \underbrace{a\,s^{\alpha}}_{\substack{\text{contextualizer}\\\text{starvation}}}
  \;+\;
  \underbrace{b\,s^{-\beta}}_{\substack{\text{embedder}\\\text{starvation}}}
  \;+\; E,
  \label{eq:share-fit}
\end{equation}
with closed-form optimum
$s^{\star}\!=\!(b\beta/(a\alpha))^{1/(\alpha+\beta)}$ and
$\mathrm{sign}\!=\!+1$ for smaller-better metrics.  The smallest swept
cell carries a nominal target share of $0\%$, but a minimum embedder
width (the text encoder is never removed) means it realizes
$s\!\approx\!0.5\%$; we therefore fit and plot it at its realized
$s_{\text{floor}}\!\approx\!0.5\%$ rather than at the nominal $0\%$.

\begin{figure}[!t]
  \centering
  \includegraphics[width=0.95\linewidth]{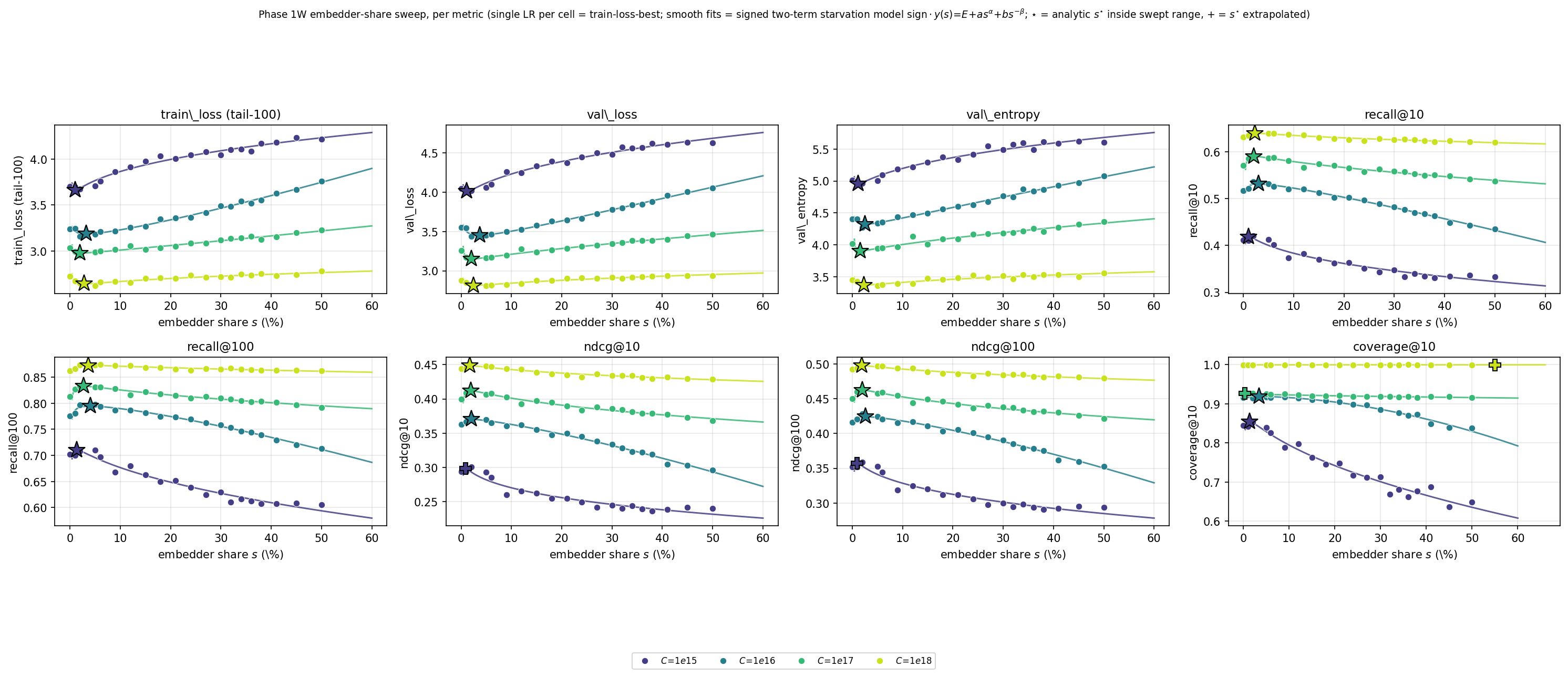}
  \caption{\textbf{Width sweep: every headline eval metric vs.\
    target embedder parameter share~$s$.}  One curve per compute budget;
    solid lines are the per-(metric, budget) two-term starvation fits of
    \eqref{eq:share-fit}.  $\star$ = analytic $s^{\star}$ inside the
    swept range $s\!\in\![1\%,50\%]$; $+$ = boundary case ($s^{\star}$
    extrapolated outside the sweep, almost always \texttt{coverage@10}
    at large~$C$).  The Kaplan-FLOP-share view of the same cells is
    Appendix~Figure~\ref{fig:phase1w_metrics_compute}; a
    \texttt{val\_loss}-only zoom on the small-share band is
    Figure~\ref{fig:phase1w_twostage}.}
  \label{fig:phase1w_metrics}
\end{figure}

Figure~\ref{fig:phase1w_metrics} summarizes every headline metric vs.\
target embedder share~$s$, with the per-(metric, budget) two-term fits
of~\eqref{eq:share-fit} overlaid and the analytic optimum $s^{\star}$
marked per panel; a \texttt{val\_loss}-only zoom on the small-share band
is Figure~\ref{fig:phase1w_twostage}.  The Kaplan FLOP-share cross-check
(same cells re-plotted vs.\ the realized embedder-side FLOP share~$f$
from~\eqref{eq:flops}) and the justification for parameterizing in~$s$
rather than~$f$ are in Appendix~\ref{app:phase1w}.  The reason is that
$s$ linearly trades off the two parameter pools, while $f$ is a nonlinear,
cell-dependent pushforward that breaks the two-term fit's conditioning.

\begin{figure}[!t]
  \centering
  \includegraphics[width=0.7\linewidth]{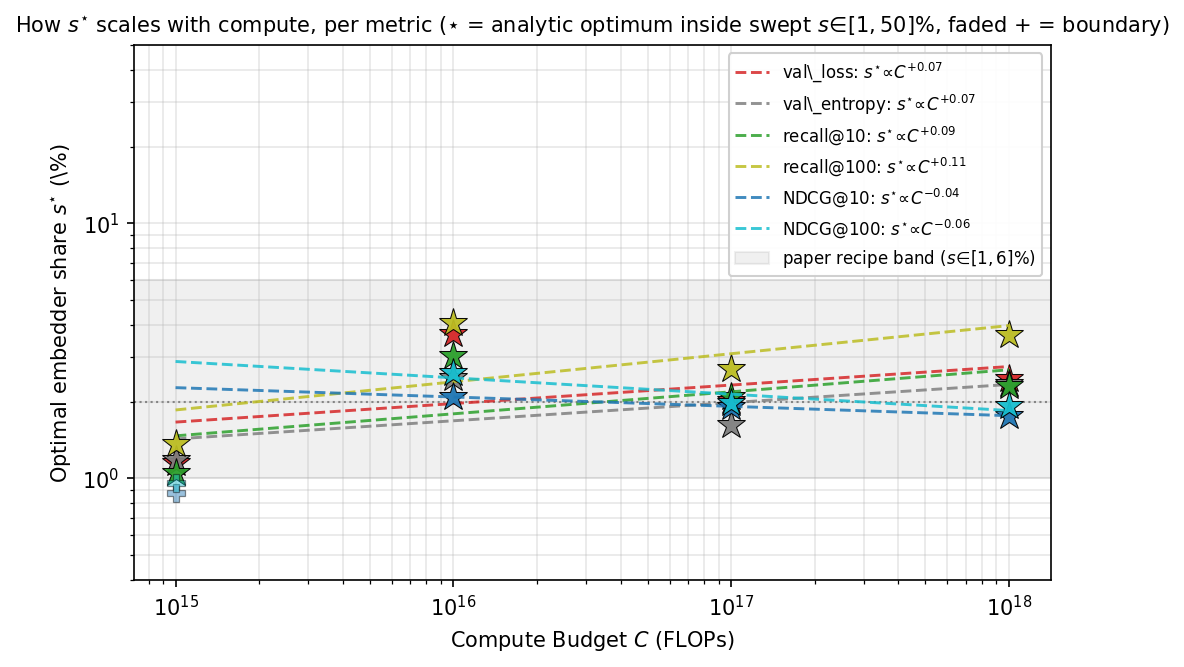}
  \caption{\textbf{$s^{\star}(C)$ summary, per metric.}  Visual companion
  to Table~\ref{tab:phase1w_sstar}: stars are the analytic $s^{\star}$
  from the per-(metric, budget) two-term fit when it lands inside the
  swept range $s\!\in\![1\%,50\%]$; faded $+$ markers are boundary-
  extrapolated $s^{\star}$ and are excluded from the dashed power-law
  fits.  Every loss / ranking metric fits
  $s^{\star}\!\propto\!C^{\rho}$ with $|\rho|\!\leq\!0.11$, effectively
  flat over four decades of compute.  \texttt{recall@100} drifts upward
  most strongly ($\rho\!=\!+0.11$), \texttt{NDCG@10} downward most
  strongly ($\rho\!=\!-0.10$); the gray band is the practitioner-recipe
  $s\!\in\![1\%,6\%]$ that contains every interior $s^{\star}$ in
  Table~\ref{tab:phase1w_sstar}.}
  \label{fig:phase1w_sstar_vs_C}
\end{figure}

\textbf{Findings.}  The share--loss relationship from $0\%$ to $50\%$ is
clean: loss is essentially flat in the small-share band $s\!\in\![2,6]\%$
($\Delta\!\leq\!0.06$ nat at every budget); monotonically increasing above
$6\%$, where going from $6\%$ to $27\%$ costs $\sim\!0.40$ nat at $C\!=\!10^{15}$,
falling to $\sim\!0.09$ nat at $C\!=\!10^{18}$; and shallow below $2\%$.
The two-term fit
$L(s)\!=\!E\!+\!a\,s^{\alpha}\!+\!b\,s^{-\beta}$ collapses every
budget into a single closed-form optimum
$s^{\star}\!\in\![1.1\%,\,3.7\%]$:
\emph{the optimal embedder share is constant at
$s^{\star}\!\approx\!2$--$3\%$ across all four budgets} (val\_loss
fits $s^{\star}\!\propto\!C^{+0.07}$, slope indistinguishable from
zero given a $0.5\%$ swept-share resolution).  Repeating the
analysis for every other headline eval metric agrees within
$1\sigma$: $s^{\star}\!\propto\!C^{\pm 0.10}$ across val\_entropy,
\texttt{recall@10}, \texttt{NDCG@10}, \texttt{NDCG@100}, all clustering
inside the $[1\%,\,6\%]$ band (Appendix~\ref{app:phase1w},
Figure~\ref{fig:phase1w_sstar_vs_C}); \texttt{recall@100} drifts
upward most, with $s^{\star}\!\in\![1.4\%,\,4.1\%]$ and an
$s^{\star}\!\propto\!C^{+0.11}$ trend, since a $100$-deep
recommendation list tolerates more capacity in the catalogue
representation.

\begin{table}[!t]
  \centering
  \small
  \caption{\textbf{Analytic $s^{\star}$ from the two-term starvation
  fit (\ref{eq:share-fit}), per budget and per metric.}  All cells
  fall inside the swept range $s\!\in\![1\%,50\%]$ (bold) except for
  three boundary cases (italic ``boundary''), where the fit's
  closed-form optimum falls outside the sweep because the curve is flat
  enough that the analytic minimum is not bracketed.  $\beta$ is the
  embedder-starvation exponent (penalty for $s\!\to\!0$); $\alpha$ is
  the contextualizer-starvation exponent (penalty for over-allocation
  to the embedder).  Both exponents are reported only for
  \texttt{val\_loss}; full per-metric exponents are in
  Appendix~\ref{app:phase1w}.  Figure~\ref{fig:phase1w_sstar_vs_C}
  is the visual summary of the $s^{\star}$ column across budgets and
  metrics; the Kaplan-FLOP-share cross-check is
  Appendix~Figure~\ref{fig:phase1w_metrics_compute}.}
  \label{tab:phase1w_sstar}
  \begin{tabular}{lcccccc}
    \toprule
                & \multicolumn{4}{c}{Analytic $s^{\star}$ (\%) (discrete-grid argmax in parentheses)}
                & \multicolumn{2}{c}{val\_loss exponents} \\
    \cmidrule(lr){2-5} \cmidrule(lr){6-7}
    Budget & val\_loss             & recall@10           & NDCG@10            & coverage@10
           & $\alpha$              & $\beta$            \\
    \midrule
    $10^{15}$ & \textbf{$1.1$} ($2$)   & \textbf{$1.1$} ($2$)   & \emph{bound.} ($2$)   & \textbf{$1.3$} ($0$)   & $0.40$ & $1.03$ \\
    $10^{16}$ & \textbf{$3.7$} ($2$)   & \textbf{$3.0$} ($2$)   & \textbf{$2.8$} ($2$)   & \textbf{$3.5$} ($0$)   & $0.94$ & $0.26$ \\
    $10^{17}$ & \textbf{$2.0$} ($2$)   & \textbf{$2.1$} ($2$)   & \textbf{$1.9$} ($2$)   & \emph{bound.} ($0$)   & $0.75$ & $1.65$ \\
    $10^{18}$ & \textbf{$2.4$} ($5$)   & \textbf{$2.3$} ($2$)   & \textbf{$1.8$} ($2$)   & \emph{bound.} ($0$)   & $0.48$ & $1.06$ \\
    \bottomrule
  \end{tabular}
\end{table}

\textbf{Why the embedder is so small.}  Two asymmetries between the
embedder and the contextualizer push the compute-optimal split into
the small-$s$ band.  \emph{(i)
Compute asymmetry}: the contextualizer runs once per $B\!\times\!L$ batch,
while the text encoder inside the embedder runs on $B\,L$ short sequences
per step.  Each embedder parameter is therefore touched $\sim\!t$ times
more often per item than each contextualizer parameter, which is
exactly the $N_\text{eff}\!=\!p\!+\!t\,p_e$ correction of \S\ref{sec:setup}.
\emph{(ii) Effective-epoch asymmetry}: the contextualizer effectively never
sees the same $L$-event window twice in our sub-one-epoch schedule, while
the embedder sees every popular item hundreds to thousands of times.  The
embedder is thus much more exposed to memorization than the contextualizer
under the same global regularization, and the $s^{\star}\!\approx\!2\%$
recommendation should be read as the answer for the current (essentially
unregularized) recipe.  Adding embedder-specific regularization may shift
$s^{\star}$ upward; we treat this as future work.

\subsection{Depth as a Secondary Knob}
\label{sec:phase1d}

The depth sweep is a complementary check on the width result: once
embedder share is already small, how should the remaining capacity
split between a shallow text encoder and a deeper contextualizer at
fixed compute?  We
trade $L_{\text{text}}$ against $L_{\text{ctx}}$ on an iso-FLOP
depth-sum constraint ($L_{\text{ctx}}\!+\!L_{\text{text}}$ fixed per
budget) and pick the best learning rate per $(L_{\text{ctx}},L_{\text{text}})$
cell on training loss, matching the width-sweep protocol.  Per-cell
grids and per-metric curves are in Appendix~\ref{app:phase1d}.

\textbf{Findings.}  We extract per-budget optima as the analytic
minimum of an iso-FLOP parabola in $\log s$, following the
model/data allocation sweep's
treatment of $(N,D)$ allocation (\S\ref{sec:phase3}).  The parabolic
fit recovers $s^{\star}\!=\!1.82\%$ at $C\!=\!10^{16}$ and
$s^{\star}\!=\!1.80\%$ at $C\!=\!10^{17}$, both within $0.2\%$ of
the embedder-share recommendation $s^{\star}\!\approx\!2\%$.
At $C\!=\!10^{18}$ the parabolic $s^{\star}\!=\!0.80\%$, but the
grid saturates at $L_{\text{ctx}}\!=\!32$ so the true minimum may
sit slightly deeper.  At $C\!=\!10^{15}$ the val\_loss landscape
across $L_{\text{ctx}}\!\in\![1,12]$ is too flat
($\leq\!0.21$~nats end-to-end) for the parabola to resolve an
interior minimum (Appendix Table~\ref{tab:phase1d_parabola},
Figure~\ref{fig:phase1d_parabolic_fits}); the underlying depth-sum
sweep covers $L_{\text{ctx}}^{\star}\!\in\!\{12,6,4,32\}$ across
the four budgets (Appendix Table~\ref{tab:phase1d_summary}), but
the small-$s$ valley is flat enough that picking $s\!=\!2\%$
instead of the per-budget optimum costs $\leq\!0.01$~nats under
the parabolic projection at
$C\!\in\!\{10^{16},10^{17},10^{18}\}$, and ranking metrics at the
small-$s$ cells agree with \texttt{val\_loss} (Appendix
Figure~\ref{fig:phase1d_eval_metrics_two_views}).  We therefore
treat depth as a second-order knob: any configuration in the
small-$s$ valley is acceptable once $s\!\approx\!2\%$ is set.  We
keep $s^{\star}\!\approx\!2\%$ from the width sweep as the
primary architectural recommendation.  At $C\!=\!10^{18}$ deeper
extensions off the iso-FLOP diagonal do not beat the depth-sum-$34$
winner on held-out metrics (Appendix Table~\ref{tab:phase1d_eval}).

% =====================================================================
\section{Critical Batch Size Across Metrics}
\label{sec:phase2}

Batch size is usually treated as an optimization detail, but in a scaling-law
recipe it determines how much data efficiency is traded for hardware
throughput.  We therefore ask how large the global batch can grow before extra
examples stop buying proportional progress.  At the fixed architecture selected
above, the answer depends on the metric: loss-like metrics and recall have a
critical batch near $\sim\!570$, while top-weighted ranking metrics saturate at
roughly one third of that value.

\textbf{Setup.}  We train at $B\!\in\!\{64,128,256,512,1024,2048\}$ with
square-root LR scaling and a constant LR schedule (so updates-to-target
reflects optimization efficiency, not schedule shape), with periodic
held-out evaluation during training. We then read off $S_m(B)$, the smallest update at which the EWMA-smoothed metric
first crosses a per-metric iso-target $T_m$, and fit the
Kaplan/McCandlish~\citep{mccandlish2018} critical-batch model per metric:
\begin{equation}
  S_m(B) \;=\; S^{\min}_m\!\left(1 + \frac{\Bcrit^{(m)}}{B}\right).
  \label{eq:bcrit}
\end{equation}
Smoothing, iso-target derivation, and the per-metric trajectories that
$S_m(B)$ is read off of are in Appendix~\ref{app:phase2}.

\begin{table}[!t]
  \centering
  \small
  \caption{\textbf{Per-metric critical batch size.}  Updates-to-target
  $S_m(B)$ on the EWMA-smoothed trajectory at the per-metric iso-target
  $T_m$, and the Kaplan fit \eqref{eq:bcrit}.  Validation loss and
  \texttt{recall@10} share a $\Bcrit$ near $\sim\!550$; position-weighted
  ranking metrics sit substantially lower ($\sim\!200$--$275$).
  $S_m(2048)\!\ge\!S_m(1024)$ for every metric except \texttt{val\_entropy}.}
  \label{tab:phase2}
  \begin{tabular}{lrrrrrrrrr}
    \toprule
    Metric & $T_m$ & $S(64)$ & $S(128)$ & $S(256)$ & $S(512)$
    & $S(1024)$ & $S(2048)$ & $\Bcrit$ & $R^2$ \\
    \midrule
    val\_loss     & $3.86$  & 7700 & 3250 & 1950 & 1450 & 1300 & 1550 & $574$  & $0.96$ \\
    recall@10     & $0.486$ & 7350 & 3550 & 2200 & 1450 & 1250 & 1250 & $544$  & $0.99$ \\
    NDCG@10       & $0.314$ & 6550 & 3700 & 2500 & 1750 & 1500 & 1650 & $274$  & $0.99$ \\
    MRR@10        & $0.257$ & 5950 & 3650 & 2500 & 1850 & 1600 & 1850 & $201$  & $0.99$ \\
    val\_entropy  & $5.39$  & 7650 & 2100 &  850 &  600 &  350 &  300 & --$^*$ & $0.92$ \\
    \bottomrule
  \end{tabular}
  \\[2pt]
  {\footnotesize $^*$\texttt{val\_entropy} decreases monotonically in $B$
  ($S\!=\!7650$ at $B\!=\!64$ to $300$ at $B\!=\!2048$) without a Kaplan
  plateau; $\Bcrit$ is poorly constrained and we omit a point estimate
  (likely above $2048$).}
\end{table}

\begin{figure}[!t]
  \centering
  \includegraphics[width=0.88\linewidth]{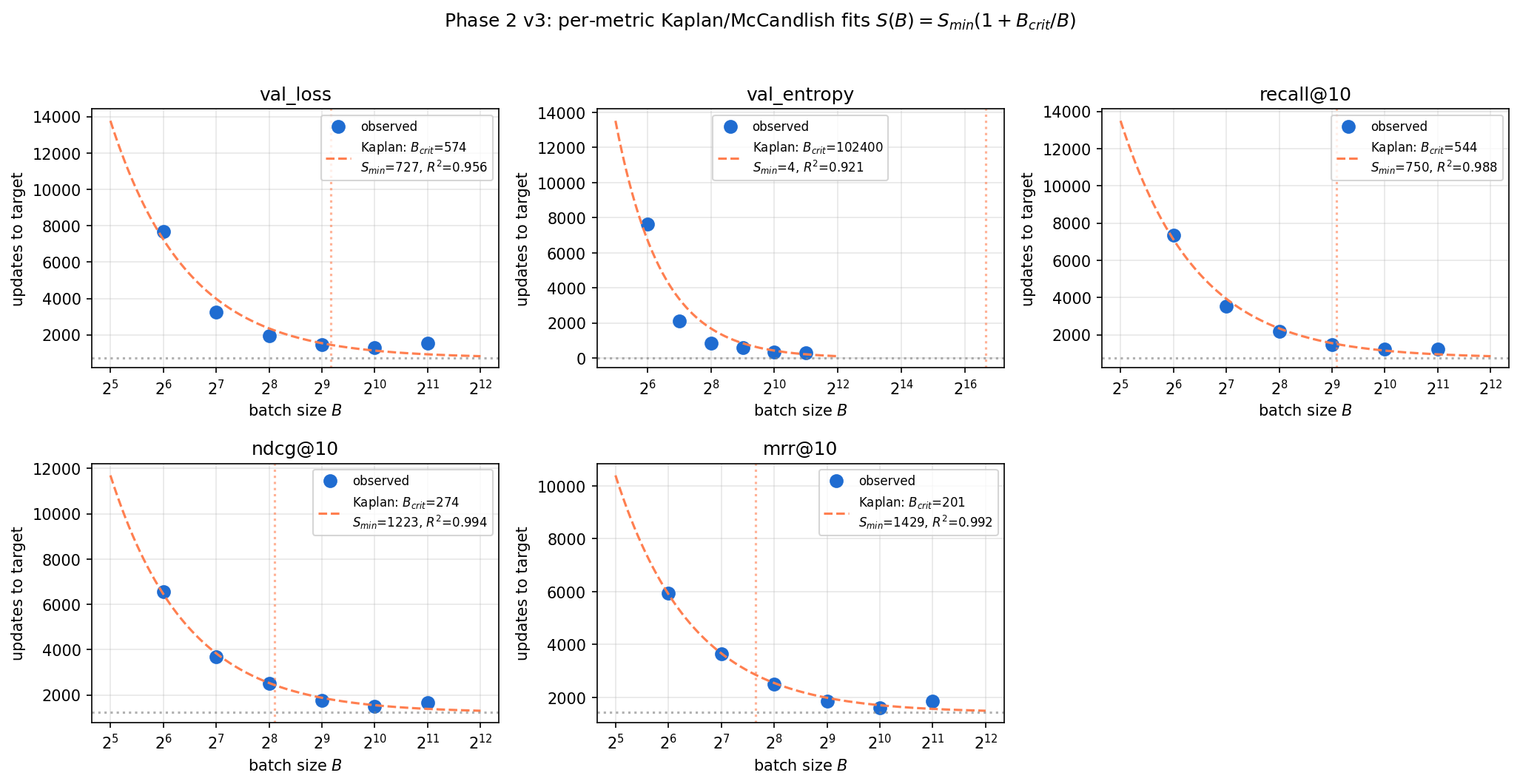}
  \caption{\textbf{Per-metric Kaplan fits} of \eqref{eq:bcrit}; dotted
  lines mark $\Bcrit^{(m)}$ and $S^{\min}_m$ per panel.
  \texttt{val\_loss} and \texttt{recall@10} coincide
  ($\Bcrit\!\approx\!574$ and $544$); \texttt{NDCG@10} and
  \texttt{MRR@10} sit notably lower; \texttt{val\_entropy} has no plateau
  within our $B$ range (Tab.~\ref{tab:phase2} footnote).
  EWMA-smoothed trajectories and iso-target crossings are in
  Appendix~\ref{app:phase2}.}
  \label{fig:phase2}
\end{figure}

\textbf{Findings.}  \emph{(i)~Loss and recall share a knee.}
\texttt{val\_loss} and \texttt{recall@10} have essentially the same
$\Bcrit\!\approx\!570$; the batch-size knee for cross-entropy and for the
dominant retrieval metric coincide.  \emph{(ii)~Position-weighted ranking
saturates earlier.} \texttt{NDCG@10} and \texttt{MRR@10} sit at
$\Bcrit\!\approx\!200$--$275$, and \texttt{val\_entropy} has no plateau
within the swept range.  This strict ordering matches a simple
sensitivity argument: if a downstream metric $M\!=\!f(L)$ has small
$|f'(L)|$ in the operating band, the gradient-noise reduction that larger
$B$ buys translates into a proportionally smaller improvement in $M$ and
the Kaplan curvature regime is reached at a smaller $B$ (top-weighted
ranking saturates quickly once the head is approximately correct;
entropy is at the high-sensitivity extreme).  \emph{(iii)~$B\!=\!2048$
sits just past the val-loss knee.}  $S_m(2048)\!\geq\!S_m(1024)$ for
every non-entropy metric, the predicted data-efficiency penalty.  This
is the largest single-node batch we measured without GPU-efficiency
loss, so we recommend it for throughput-limited production training
(\S\ref{sec:discussion}); the smaller batches used in the model/data
allocation sweep are picked
for iso-FLOP feasibility on the allocation grid (\S\ref{sec:phase3}).

% =====================================================================
\section{Model/Data Allocation Across Compute Budgets}
\label{sec:phase3}

We next ask the Chinchilla question for behavioral models: given a fixed compute
budget, should we train a larger model on fewer event tokens, or a smaller model
on more data?  The optimum is more data-heavy than the text-LM rule at small
budgets, but it moves steadily toward the Chinchilla heuristic as compute grows.
Across metrics, the fitted model-size exponents are close, even when the exact
best cell at a particular budget changes.

\textbf{Setup.}  We train $45$ contextualizers on the primary
fixed-architecture grid
$(h, L_{\text{ctx}})\!\in\!\{(128,4),(192,4),(256,6),(320,6),(384,6),(512,8)\}$ plus
six additional larger anchors at $C\!=\!10^{19}$
($h\!\in\!\{768,\dots,1408\}$, $L_{\text{ctx}}\!\in\![12,20]$), all with cosine
learning-rate decay and every
$(h, L_{\text{ctx}}, \mathrm{LR})$ cell trained from scratch.  Every cell uses embedder
share $s\!=\!2\%$ (\S\ref{sec:phase1w}).  The global training batch is
budget-scaled ($B\!\in\!\{64,128,256,512,512\}$ at
$C\!\in\!\{10^{15},10^{16},10^{17},10^{18},10^{19}\}$ so every iso-FLOP cell runs with feasible GPU throughput; the
larger-$B$ case for throughput-oriented training is in \S\ref{sec:phase2}.
We take \texttt{val\_loss} on the held-out set (batch-local pool, fixed
\texttt{eval\_batch} per budget) as the primary objective, since it
aligns with the headline ranking metrics under the same protocol and is
less optimizer-noisy than the tail-$100$ training surrogate (the
train-surrogate parallel allocation table is Appendix~\ref{app:phase3_train}).
Table~\ref{tab:phase3_alloc} lists, per budget, the \texttt{val\_loss}-best
grid cell alongside the Hoffmann Approach~2 parabolic optimum in $\log N$
and the closest swept $(h/L_{\text{ctx}})$ anchor.

\begin{table}[!t]
  \centering
  \small
  \setlength{\tabcolsep}{4pt}
  \caption{\textbf{Val-loss-optimal allocation per budget.}  \emph{Grid
  winner:} $\Nstar$, $\Dstar$, and $D/N$ at the run minimizing
  \texttt{val\_loss} on the iso-FLOP sweep, with observed val\_loss,
  R@10, and NDCG@10.  \emph{Parabolic (Approach~2):} analytic minimum of
  the quadratic-in-$\log N$ fit per budget
  (Fig.~\ref{fig:phase3_isoflop_grid}, \texttt{val\_loss} panel);
  $\Dstar\!=\!C/(6\Nstar)$.  The negative-sampling sweep sizes
  contextualizers from
  the grid-winner $\Nstar$ column.  Train-surrogate minima differ slightly
  (Appendix~\ref{app:phase3_train}).}
  \label{tab:phase3_alloc}
  \renewcommand{\arraystretch}{1.12}
  \begin{tabular*}{\linewidth}{@{\extracolsep{\fill}}l
      r r r r r r
      r r r@{}}
    \toprule
    & \multicolumn{2}{c}{\shortstack[c]{Grid\\winner}}
    & \multicolumn{4}{c}{\shortstack[c]{Observed at\\grid winner}}
    & \multicolumn{3}{c}{\shortstack[c]{Parabolic\\(Approach~2)}} \\
    \cmidrule(lr){2-3} \cmidrule(lr){4-7} \cmidrule(lr){8-10}
    Budget & $\Nstar$ & $\Dstar$ & val\_loss & R@10 & NDCG@10 & $D/N$
           & $\Nstar$ & $\Dstar$ & $D/N$ \\
    \midrule
    $10^{15}$ & 532\,k  & 84\,M   & 3.914 & 0.453 & 0.302 & 157
              & 695\,k  & 240\,M  & 345 \\
    $10^{16}$ & 1.95\,M & 251\,M  & 3.376 & 0.544 & 0.380 & 128
              & 2.51\,M & 664\,M  & 265 \\
    $10^{17}$ & 10.9\,M & 537\,M  & 3.108 & 0.594 & 0.414 & 49
              & 8.67\,M & 1.92\,B & 222 \\
    $10^{18}$ & 19.4\,M & 2.63\,B & 2.832 & 0.649 & 0.455 & 135
              & 38.9\,M & 4.28\,B & 110 \\
    $10^{19}$ & 251\,M  & 3.65\,B & 2.641 & 0.673 & 0.476 & 15
              & 216\,M  & 7.73\,B & 36 \\
    \bottomrule
  \end{tabular*}
  \renewcommand{\arraystretch}{1.0}
\end{table}

\begin{figure}[!t]
  \centering
  \includegraphics[width=0.99\linewidth]{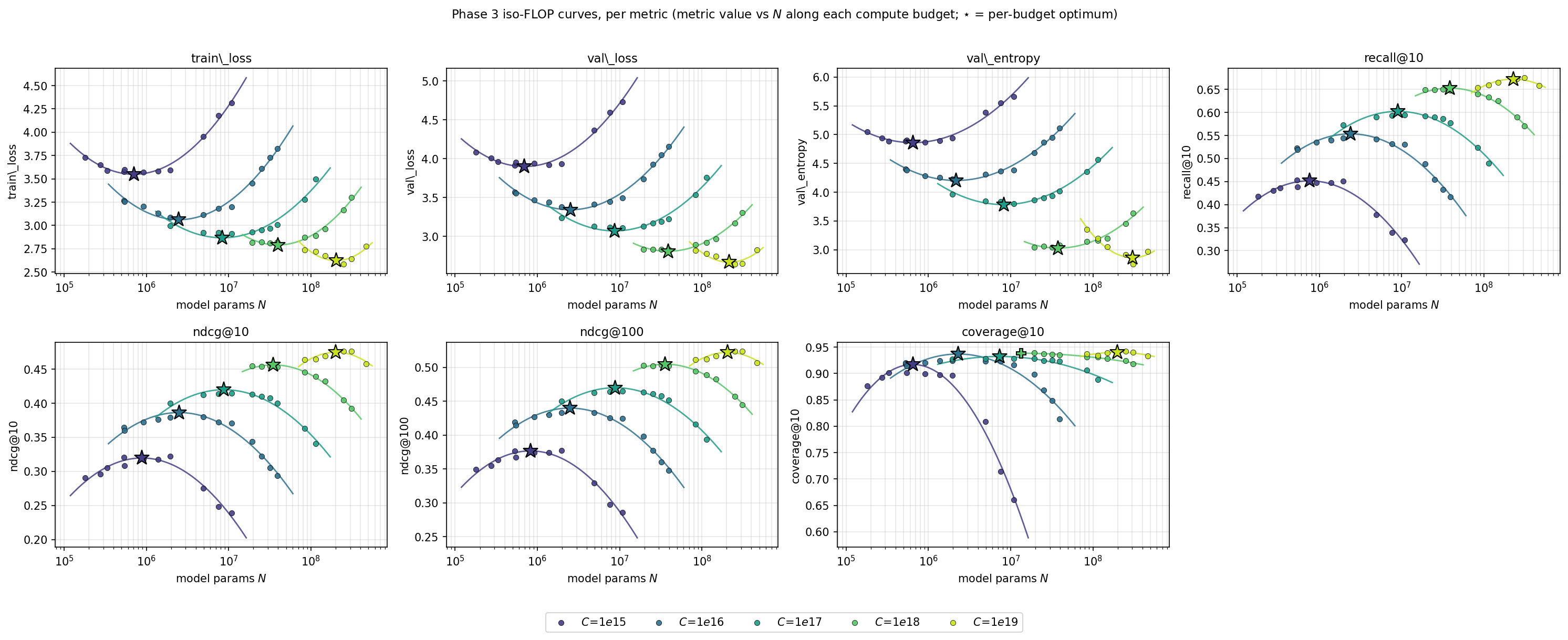}
  \caption{\textbf{Iso-FLOP curves per metric, with parabolic fits.}  Each
  panel scatters one headline metric vs.\ $N$ along each compute budget
  and overlays the quadratic-in-$\log N$ fit per budget; $\star$ sits at
  the parabola's analytic optimum (minimum for losses/entropy, maximum
  for ranking metrics).  The plus-marker variant is used at
  $(\textsc{coverage@10},C\!=\!10^{18})$ where the coverage curve is
  essentially flat ($a\!\approx\!0$) and the parabola maximum falls just
  outside the swept $N$ range.  These analytic optima feed the
  $(h/L_{\text{ctx}})$ closest-cell column of Tab.~\ref{tab:exponents} and the
  per-metric power-law fits in Fig.~\ref{fig:phase3_frontiers}.}
  \label{fig:phase3_isoflop_grid}
\end{figure}

\textbf{Power-law fits (val optima, parabolic).}  Fitting
$\Nstar(C)\!=\!aC^{b}$ and $\Dstar(C)\!=\!a'C^{b'}$ in log-space on the
five per-budget parabolic minima (Fig.~\ref{fig:phase3_isoflop_grid},
\texttt{val\_loss} panel; Hoffmann Approach~2) gives
\begin{align}
  \Nstar(C) &\;=\; 3.35\!\times\!10^{-4}\, C^{0.617\pm 0.025}, \label{eq:nstar} \\
  \Dstar(C) &\;=\; 4.97\!\times\!10^{2}\, C^{0.383\pm 0.025}, \label{eq:dstar}
\end{align}
with $b\!+\!b'\!=\!1$ exactly by Approach~2 construction.  The allocation
is moderately parameter-heavy ($b_N\!>\!b_D$); the train-surrogate
parallel ($b_N\!=\!0.612\!\pm\!0.024$) lands within $0.005$ of the
val-surrogate exponent (Appendix~\ref{app:phase3_train}).

\textbf{Data-heavy relative to text LMs, narrowing toward Chinchilla.}
Val-optimal points give $\Dstar/\Nstar$ that decreases monotonically
in $C$: $344\!\to\!265\!\to\!222\!\to\!110\!\to\!36$ across
$C\!=\!10^{15}\!\to\!10^{19}$.  Over the lower four budgets we sit
nearly an order of magnitude above the text-LM Chinchilla heuristic
$D/N\!\approx\!20$; the $C\!=\!10^{19}$ anchor brings the optimum to
within $\sim\!2\!\times$ of Chinchilla and extrapolates the trajectory
toward the text-LM heuristic at production scale.

\begin{figure}[!t]
  \centering
  \includegraphics[width=0.99\linewidth]{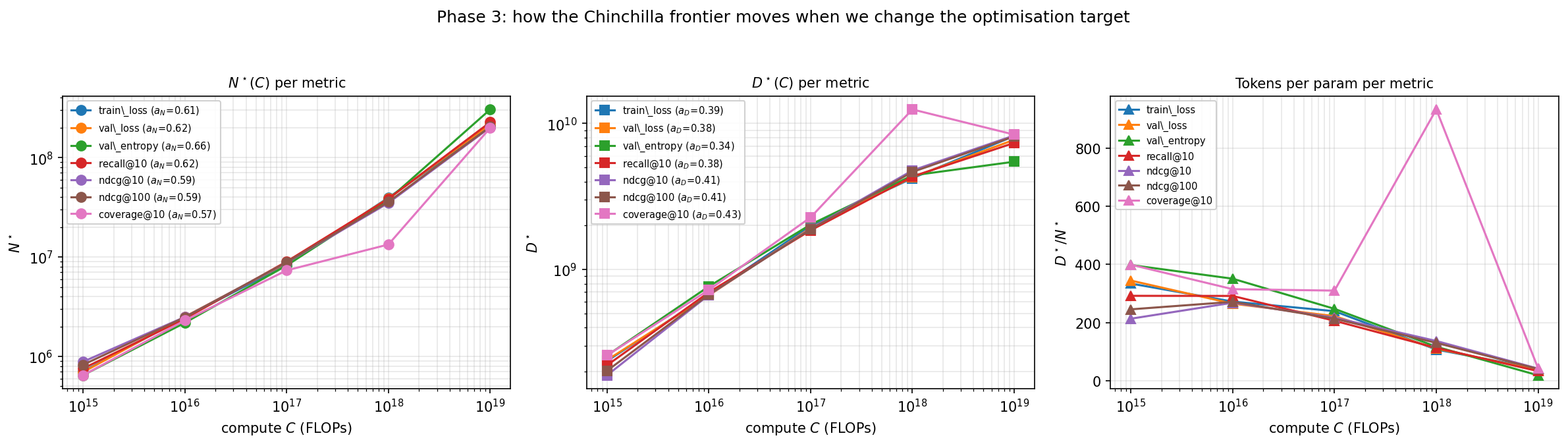}
  \caption{\textbf{Chinchilla frontier per metric.}  Left: $\Nstar(C)$;
  middle: $\Dstar(C)$; right: $\Dstar/\Nstar$.  Switching the target from
  \texttt{val\_loss}/\texttt{recall@10} to \texttt{NDCG@10} pushes the
  optimum toward smaller models trained on more tokens at every budget.}
  \label{fig:phase3_frontiers}
\end{figure}

\textbf{Per-metric allocation laws are metric-robust.}
Refitting $\Nstar\!\propto\!C^{a_N}$ on each metric's analytic
parabola optimum (Table~\ref{tab:exponents}) lands all five headline
metrics in the tight band $a_N\!\in\![0.57,\,0.66]$
($0.617\!\pm\!0.025$ \texttt{val\_loss}, $0.616\!\pm\!0.029$
\texttt{recall@10}, $0.586\!\pm\!0.033$ \texttt{NDCG@10},
$0.574\!\pm\!0.079$ \texttt{coverage@10}); the
\texttt{val\_loss}-vs-\texttt{NDCG@10} separation
$\Delta a_N\!\approx\!0.03$ is inside its own slope uncertainty.  The
per-budget winners do still disagree across metrics on which $(h,L)$
cell to ship: the scaling law is shared, but the recipe is not.
\S\ref{sec:metrics} consolidates this.
Figure~\ref{fig:phase3_isoflop_grid} plots the iso-FLOP curves and
parabolic fits per metric; Figure~\ref{fig:phase3_frontiers} shows the
implied frontiers in $(\Nstar,\Dstar,D/N)$-space.

\begin{table}[!t]
  \centering
  \small
  \caption{\textbf{Hoffmann-style exponents under each eval metric.}
  Per-budget optima are taken as the analytic minimum (or maximum, for
  ranking metrics) of the iso-FLOP parabola in $\log N$
  (Fig.~\ref{fig:phase3_isoflop_grid}) on the primary grid plus six
  $C\!=\!10^{19}$ anchors.  Slope
  errors are the 1-sigma OLS uncertainties of the five-budget log-log
  power-law fit; $a_N\!+\!a_D$ is exact at $1$ by construction since
  $D^{\star}\!=\!C/(6\Nstar)$.  Cell columns list the discrete $(h/L_{\text{ctx}})$
  closest to each parabola optimum.}
  \label{tab:exponents}
  \begin{tabular}{lcccl}
    \toprule
    Eval metric & $a_N$ & $a_D$ & $a_N\!+\!a_D$ & Best cells $(h/L_{\text{ctx}})$ at $C\!\in\!\{10^{15},\ldots,10^{19}\}$ \\
    \midrule
    \texttt{val\_loss}    & $0.617\!\pm\!0.025$ & $0.383\!\pm\!0.025$ & $1.000$ & $96/4,\ 192/4,\ 320/6,\ 640/8,\ 1152/16$ \\
    \texttt{val\_ppl}     & $0.617\!\pm\!0.025$ & $0.383\!\pm\!0.025$ & $1.000$ & $96/4,\ 192/4,\ 320/6,\ 640/8,\ 1152/16$ \\
    \texttt{recall@10}    & $0.616\!\pm\!0.029$ & $0.384\!\pm\!0.029$ & $1.000$ & $128/4,\ 192/4,\ 320/6,\ 640/8,\ 1152/16$ \\
    \texttt{NDCG@10}      & $0.586\!\pm\!0.033$ & $0.414\!\pm\!0.033$ & $1.000$ & $128/4,\ 192/4,\ 320/6,\ 576/8,\ 1152/16$ \\
    \texttt{coverage@10}  & $0.574\!\pm\!0.079$ & $0.426\!\pm\!0.079$ & $1.000$ & $96/4,\ 192/4,\ 320/6,\ 448/8,\ 1152/16$ \\
    \bottomrule
  \end{tabular}
\end{table}

% =====================================================================
\section{Scaling the Negative Candidate Pool}
\label{sec:phase4}

After the embedder is frozen, the cached catalogue makes it possible to train
against many more sampled negatives.  This creates a new scaling knob: how large
should the extra negative pool $K$ be at fixed compute?  The useful range is
broad but not arbitrary.  Smooth iso-FLOP fits place most metric-specific optima
in the low $10^5$--$10^6$ range; at the largest budget, the limiting constraint
is no longer FLOPs but candidate-axis memory.

\textbf{Setup.}  All $\Kstar$ claims below are read from \emph{full-catalogue}
evaluation (the deployed full-catalogue candidate set; \S\ref{sec:setup}), not
from the training sampled-softmax whose candidate count
$V_\text{softmax}\!=\!16{,}384\!+\!K$ changes with~$K$.  We sweep
$K\!\in\!\{0,16\mathrm{k},32\mathrm{k},64\mathrm{k},131\mathrm{k},262\mathrm{k},524\mathrm{k},1\mathrm{M},2\mathrm{M}\}$ at
five compute budgets ($102$ training runs, $123$ full-catalogue evals), with
the contextualizer sized at the model/data grid-winner $\Nstar$ for each
budget (Table~\ref{tab:phase3_alloc}, grid-winner column).  We model the
iso-FLOP curve as a sum of two opposing terms plus an irreducible floor,
\begin{equation}
  y(K) \;=\; \underbrace{a\,K^{\alpha}}_{\text{starvation}}
  \;+\;
  \underbrace{b\,K^{-\beta}}_{\text{sampling bias}}
  \;+\; E,
  \label{eq:LK}
\end{equation}
with closed-form optimum $\Kstar\!=\!(b\beta/(a\alpha))^{1/(\alpha+\beta)}$
(the starvation term is the linear-in-$K$ per-step cost shrinking the
available step count $T$ at fixed $C$; the sampling-bias term is the
$\propto\!1/K$ variance of the sampled-softmax partition-function
estimator).  In-batch vs extra-negative channel decomposition and
per-metric bias-exponent values are in Appendix~\ref{app:phase4}.

\begin{table}[!t]
  \centering
  \small
  \caption{\textbf{Analytic $\Kstar$ from the fitted starvation/bias
  model (\ref{eq:LK}), per budget and per criterion.}  The eval curves
  are nearly flat in $\log K$, so the analytic optima carry wide CIs;
  even so, $\Kstar$ for every metric clusters around
  $\sim\!10^{5}$--$10^{6}$.  Bold cells are inside the swept range
  $K\!\in\![16\mathrm{k},2\mathrm{M}]$; italic ``boundary'' cells indicate the
  extrapolated $\Kstar$ falls outside the sweep so we report the right
  edge; values in parentheses are the discrete-grid winners.  $\beta$
  is the per-metric bias-decay exponent (sampling-bias regime; see
  \S\ref{sec:phase4}); $\alpha$ is the (shared) starvation exponent.}
  \label{tab:phase4}
  \begin{tabular}{lcccccc}
    \toprule
                & \multicolumn{4}{c}{Analytic $\Kstar$ (discrete-grid argmax in parentheses)}
                & \multicolumn{2}{c}{Bias exponent $\beta$} \\
    \cmidrule(lr){2-5} \cmidrule(lr){6-7}
    Budget & val\_loss            & recall@10            & NDCG@10            & MRR@10
           & val\_loss            & recall@10            \\
    \midrule
    $10^{15}$ & \textbf{$360$k} (2\,M)    & \emph{bound.} (0)         & \textbf{$125$k} (33\,k)   & \textbf{$212$k} (33\,k)   & $1.24$ & $2.00^{\dagger}$ \\
    $10^{16}$ & \emph{bound.} (2\,M)      & \textbf{$437$k} (16\,k)   & \textbf{$412$k} (65\,k)   & \textbf{$420$k} (65\,k)   & $1.19$ & $0.39$           \\
    $10^{17}$ & \textbf{$403$k} (1\,M)    & \textbf{$249$k} (65\,k)   & \textbf{$245$k} (65\,k)   & \textbf{$237$k} (65\,k)   & $1.24$ & $0.40$           \\
    $10^{18}$ & \emph{bound.} (2\,M)      & \textbf{$632$k} (2\,M)    & \emph{bound.} (2\,M)      & \textbf{$489$k} (2\,M)    & $1.17$ & $0.41$           \\
    $10^{19}$ & \emph{bound.} (2\,M)      & \emph{bound.} (2\,M)      & \emph{bound.} (2\,M)      & \emph{bound.} (2\,M)      & $0.55$ & $0.87$           \\
    \bottomrule
  \end{tabular}
  \\[2pt]
  {\footnotesize $^{\dagger}$ At $C\!=\!10^{15}$ the recall@10 fit hits
  $\beta\!=\!2$ (upper bound) because the swept-range curve is monotone;
  no interior $\Kstar$.}
\end{table}

\begin{figure}[!t]
  \centering
  \includegraphics[width=0.95\linewidth]{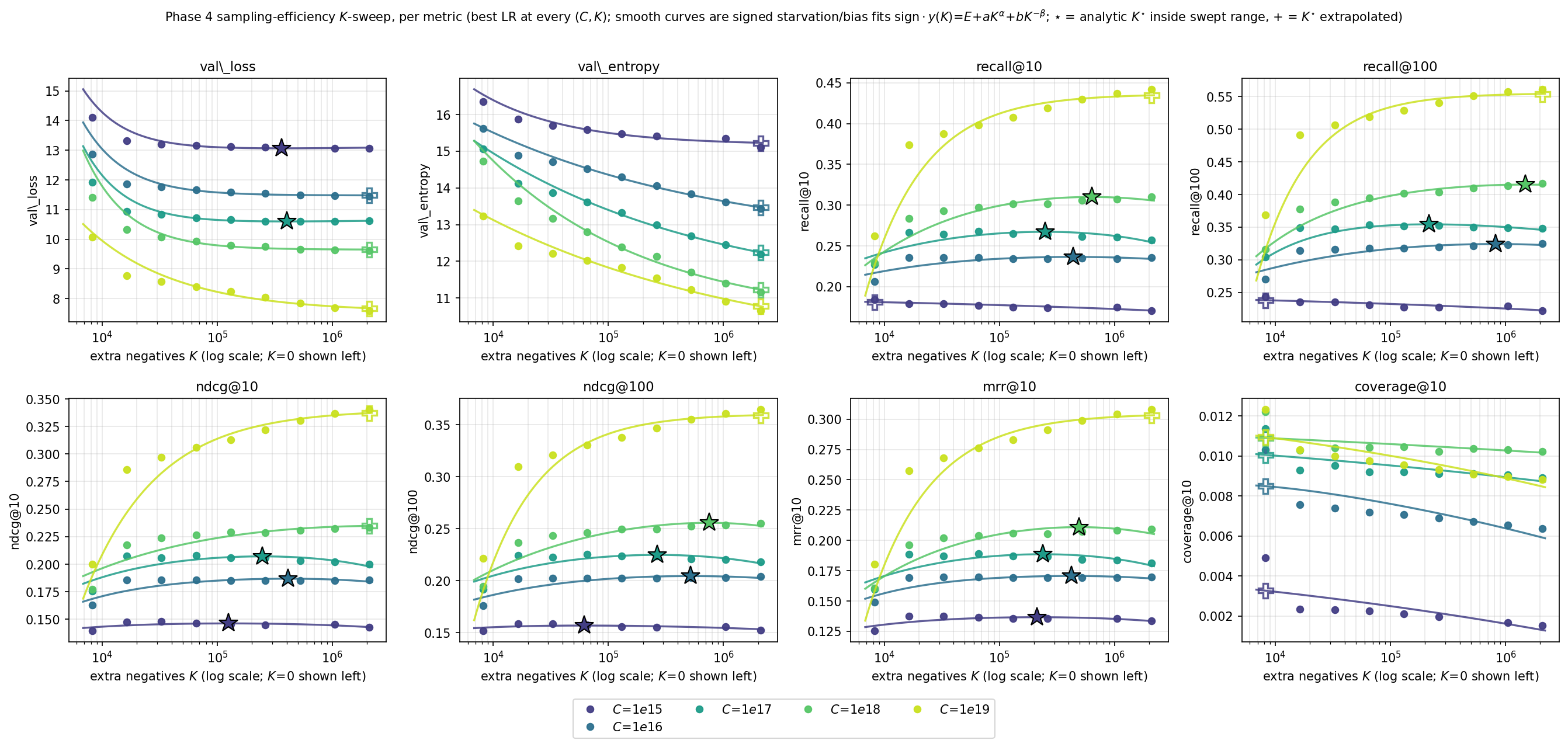}
  \caption{\textbf{Negative-sampling $K$-sweep, per headline metric, with the
  per-(metric,$C$) starvation/bias fit \eqref{eq:LK} overlaid.}
  $\star$: analytic $\Kstar$ inside the swept range
  $K\!\in\![16\mathrm{k},2\mathrm{M}]$; $+$: $\Kstar$ extrapolated outside the
  sweep (boundary cases in Table~\ref{tab:phase4}; $13$ of $25$
  metric$\times$budget cells).  Three things are visible directly off
  the parabolas: (a)~at $C\!\leq\!10^{17}$ the ranking panels are flat
  enough across $K\!\in\![10^{5},10^{6}]$ that the discrete-grid argmax
  moves nearly a decade between adjacent cells on noise alone, motivating
  the smooth-fit estimator of (i) below; (b)~\texttt{val\_loss}
  and \texttt{val\_entropy} keep falling all the way to $K\!=\!2$M at
  $C\!\in\!\{10^{16},10^{18},10^{19}\}$, the right-edge boundary
  cases of Table~\ref{tab:phase4}; and (c)~at $C\!=\!10^{19}$ every
  panel is still trending at $K\!=\!2$M, consistent with the
  memory-bound flip in finding (iii).  The compact $\Kstar(C)$ summary
  on the interior cells is Figure~\ref{fig:phase4_kstar_vs_C} in
  Appendix~\ref{app:phase4}.}
  \label{fig:phase4_metrics}
\end{figure}

\textbf{Findings.}  The main result is that the ``right'' $K$ is
metric-dependent.  On the discrete grid, full-catalogue
\texttt{val\_loss} and \texttt{recall@10} can select very different
points: at $C\!=\!10^{17}$ the grid winners are $1$M and $65$k negatives,
respectively.  Reading from the smooth fit makes the disagreement much
smaller ($400$k vs.\ $250$k), but it does not remove it.  Loss-like
metrics still prefer more negatives than ranking metrics whenever both
have interior optima (Table~\ref{tab:phase4}).

The actionable range is narrower than the raw grid suggests.  Across the
interior fits, the ranking-metric optima lie in
$\Kstar\!\in\![125\text{k},\,870\text{k}]$.  The fitted cross-budget
slope is weak ($\Kstar\!\propto\!C^{0.08}$--$C^{0.15}$), and many curves
are shallow in $\log K$, so we treat the band rather than the slope as the
practitioner summary.  The fitted bias exponent also separates loss from
ranking: $\beta\!\approx\!1.2$ for \texttt{val\_loss} versus
$\beta\!\in\![0.39,0.56]$ for ranking metrics (Appendix~\ref{app:phase4}).

At the largest budget, $K$ stops looking compute-limited.  At
$C\!=\!10^{19}$, every headline metric is still improving at the largest
value we trained ($K\!=\!2$M), and the analytic optimum falls beyond the
swept range for every metric.  The bottleneck is then the candidate-axis
softmax-logit memory footprint, which grows linearly with $K$, not the
available FLOPs.  Pushing beyond this regime requires candidate-axis
checkpointing, candidate sharding, or a sampled/hierarchical approximation
to the partition function.

The resulting recipe is simple: use $K$ in the low hundreds of thousands
when memory is not binding, cap near $10^6$ around $C\!=\!10^{18}$, and
treat $C\!\geq\!10^{19}$ as a memory-engineering problem rather than a
pure compute-allocation problem (Table~\ref{tab:recipe}).

% =====================================================================
\section{Cross-Metric and Cross-Regime Evaluation}
\label{sec:metrics}

The experiments above point to a single methodological lesson: in behavioral
foundation models, the evaluation metric is part of the scaling law.  Changing
it can change the compute-optimal recipe.  The optimizer sees a sampled-softmax
loss, while the deployed system serves a full-catalogue ranking metric.  Those
quantities are often correlated, but they do not always choose the same batch
size, architecture cell, or negative-sampling recipe.  This
section gathers the evidence across the study: metric-specific critical batch
sizes, metric-specific $\Kstar$, a compute-dependent loss--ranking sign flip,
and the asymmetry between batch-local and full-catalogue evaluation.  The
analysis pools the $408$ validation evaluations across the architecture,
allocation, and sampling experiments.

\begin{figure}[!t]
  \centering
  \begin{subfigure}[t]{0.45\linewidth}
    \includegraphics[width=\linewidth]{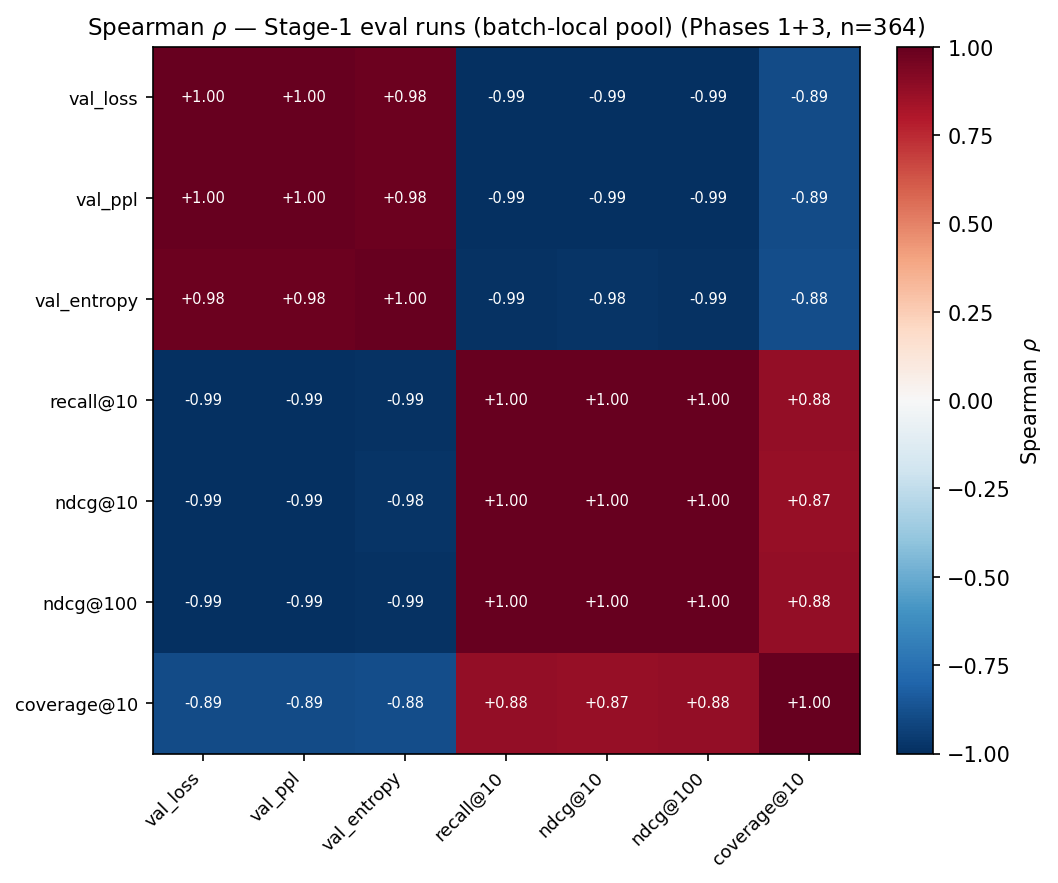}
    \caption{Stage~1 architecture/allocation sweeps ($n\!=\!285$).}
  \end{subfigure}\hfill
  \begin{subfigure}[t]{0.45\linewidth}
    \includegraphics[width=\linewidth]{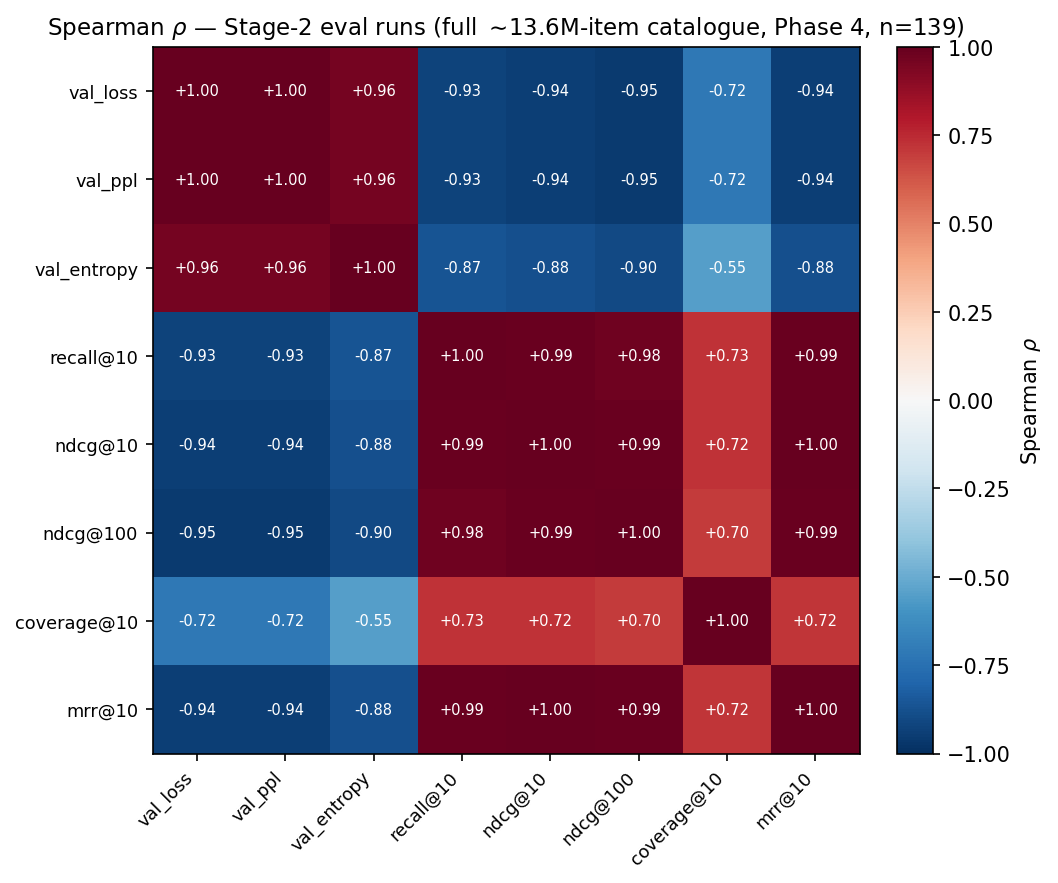}
    \caption{Negative-sampling sweep after freezing ($n\!=\!123$).}
  \end{subfigure}
  \caption{\textbf{Spearman rank correlations between headline metrics.}
  Within either stage the loss/perplexity/entropy/ranking metrics are
  essentially one quantity; coverage is the only metric that decouples
  meaningfully.}
  \label{fig:metric_corr}
\end{figure}

\textbf{Within-stage correlations are tight.}  Within either stage,
\texttt{val\_loss}, \texttt{val\_ppl}, \texttt{val\_entropy} and the
ranking metrics (\texttt{recall@}$k$, \texttt{NDCG@}$k$ for
$k\!\in\!\{1,5,10,20,50,100\}$) are mutually correlated at
$|\rho_S|\!\geq\!0.94$ (Figure~\ref{fig:metric_corr}).  Coverage is the
only metric that decouples meaningfully, and the loss--coverage link
weakens monotonically with scale (full table in
Appendix~\ref{app:metrics}): once catalogue coverage saturates,
\emph{which} cells happen to spread the head distribution furthest is
essentially independent of which cells minimize loss.

\textbf{High correlations are not the whole story.}  Two metrics with
$\rho_S\!=\!-0.99$ can still disagree about which cell wins each
budget when the leaderboard is tightly bunched.  The architectural takeaways
(\S\ref{sec:phase1}) are metric-robust: the optimal embedder share and
depth coincide under loss, recall and NDCG.  The per-budget
winners and within-regime correlations on every other axis are not.
The four axes (a)--(d) below itemize where.

\paragraph{(a)~Per-metric critical batch size (\S\ref{sec:phase2}).}
The Kaplan fit~\eqref{eq:bcrit} returns $\Bcrit\!\approx\!200$--$275$
for position-weighted ranking (\texttt{NDCG@10}, \texttt{MRR@10}),
$\sim\!570$ for \texttt{val\_loss} and \texttt{recall@10}, and no
plateau within the swept range for \texttt{val\_entropy}: same model,
same iso-target machinery, four different knees
(Table~\ref{tab:phase2}).  A downstream metric $M\!=\!f(L)$ with
small $|f'(L)|$ in the operating band reaches its Kaplan curvature
regime at smaller $B$, which is the source of the spread
(\S\ref{sec:phase2}, finding~(ii)).

\paragraph{(b)~Per-metric Stage~2 $\Kstar$ (\S\ref{sec:phase4}).}
The analytic minima of the iso-$C$ fit~\eqref{eq:LK}
disagree across metrics at every budget: at $C\!=\!10^{17}$,
$\Kstar_{\text{loss}}\!=\!400$k vs.\
$\Kstar_{\text{recall@10}}\!=\!250$k, and across the four interior
budgets every ranking-metric $\Kstar$ lies in
$[125\mathrm{k},\,870\mathrm{k}]$.  The loss landscape in $\log K$ is shallow at
the larger budgets, so the analytic optima carry wide confidence
intervals, but the loss-vs-ranking ordering is preserved at every budget:
loss prefers more negatives than the ranking metrics
where both have interior optima.

\begin{figure}[!t]
  \centering
  \includegraphics[width=0.95\linewidth]{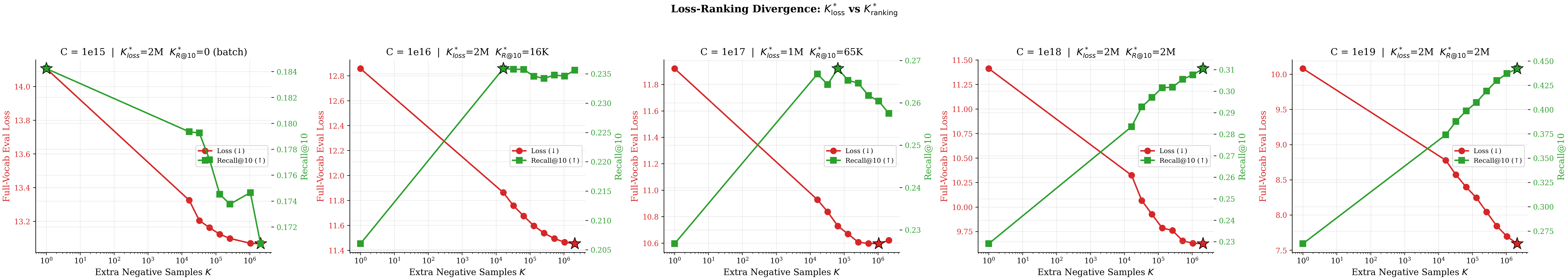}
  \caption{\textbf{Full-catalogue \texttt{val\_loss} vs.\ \texttt{recall@10}.}
  At fixed $C$, each column is one iso-FLOP budget; curves use the same
  deployed-catalogue eval pool (only training $K$ differs).  Red:
  \texttt{val\_loss} ($\downarrow$); green: \texttt{recall@10}
  ($\uparrow$); stars mark the swept-grid argmax per metric.}
  \label{fig:phase4_lvr}
\end{figure}

\paragraph{(c)~Stage~2 loss-vs-ranking rank correlation flips sign with
compute.}  In the Stage~2 $K$-sweep, the Spearman correlation between
\texttt{val\_loss} (smaller-better) and \texttt{recall@10}
(larger-better) runs from $\rho_S\!=\!+0.93$ at $C\!=\!10^{15}$
(misaligned) to $\rho_S\!=\!-1.00$ at $C\!=\!10^{18}$ (perfectly
aligned) (Table~\ref{tab:corr_by_budget_stage2}).  By contrast the Stage~1
loss-vs-ranking link is locked at $\rho_S\!\approx\!-0.98$ across all
budgets (Table~\ref{tab:corr_by_budget}).  Figure~\ref{fig:phase4_lvr}
shows the same effect directly on the full-catalogue curves: the loss
and recall winners can separate at low compute and realign at larger
budgets.  The sign flip arises because
the source of variation differs: Stage~1 varies \emph{architecture}
(bigger model $\to$ both lower loss and higher recall, locked), while
Stage~2 varies $K$.  More negatives sharpen the conditional but can
eventually flatten or worsen ranking.  The alignment at $C\!=\!10^{18}$
confirms that once popularity bias is removed (large enough $K$ for the
sampling distribution to approach uniform), the two regimes agree on
the ranking of $K$-cells.

\textbf{Why training loss and ranking can disagree: mechanism for
(c) and (d).}  At each step the model is scored against a candidate set
$\mathcal{B}\!\subset\!\mathcal{V}$ drawn from a sampling distribution
$q$ and minimizes the sampled cross-entropy
\[
\mathcal{L}(h,\mathcal{B})
\;=\;
-\log\frac{\exp\langle h,e_\theta(y^{+})\rangle}
            {\sum_{x\in\mathcal{B}}\exp\langle h,e_\theta(x)\rangle}.
\]
By a standard importance-weighting argument~\citep{bengio2008adaptive},
this is an unbiased estimator of the full softmax only if each candidate
logit is corrected by $-\log q(x)$; without that correction the unique
minimizer satisfies
$f_\theta^{\star}(y\!\mid\!h)\!\propto\!p(y\!\mid\!h)/q(y)$.  In
Stage~1 each candidate is itself a target, so $q\!\propto\!p$ and the
loss minimizer is uniform on the catalogue (absolute popularity is
unidentifiable from the in-batch objective; only within-batch rank
order is learned).  In Stage~2 with $K$ uniform extras, $q$ approaches
uniform as $K\!\to\!\infty$ and the minimizer approaches the true
conditional.  Eval metrics score against the full catalogue and depend
on the marginal $f_\theta$, not just its within-batch ranks; the $1/q$
correction the model never had to learn shows up directly in the
loss-to-NDCG mapping, mechanistically producing both the sign-flip of
(c) and the cross-regime asymmetry of (d).

\begin{figure}[!t]
  \centering
  \includegraphics[width=0.99\linewidth]{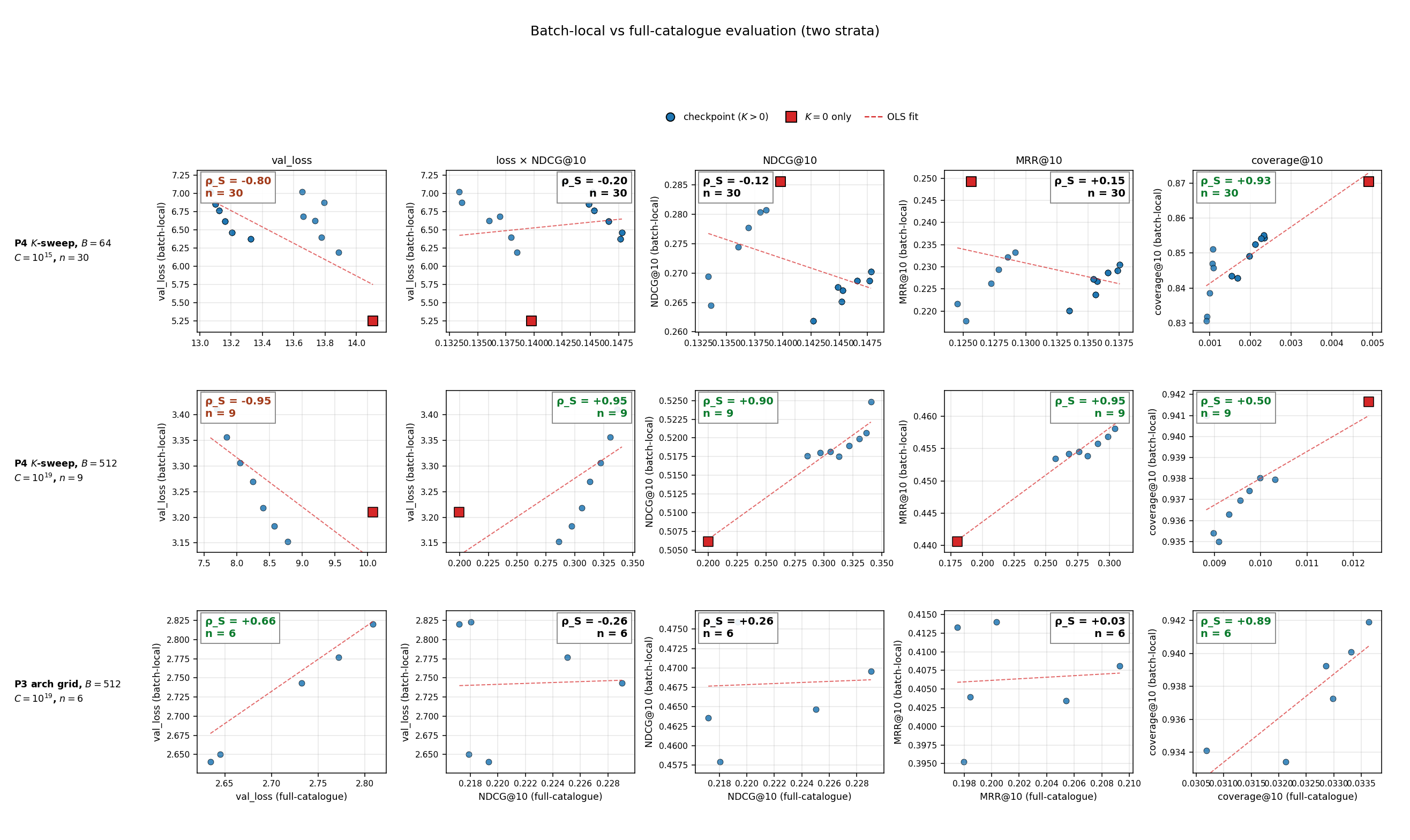}
  \caption{\textbf{Batch-local vs.\ full-catalogue evaluation.}
  Each point is one checkpoint scored under both regimes.
  Red squares mark Phase~4 $K\!=\!0$ only.
  The first two rows are Stage~2 $K$-sweep strata ($B\!=\!64$, $C\!=\!10^{15}$
  and $B\!=\!512$, $C\!=\!10^{19}$).  The third row is the Phase~3 iso-FLOP
  architecture grid at $C\!=\!10^{19}$, $B\!=\!512$ ($n\!=\!6$ cells
  re-evaluated under full-catalogue).  Column~2 is the cross-metric panel:
  batch-local \texttt{val\_loss} vs.\ full-catalogue \texttt{NDCG@10}; the
  other columns match metrics on both axes.  In the Phase~4 $K$-sweep, loss is
  anti-correlated across regimes while ranking metrics stay strongly positively
  correlated.  In the Phase~3 arch grid, the cells cluster tightly within each
  regime, so \texttt{NDCG@10} and \texttt{MRR@10} carry essentially no
  cross-regime ranking signal whereas \texttt{coverage@10} and
  \texttt{val\_loss} remain positively correlated.}
  \label{fig:p4_local_vs_global}
\end{figure}

\paragraph{(d)~Batch-local and full-catalogue evaluation disagree mainly on
loss.}  Batch-local evaluation and full-catalogue evaluation use different
candidate sets, so their losses need not rank checkpoints the same way.  This
is most visible in the Stage~2 $K$-sweep, where changing $K$ also changes the
sampling distribution behind the sampled-softmax objective.  In paired
re-evaluations of the same checkpoints, batch-local \texttt{val\_loss} is
therefore a poor proxy for full-catalogue \texttt{val\_loss}
($\rho_S\!=\!-0.95$ at $B\!=\!512$, $C\!=\!10^{19}$).  Ranking metrics behave
differently: batch-local ranking metrics, especially \texttt{NDCG@10} and
\texttt{MRR@10} at $B\!=\!512$, remain strongly correlated with full-catalogue
ranking metrics ($\rho_S\!=\!+0.90$ and $+0.95$ respectively;
Figure~\ref{fig:p4_local_vs_global}).  The Phase~3 iso-FLOP architecture grid
at the same budget tells a slightly different story: the six cells cluster
tightly enough that the cross-regime ranking signal is dominated by noise on
\texttt{NDCG@10} ($\rho_S\!=\!+0.26$) and \texttt{MRR@10}
($\rho_S\!=\!+0.03$), while \texttt{coverage@10} stays the most stable proxy
($\rho_S\!=\!+0.90$, $n\!=\!6$) and \texttt{val\_loss} actually correlates
\emph{positively} ($\rho_S\!=\!+0.66$).  So batch-local ranking metrics are a
good proxy for full-catalogue ranking metrics when the cells span a real
quality gap (the $K$-sweep), but lose discriminative power when the cells are
quality-equivalent (the iso-FLOP arch grid).  We therefore use batch-local
ranking metrics as a practical proxy for comparing architecture/allocation
cells, with the caveat that the marginal ranking gap between near-tied cells
is unreliable across regimes.

\paragraph{(e)~Ranking stability across scoring-history lengths depends on
the sweep.}  For each evaluated checkpoint we log metrics at scoring-history
lengths $cl\!\in\!\{3,5,10,20,50,100\}$.  For each metric and compute budget,
we then ask whether cells are ranked similarly at different scoring-history
lengths: for example, do the same checkpoints win at $cl\!=\!3$ and
$cl\!=\!100$?  Architecture/allocation sweeps are stable under this test:
ranking metrics have worst-case pairwise Spearman correlations
$\rho_{\min}\!\geq\!0.93$ at $C\!\leq\!10^{18}$.  Stage~2 $K$ sweeps are less
stable at smaller budgets: for $C\!\leq\!10^{17}$, the worst-case correlations
across scoring-history lengths drop to $\rho_{\min}\!\in\![0.26,0.73]$,
meaning that the best $K$ can depend on whether evaluation emphasizes short-
or long-history queries.  By $C\!\geq\!10^{18}$, these Stage~2 rankings
realign ($\rho_{\min}\!\geq\!0.87$).  Full 7$\times$7 slice-pair matrices and
the worst-case-$\rho$-vs-$C$ summary are in Appendix~\ref{app:context_length}
(Figures~\ref{fig:ctxlen_minrho}, \ref{fig:ctxlen_heatmap_bl},
\ref{fig:ctxlen_heatmap_fv}).

\textbf{Cross-budget exponents are metric-robust.}  Despite (a)--(d)
above, the cross-budget Hoffmann \emph{exponents} agree across metrics
to within slope uncertainty (\S\ref{sec:phase3},
Table~\ref{tab:exponents}): every metric lies in
$a_N\!\in\![0.57,\,0.66]$, with the
\texttt{val\_loss}-vs-\texttt{NDCG@10} separation
$\Delta a_N\!\approx\!0.03$.  So the disagreement is in the
per-budget recipe, namely which $(h,L)$ cell or which $K$ to ship, not
in the scaling law itself.

\textbf{Pick the target before fitting the scaling law.}  How much metric choice matters varies by axis.  In the model/data allocation sweep, the
per-metric Hoffmann
exponents agree within slope uncertainty
($\Delta a_N\!\approx\!0.04$) and per-budget winners differ by
$\leq\!15\%$ in $\Nstar$: the scaling law transfers across metrics,
the per-budget recipe approximately so.  In the negative-sampling sweep, the per-budget
$\Kstar$ spreads $\sim\!1.5$--$3\!\times$ across metrics at the same
budget, the cross-budget exponent itself spreads
($\Kstar\!\propto\!C^{0.08}$--$C^{0.15}$), and the bias-decay
exponent $\beta$ splits cleanly between loss-like and ranking
metrics ($\beta_{\text{loss}}\!\approx\!1.2$ vs.\
$\beta_{\text{ranking}}\!\in\![0.39,\,0.56]$): the scaling law
itself does not transfer.  The strong within-stage correlations of
Figure~\ref{fig:metric_corr} do not save the practitioner from this
asymmetry: two metrics that move together can still disagree on the
per-budget $\Kstar$.  Selecting the deployed target metric
\emph{before} fitting a scaling law, not after, is the single
practical recommendation we would carry to other behavioral
foundation models, and matters most for the negative-sampling axis.

% =====================================================================
\section{Related Work}
\label{sec:related}

\textbf{Scaling laws for language models.}  The empirical-scaling-laws
program of Kaplan et~al.~\citep{kaplan2020} and
Hoffmann et~al.~\citep{hoffmann2022} established that loss is well-described
by power laws in compute, parameters and tokens, and that the
compute-optimal allocation lies near $D/N\!\approx\!20$ for text language
modeling.  McCandlish et~al.~\citep{mccandlish2018} introduced the critical
batch size as an orthogonal axis, and Yang and Hu~\citep{yang2021} reframed
initialization and LR transfer.  Our setup transposes this program to a
different objective (sampled in-batch contrastive loss), a different
architecture (two-part feature embedder $\rightarrow$ contextualizer), and
an evaluation regime in which the training loss and the deployed ranking
metric do not coincide (\S\ref{sec:metrics}).

\textbf{Recommender systems and feature-rich foundation models.}  The
two-tower factorization has its roots in the YouTube deep
recommender~\citep{covington2016}.  Recent industrial systems generalize the
recipe to dynamic catalogues with feature-based encoders, including Visa
TREASURE~\citep{visa2025treasure}, TransactionGPT~\citep{visa2025tgpt},
Stripe PFM~\citep{stripe2025pfm}, Revolut PRAGMA~\citep{revolut2026pragma},
and J.P.~Morgan TradeFM~\citep{jpmorgan2025tradefm}.  Generative recommenders
such as HSTU~\citep{zhai2024hstu} and Wukong~\citep{zhang2024wukong} demonstrate
favorable single-axis scaling in $N$.  Ardalani
et~al.~\citep{ardalani2022dlrm} report scaling laws for DLRM-style hybrids and
Netflix's foundation model~\citep{netflix2025fm} reports that scaling-up
monotonically improves quality without quantitative laws.  None of these works
jointly varies architecture, batch, $(N,D)$ allocation and negative sampling on
a single stack, none reports per-metric power-law exponents, and none
quantifies the embedder/contextualizer share as a scaling axis.

\textbf{Behavioral foundation models.}  The BehaviorGPT
line~\citep{unboxai2025consumption, unboxai2025workforce,
unboxai2025aesthetics} and its generalization to ``Large Behavioral
Models''~\citep{unboxai2026lbm} argue that the embedder must be trained
end-to-end on the sequential task and frame action-sequence modeling as its
own foundation-model paradigm, complementary to language modeling.  Within
that paradigm, the present paper supplies the joint scaling-law calibration
that the scaling-law guidance of Table~\ref{tab:recipe} relies on.

% =====================================================================
\section{Discussion}
\label{sec:discussion}

The four sweeps give a practical recipe for the current two-part behavioral
modeling stack and its two-stage training recipe.  The robust recommendation is
not a single number so much as an ordering of decisions.  First choose the
deployment metric; then set a small
event embedder, choose the batch size based on the data-efficiency/throughput
tradeoff, allocate compute between $N$ and $D$, and finally tune the frozen-embedder
negative pool under the full-catalogue ranking metric.

\begin{table}[!t]
  \centering
  \footnotesize
  \setlength{\tabcolsep}{3pt}
  \caption{\textbf{Scaling-law guidance distilled from the four sweeps.}
  Depth is omitted: the depth sweep shows $L_{\text{ctx}}$ at the val
  minimum varies by budget while induced $s$ stays in the width-sweep
  band (Appendix~\ref{app:phase1d}).}
  \label{tab:recipe}
  \begin{tabular}{p{0.17\linewidth}p{0.62\linewidth}p{0.13\linewidth}}
    \toprule
    Knob & Guidance & Source \\
    \midrule
    Initialization      & Default (truncated-normal--style) & Appendix~\ref{app:mup} \\
    Embedder share      & $s\!\approx\!2\%$ (primary architectural knob) & \S\ref{sec:phase1w} \\
    Batch size          & $B\!\approx\!2048$: largest single-node batch tested, past the $\Bcrit$ for \texttt{val\_loss}, and throughput-optimal & \S\ref{sec:phase2} \\
    LR                  & $\sim\!2\!\cdot\!10^{-3}\!\sqrt{B/512}$, cosine $+$ $5\%$ warm-up & \S\ref{sec:phase2}--\ref{sec:phase3} \\
    $D/N$               & Parabolic \texttt{val\_loss} optima decrease with compute: $344\!\to\!265\!\to\!222\!\to\!110\!\to\!36$ from $10^{15}$ to $10^{19}$ FLOPs, approaching the text-LM heuristic & \S\ref{sec:phase3} \\
    Extra negatives $K$ & $K\!\in\![2.5\!\cdot\!10^{5},\,9\!\cdot\!10^{5}]$ from the analytic $\Kstar$ band; cap near $10^{6}$ at $C\!=\!10^{18}$ & \S\ref{sec:phase4} \\
    \bottomrule
  \end{tabular}
\end{table}

\paragraph{Maximal Update Parameterization: a negative result.}
We swept Maximal Update Parameterization (MuP)~\citep{yang2021} across
four model sizes ($\sim\!10$\,M to $\sim\!500$\,M total trainable
parameters) and learning rates in
$[10^{-5},\,5\!\cdot\!10^{-2}]$, against our default
truncated-normal--style initialization.  MuP delivers most of its core
LR-transferability promise: the MuP-optimum LR span across widths is
$0.30$ decades vs.\ $0.70$ for Default, and every MuP optimum is a
verified local minimum.  Default nevertheless reaches a strictly lower training
loss at every scale by $0.68$--$0.92$~nats (Appendix~\ref{app:mup},
Table~\ref{tab:mup}).  The pattern holds for every other MuP variant we
tried, including per-layer MuP and a FLOP-budget-matched variant; we
therefore retain the default initialization and pay the modest cost of a
per-phase LR sweep instead.

\paragraph{Limitations: Stage~1 evaluation is batch-local.}  The architecture
and allocation sweeps score checkpoints against the unique target embeddings in
each validation batch ($\sim\!5$--$10$k items), not against the full catalogue.
Eval batch size is fixed within each
compute budget, so within-budget rankings are apples-to-apples; this is
what every architecture/allocation takeaway depends on.  Absolute losses
across budgets are not directly comparable.  The mechanism argument for
why batch-local rankings should still transfer to the deployed
full-catalogue metric on the architecture axis is axis~(d) of
\S\ref{sec:metrics}; Figure~\ref{fig:p4_local_vs_global} sharpens it on
identical checkpoints from Stage~2 $K$-sweep cells.
% and a $C\!=\!10^{19}$ Phase~3 architecture-grid spot check (bottom row, $n\!=\!6$).
Remaining open work is to extend full-catalogue re-evaluation
across the full architecture/allocation grid and additional Stage~2
checkpoints at intermediate eval batch sizes.

\paragraph{Limitations: Architecture coverage.}  The width sweep does not vary hidden size
at fixed depth; the depth sweep does not vary text-encoder share.  A full
$5$D grid (share $\times$ width $\times$ depth $\times$ embedder-depth $\times$
LR) is the natural next step.  The $s^{\star}\!\approx\!2\%$ result is
also for the essentially-unregularized recipe we trained with; embedder
regularization may shift $s^{\star}$ upward (\S\ref{sec:phase1w}).

\paragraph{Limitations: Context length.}  All sweeps fix the
\emph{training} contextualizer sequence length at $\Lseq\!=\!256$ event tokens
per example.  Every eval batch is additionally stratified by history
position so each metric is also logged at
$cl\!\in\!\{3,5,10,20,50,100\}$ events of context (\S\ref{sec:setup}),
which axis~(e) of \S\ref{sec:metrics}
(Figure~\ref{fig:ctxlen_minrho}) uses to bound the \emph{scoring}
half of the question: Stage~1 architectural winners are
context-length-robust (\texttt{recall@10}, \texttt{NDCG@10} and
\texttt{NDCG@100} all $\rho_{\min}\!\geq\!0.93$ at $C\!\leq\!10^{18}$)
so architecture/allocation winners transfer to shorter or longer serving contexts; Stage~2
$\Kstar$ winners are not at $C\!\leq\!10^{17}$ (ranking
$\rho_{\min}\!\in\![0.26,0.73]$), realigning at $C\!\geq\!10^{18}$ on
the same compute threshold as axis~(c).  What none of this measures is the \emph{training} half:
how $\Nstar$, $\Dstar$, $s^{\star}$ and $\Kstar$ shift when models
are trained from scratch at smaller or larger~$\Lseq$, how the in-batch
$3\,B\Lseq h$ contrastive cost in~\eqref{eq:flops} scales with~$\Lseq$, and
whether the iso-FLOP trade-offs continue to be captured by the
Kaplan formula once attention's $L\!\cdot\!\Lseq^{2}$ term becomes
non-negligible.  That sweep is a planned future axis we did not vary.

\paragraph{Limitations: Compute range.}  Our primary model/data allocation budgets span
$C\!\in\![10^{15},10^{19}]$ FLOPs. Exploring larger budgets remains open work.

% =====================================================================
\section{Conclusion}

Behavioral foundation models need their own scaling laws.  We study the
now-common two-part stack, a feature-based event embedder feeding a
transformer contextualizer, under the two-stage recipe in which the embedder
is first trained jointly, then frozen while the contextualizer is trained with
extra negatives.  Across this setting, the compute-optimal event embedder is
small: roughly two percent of parameters across the budgets we test.  The
reason is structural: embedder parameters are more expensive per step and see a
far more repetitive effective data distribution than contextualizer parameters.

The compute allocation law is also distinctive.  Behavioral models are strongly
data-heavy at small budgets, with $D/N$ far above the text-LM Chinchilla
heuristic, but the optimum moves toward the language-model regime as compute
increases.  The recipe around that allocation is metric-dependent: critical
batch size changes with the target metric, the useful number of negatives after
freezing changes with both compute and metric, and at $C\!=\!10^{19}$ the
negative-sampling axis becomes limited by candidate-axis memory rather than
FLOPs.

Finally, evaluation regime matters.  Batch-local loss is not a reliable proxy
for full-catalogue loss, while batch-local ranking metrics are a more practical
proxy for comparing architecture/allocation cells.  For practitioners, the most
portable lesson is therefore simple: choose the metric the system will serve,
then fit the scaling law to that metric.  In behavioral foundation models, the
evaluation metric is part of the scaling law because changing it can change the
compute-optimal recipe.
% =====================================================================
\section*{Acknowledgements}
We thank Adam Fredriksson, Alexander Junco Hagberg, Alexandros Lemonaris,
Erik Guander, Gabriel Melin, Gon\c{c}alo Marques,
Jens Palmborg, Marcel R\o d, Nicolas Sanchez, Simon Granstr\"om, and
Tom Boustedt for their support and contributions.
% =====================================================================
\bibliographystyle{plainnat}

% =====================================================================
\appendix
\phantomsection
\addcontentsline{toc}{section}{Appendix}

\appendixsection{Metric Definitions}
\label{app:metricdefs}

\paragraph{Notation and candidate set.}
Every metric scores each query position against a \emph{candidate set}
$\mathcal{C}$ and ranks its items by the dot-product score
$z_{q,j}\!=\!\langle h_q, e_j\rangle$, where $h_q$ is the contextualizer's output
(query) embedding at position $q$ and $e_j$ is the cached embedding of candidate
$j\!\in\!\mathcal{C}$.  We write $N$ for the number of scored positions,
$C\!=\!|\mathcal{C}|$ for the candidate-set size, $K$ for the cutoff (the
``@$K$''), and $\ind[\cdot]$ for the indicator.  Each query has exactly one
relevant item, its true next event; $r_q$ is that item's $1$-indexed rank in the
descending score order over $\mathcal{C}$ (so $r_q\!=\!1$ is a top hit).

\emph{The choice of $\mathcal{C}$ is the central batch-local vs.\ global-catalogue
distinction in the paper}, and it governs \emph{every} metric below, not just
coverage:
\begin{itemize}\itemsep2pt
  \item \textbf{Batch-local (Stage~1).}  Before the embedder is frozen,
  $\mathcal{C}$ is the set of \emph{unique target items in the current
  validation batch} ($C\!\lesssim\!B\Lseq$), matching the in-batch sampled
  softmax used in training.
  \item \textbf{Global catalogue (Stage~2).}  After freezing, $\mathcal{C}$ is
  the \emph{full cached deployed catalogue} ($C\!\sim\!10^{8}$), matching the
  deployed retrieval setting.  For tractable repeated evaluation we score
  against a fixed $\sim\!13.6$M-item subset of this catalogue, still
  $\sim\!1500\times$ larger than the batch-local pool.
\end{itemize}
Because $\mathcal{C}$ fixes both the ranking pool (hence $r_q$) and the softmax
normalizer (hence $p_q$ below), absolute loss, entropy, and ranking scores are
comparable only \emph{within} a stage; this is exactly why \S\ref{sec:metrics}
reports loss--ranking correlations per stage.

\paragraph{Ranking metrics~\citep{manning2008ir}.}
With a single relevant item and binary relevance ($r_q\!=\!1$ best):
\begin{align}
  \text{recall@}K &= \frac{1}{N}\sum_{q=1}^{N}\ind[\,r_q\!\le\!K\,],
  &
  \text{NDCG@}K &= \frac{1}{N}\sum_{q=1}^{N}\frac{\ind[\,r_q\!\le\!K\,]}{\log_2(r_q+1)},
  &
  \text{MRR@}K &= \frac{1}{N}\sum_{q=1}^{N}\frac{\ind[\,r_q\!\le\!K\,]}{r_q}.
\end{align}
recall@$K$ (here equal to hit@$K$) counts how often the true item lands in the
top~$K$; NDCG@$K$~\citep{jarvelin2002ndcg} and MRR@$K$~\citep{voorhees1999mrr}
additionally reward placing it near the top.  The ideal DCG is~$1$ (one relevant
item), so NDCG@$K$ is just that item's rank discount $1/\log_2(r_q+1)$.

\paragraph{Coverage.}
coverage@$K$ is the catalogue fraction the model actually surfaces in its top-$K$
predictions, a recommender diversity measure~\citep{herlocker2004eval}.  It has
two protocols; we report the \textbf{batch-local} one, because the global mask
saturates toward~$1$ once enough batches are pooled (\S\ref{sec:phase2}):
\begin{equation}
  \underbrace{\text{coverage@}K = \frac{1}{B_{\text{eval}}}\sum_{b=1}^{B_{\text{eval}}}
    \frac{\bigl|\{\text{distinct top-}K\text{ items in batch }b\}\bigr|}{C}}_{\text{batch-local (reported)}}
  \;,\qquad
  \underbrace{\frac{\bigl|\bigcup_{b}\{\text{top-}K\text{ items in batch }b\}\bigr|}{C}}_{\text{global mask}},
\end{equation}
where $B_{\text{eval}}$ is the number of evaluation batches.

\paragraph{Loss and predictive uncertainty.}
Let $p_q$ be the softmax over $\mathcal{C}$,
$p_{q,j}\!=\!e^{z_{q,j}}\big/\sum_{j'\in\mathcal{C}} e^{z_{q,j'}}$.  We report the
sampled-softmax cross-entropy, its exponential, and the mean predictive entropy
(in nats):
\begin{equation}
  \text{CE} = \frac{1}{N}\sum_{q=1}^{N}\!\bigl(-\log p_q(\text{target}_q)\bigr),
  \quad
  \text{perplexity} = e^{\text{CE}},
  \quad
  \text{entropy} = \frac{1}{N}\sum_{q=1}^{N}\!\Bigl(-\!\!\sum_{j\in\mathcal{C}} p_{q,j}\log p_{q,j}\Bigr).
\end{equation}

\appendixsection{Additional Architecture Results}
\label{app:arch}

\paragraph{Per-budget iso-FLOPs accounting.}
Table~\ref{tab:iso} writes the Kaplan per-step formula \eqref{eq:flops} out
cell by cell at the near-optimal reference share $s\!=\!6\%$ we use for the
iso-FLOPs schedule (the upper edge of the flat $s\!\in\![2,6]\%$ band; the
recommendation remains $s^{\star}\!\approx\!2\%$, \S\ref{sec:phase1w}).  All
four budgets land within $0.01\%$ of their nominal target.

\begin{table}[!t]
  \centering
  \footnotesize
  \caption{\textbf{Iso-FLOPs accounting at the near-optimal reference share
  $s\!=\!6\%$.}  $D/N$ stays at the Chinchilla $\sim\!15$ that motivated the
  schedule.}
  \label{tab:iso}
  \begin{tabular}{lccccccccc}
    \toprule
    $C$ & $h_{\text{emb}}$ & $h_{\text{ctx}}$ & $L_{\text{ctx}}$ & $B$ & $T$ & $N$ & $D$ & $F_{\text{total}}$ & $D/N$ \\
    \midrule
    $10^{15}$ &  44 & 128 &  8 &  64 &   1{,}645 &   1.8\,M &   27\,M & $1.00\!\times\!10^{15}$ & 15.3 \\
    $10^{16}$ &  76 & 232 &  8 & 128 &   2{,}579 &   5.5\,M &   85\,M & $1.00\!\times\!10^{16}$ & 15.2 \\
    $10^{17}$ & 136 & 408 &  8 & 256 &   3{,}914 &  16.9\,M &  257\,M & $1.00\!\times\!10^{17}$ & 15.2 \\
    $10^{18}$ & 188 & 452 & 24 & 256 &  14{,}193 &  60.7\,M &  930\,M & $1.00\!\times\!10^{18}$ & 15.3 \\
    \bottomrule
  \end{tabular}
\end{table}

\paragraph{Validation loss vs.\ parameter share.}
\label{app:phase1w}%
\begin{figure}[!t]
  \centering
  \includegraphics[width=0.95\linewidth]{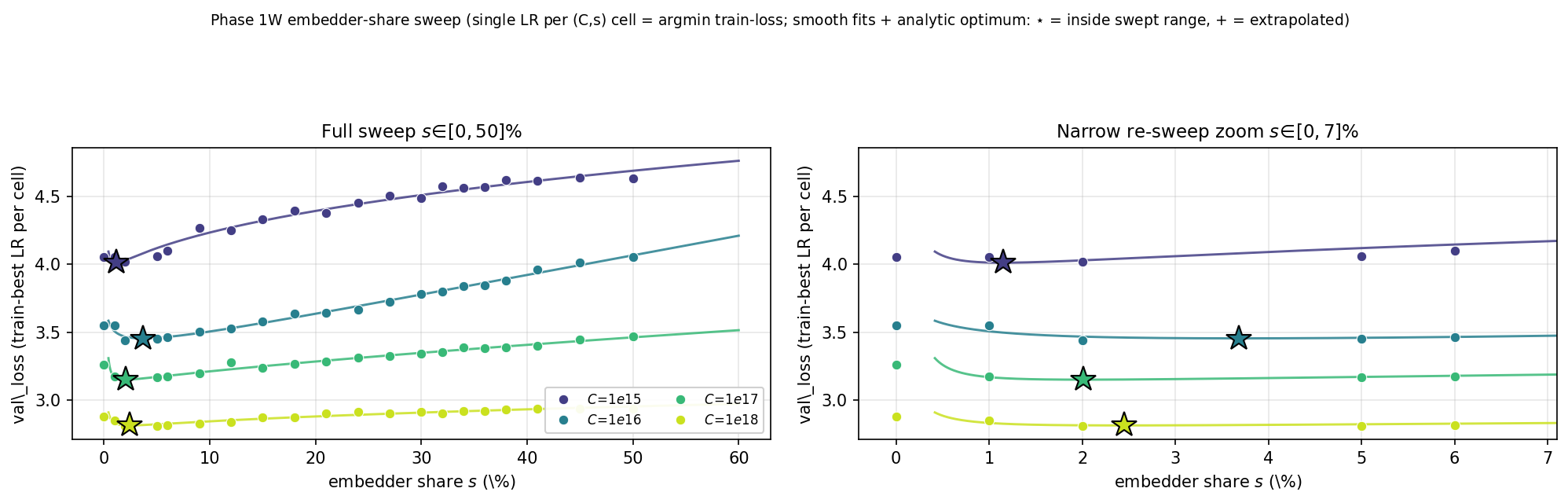}
  \caption{\textbf{Embedder-share sweep (\texttt{val\_loss} only).}
    \emph{Left:} validation loss vs.\ embedder share over $s\!\in\![0,50]\%$
    at four compute budgets.  Curves are monotone increasing in $s$ at every
    budget over $s\!\in\![6,50]\%$.  \emph{Right:} zoom on
    $s\!\in\![0,6]\%$ around the per-budget optimum.  Solid lines are
    per-budget two-term starvation fits
    $L(s)\!=\!E\!+\!a\,s^{\alpha}\!+\!b\,s^{-\beta}$; stars mark the
    closed-form analytic optimum
    $s^{\star}\!=\!(b\beta/(a\alpha))^{1/(\alpha+\beta)}$
    (Table~\ref{tab:phase1w_sstar}).  The analytic optima cluster at
    $s^{\star}\!\in\![1.1\%,\,3.7\%]$ across all four budgets (vs.\ a
    discrete-grid argmax that bounces between $s\!=\!2\%$ and $s\!=\!5\%$),
    with $s^{\star}\!\propto\!C^{+0.07}$, effectively flat in compute.
    The Kaplan-FLOP-share cross-check on the same cells is
    Figure~\ref{fig:phase1w_metrics_compute}.}
  \label{fig:phase1w_twostage}
\end{figure}

\begin{figure}[!t]
  \centering
  \includegraphics[width=0.95\linewidth]{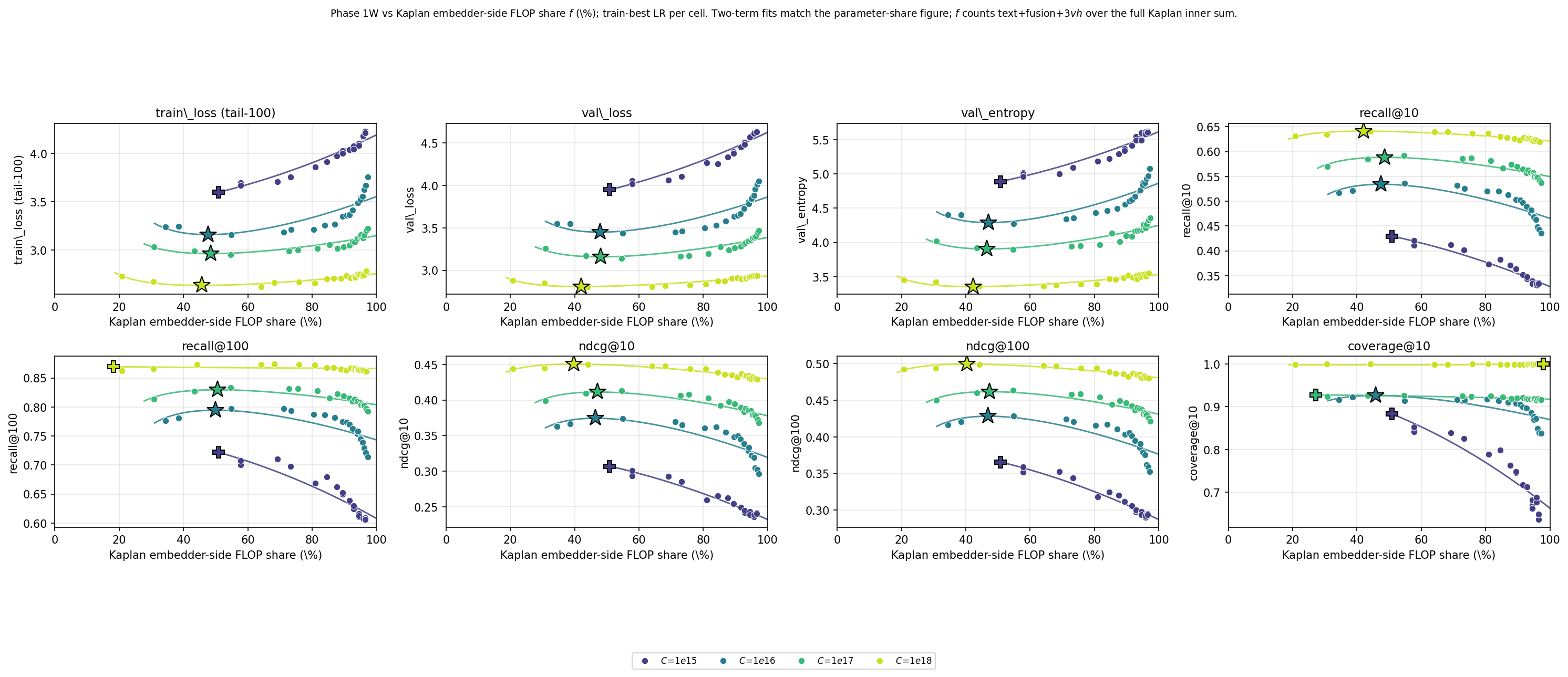}
  \caption{\textbf{Width sweep vs.\ Kaplan embedder-side FLOP share~$f$.}
  Same cells as Figure~\ref{fig:phase1w_metrics}; horizontal axis is the
  Kaplan fraction~$f$ from~\eqref{eq:flops}.  $\star$ = analytic optimum
  inside the swept range; $+$ = boundary extrapolation.  Reported
  $s^{\star}$ values use the $s$-axis fit (Table~\ref{tab:phase1w_sstar}).}
  \label{fig:phase1w_metrics_compute}
\end{figure}

\paragraph{Kaplan embedder-side FLOP share (cross-check).}
The width--share grid is controlled by a target \emph{parameter} embedder
share~$s$, which need not coincide with the fraction of the Kaplan
per-step inner sum from~\eqref{eq:flops} spent on the text embedder, the
non-text embedder stack, and the in-batch contrastive $3\,v\,h$ term
(numerator) vs.\ the contextualizer term (denominator).
Figure~\ref{fig:phase1w_metrics_compute} re-plots the same per-metric,
train-best-LR cells as the main-text Figure~\ref{fig:phase1w_metrics}
with that Kaplan fraction~$f$ on the horizontal axis; the two-term
template~\eqref{eq:share-fit} is still overlaid for visual continuity,
but we read $s^{\star}$ only from the $s$-axis fit (Table~\ref{tab:phase1w_sstar})
because the $f$-axis refit is poorly conditioned.  Here $f$ is a nonlinear,
cell-dependent pushforward of~$s$ (the $3Bsh$ term varies with the joint
embedding dim $h$ that the width search picks per cell), the swept points cover
only the band $f\!\in\![0.1,0.5]$ vs.\ $s\!\in\![0.5\%,50\%]$, and the
two-term form gets pushed to its parameter bounds.  The picture is
nonetheless useful as a sanity check: the small-embedder optimum is not a
parameter-counting artifact.

\begin{figure}[!t]
  \centering
  \begin{subfigure}[t]{0.85\linewidth}
    \centering
    \includegraphics[width=\linewidth]{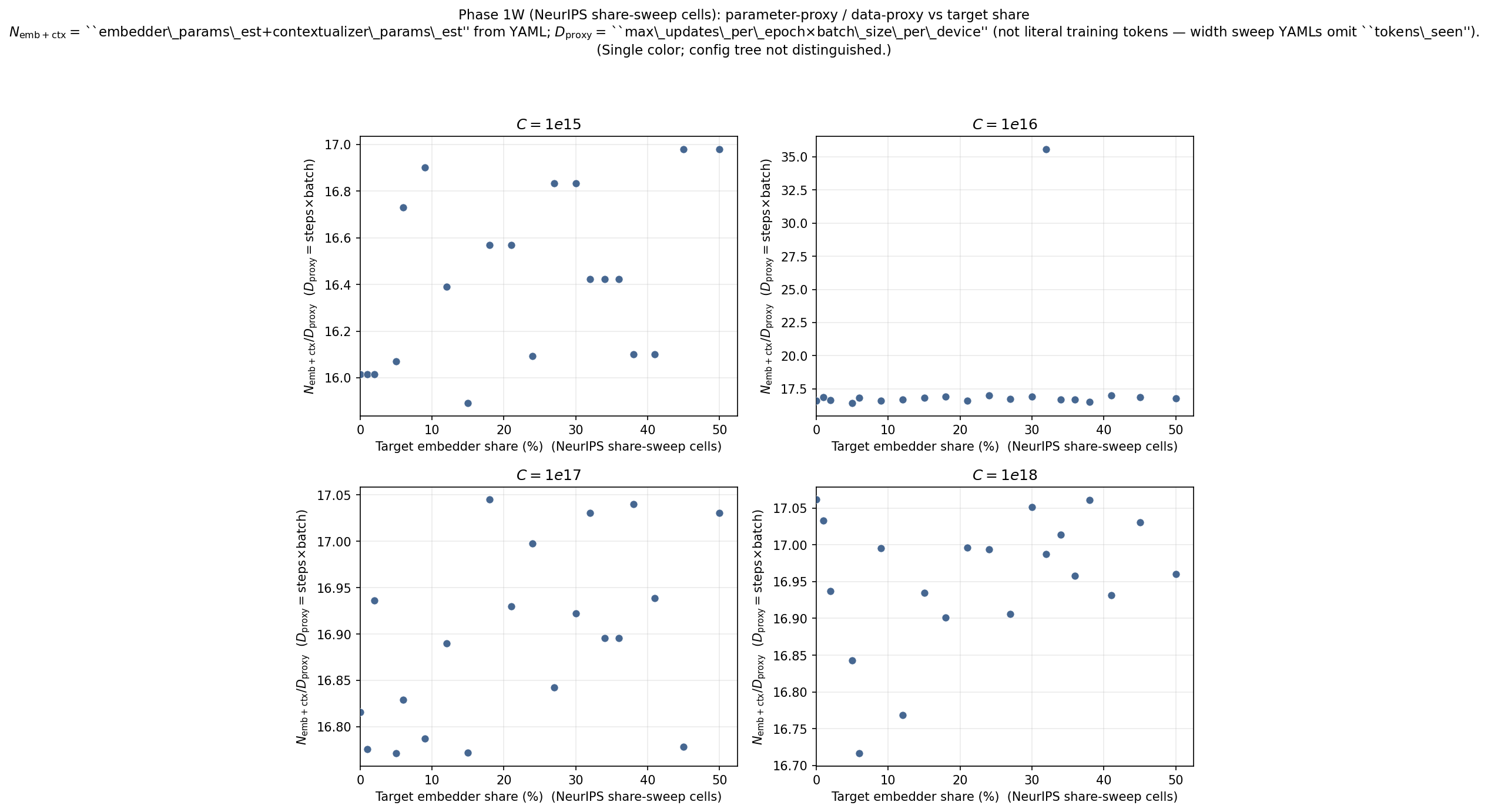}
    \caption{$N_{\mathrm{emb+ctx}}/D_{\mathrm{proxy}}$ vs.\ target embedder
    share (single color; train-best LR per cell).}
    \label{fig:phase1w_share_vs_nd_proxy}
  \end{subfigure}

  \vspace{1em}

  \begin{subfigure}[t]{0.85\linewidth}
    \centering
    \includegraphics[width=\linewidth]{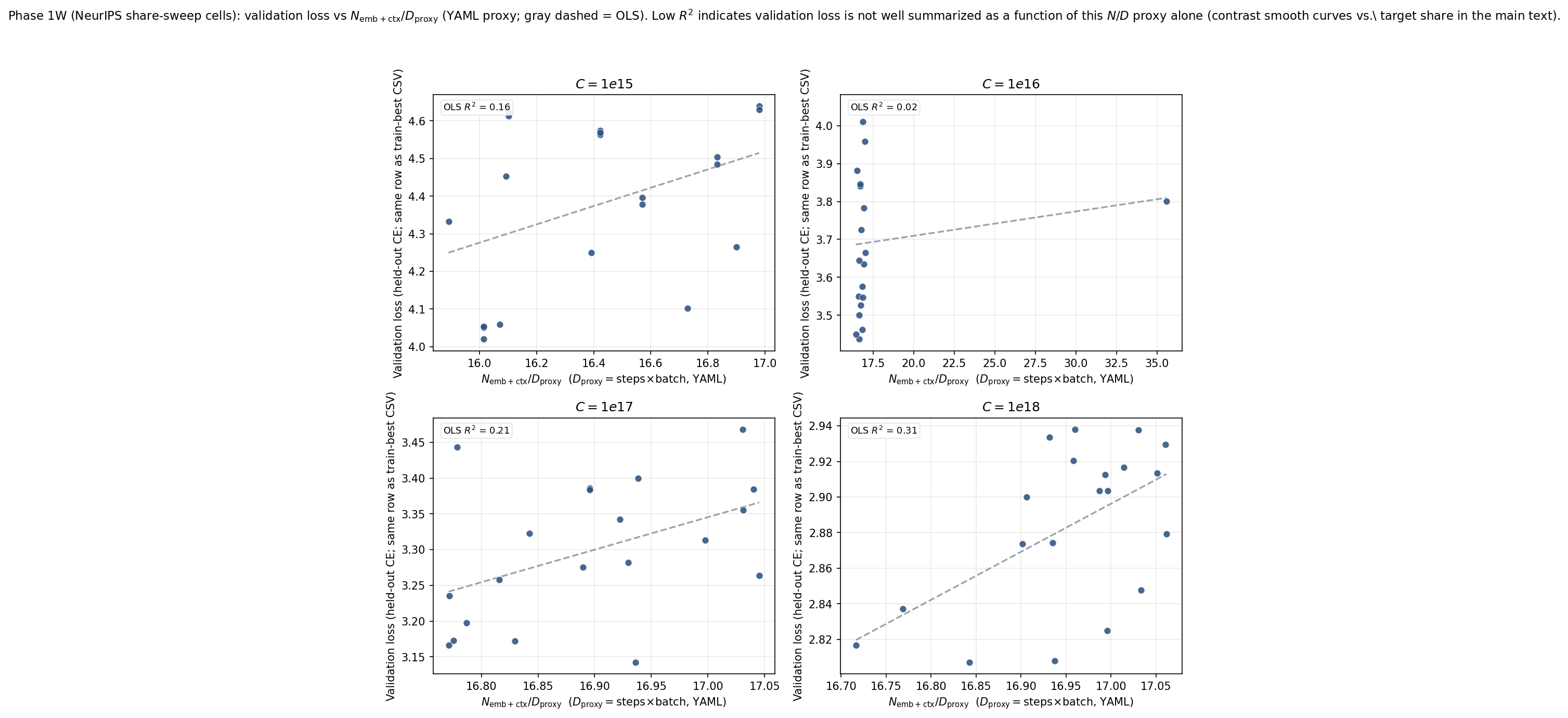}
    \caption{Validation loss vs.\ $N_{\mathrm{emb+ctx}}/D_{\mathrm{proxy}}$
    with a linear OLS overlay; per-panel $R^2$ is small relative to the
    sharp $s$-dependence in the main text.}
    \label{fig:phase1w_val_vs_nd_proxy}
  \end{subfigure}
  \caption{\textbf{$N/D$ proxy diagnostics on the width sweep.}
  Varying $s$ moves compute between embedder and contextualizer at nearly
  fixed training-to-parameter ratio: $N_{\mathrm{emb+ctx}}/D_{\mathrm{proxy}}$
  is flat in~$s$ (a), while validation loss tracks~$s$ much more strongly
  than this residual ratio (b).}
  \label{fig:phase1w_nd_proxy_appendix}
\end{figure}

\paragraph{$N/D$ proxy across the width--share sweep.}
The iso-$D/N$ framing of \S\ref{sec:phase1w} is verified empirically here.
For every train-best cell we plot a microbatch proxy for tokens seen
against total embedder-plus-contextualizer parameters (iso-FLOP accounting
at $s\!=\!6\%$ in Table~\ref{tab:iso}).
Figure~\ref{fig:phase1w_nd_proxy_appendix}(a) shows that
$N_{\mathrm{emb+ctx}}/D_{\mathrm{proxy}}$ is nearly flat in target embedder
share $s$ at each budget, so the share sweep does slide along an
approximately constant ratio rather than along a large $N/D$ move that
would confound the embedder/contextualizer tradeoff with a Chinchilla
allocation move.  Figure~\ref{fig:phase1w_nd_proxy_appendix}(b) plots
validation loss against the same ratio; the gray dashed line is an OLS fit
whose $R^2$ stays low, while loss varies much more systematically with~$s$
in Figure~\ref{fig:phase1w_twostage}, confirming that quality is driven
by the embedder--contextualizer split, not by the small residual $N/D$
movement.

\begin{figure}[!t]
  \centering
  \includegraphics[width=0.95\linewidth]{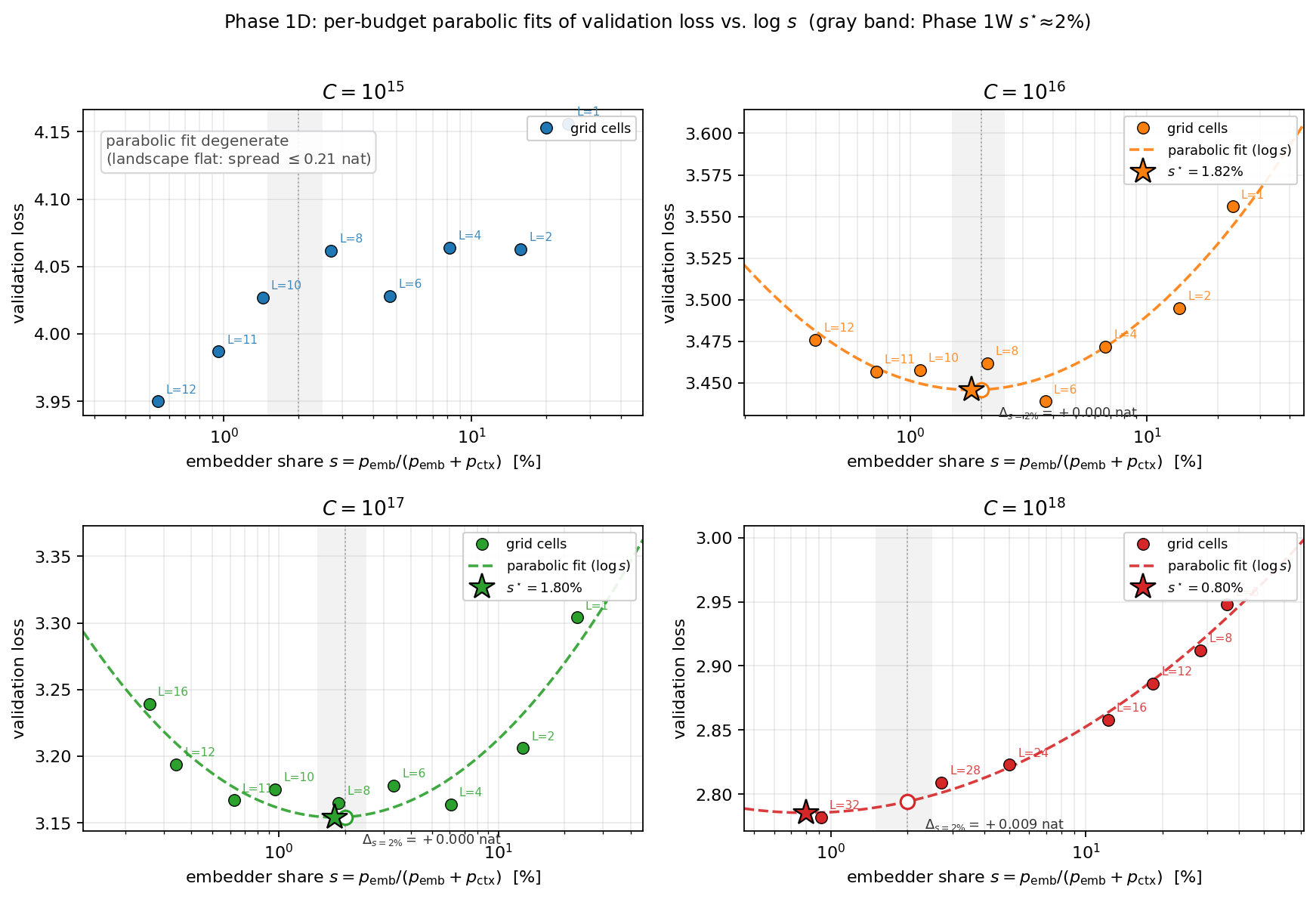}
  \caption{\textbf{Depth-sweep parabolic fits behind
  Table~\ref{tab:phase1d_parabola}.}  Per-budget panels of
  \texttt{val\_loss} against embedder share~$s$ (log scale).  Filled
  circles are the discrete depth-sum-diagonal cells (annotated with
  $L_{\mathrm{ctx}}$); the dashed curve is the parabolic fit in
  $\log_{10} s$; the star marks the analytic minimum $s^{\star}$; the
  hollow circle marks the value of the parabola at $s\!=\!2\%$ (the
  Phase~1W recommendation) with the gap $\Delta_{s=2\%}$ annotated below.
  The gray band is the Phase~1W $s\!\in\![1.5\%,\,2.5\%]$ recommendation.
  At $C\!=\!10^{15}$ the swept range is too flat ($\leq\!0.21$~nats
  end-to-end) for the parabola to resolve a sharp interior minimum;
  at $C\!=\!10^{18}$ the depth grid saturates at $L_{\mathrm{ctx}}\!=\!32$
  so the fit is monotone-trending and the analytic $s^{\star}$ should
  be read as an upper bound on the true value.  At $C\!=\!10^{16},\,10^{17}$
  the parabolic minima ($s^{\star}\!=\!1.82\%,\,1.80\%$) land
  essentially on top of the Phase~1W recommendation of $2\%$ and the
  projected loss penalty at $s\!=\!2\%$ is $\leq\!10^{-3}$~nat.  The noisy discrete
  $L_{\mathrm{ctx}}^{\star}$ jumps of Table~\ref{tab:phase1d_summary}
  are an artifact of the discrete-grid argmax, not of an underlying
  shift in the optimum.}
  \label{fig:phase1d_parabolic_fits}
\end{figure}

\begin{figure}[!t]
  \centering
  \includegraphics[width=0.65\linewidth]{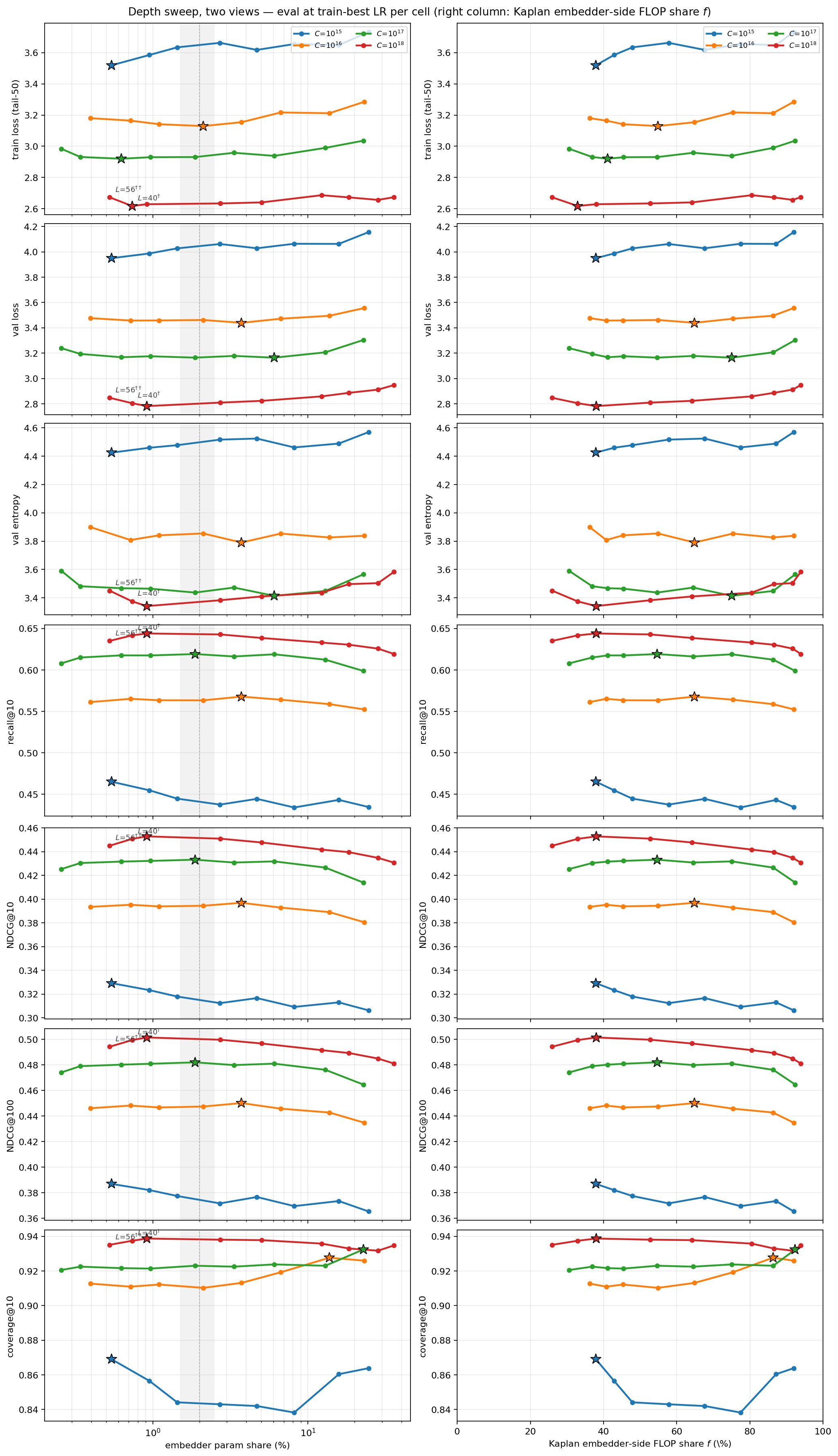}
  \caption{\textbf{Depth sweep: per-metric eval vs.\ $s$ and Kaplan FLOP
    share~$f$.}  Stars mark per-budget optima.  Gray band:
    width-sweep $s^{\star}\!\approx\!2\%$.}
  \label{fig:phase1d_eval_metrics_two_views}
\end{figure}

\paragraph{Depth sweep (per-cell val grid and per-metric eval).}%
\label{app:phase1d}%
Table~\ref{tab:phase1d_summary} lists the val-loss winner per budget;
Table~\ref{tab:phase1d_parabola} re-reads the same grid through the
parabolic estimator in $\log s$ and
Figure~\ref{fig:phase1d_parabolic_fits} shows the fits visually
overlaid on the discrete cells;
Table~\ref{tab:phase1d} is the full per-cell grid;
Figure~\ref{fig:phase1d_eval_metrics_two_views} plots every headline metric
vs.\ embedder share~$s$ and Kaplan FLOP share~$f$.

\begin{table}[!t]
  \centering \small
  \caption{\textbf{Val-loss-optimal depth per budget (appendix detail).}
  $s$ is the induced embedder share; ranking metrics are at the same
  checkpoint.  Val-optimal $L_{\text{ctx}}$ varies; $s$ stays in the
  width-sweep band.}
  \label{tab:phase1d_summary}
  \begin{tabular}{lccccc}
    \toprule
    Budget & $L_{\text{ctx}}^{\star}$ & val\_loss & \texttt{R@10} & \texttt{NDCG@10} & $s$ (\%) \\
    \midrule
    $10^{15}$ & 12 & 3.950 & 0.465 & 0.329 & 0.5 \\
    $10^{16}$ &  6 & 3.439 & 0.568 & 0.397 & 3.7 \\
    $10^{17}$ &  4 & 3.164 & 0.619 & 0.432 & 6.1 \\
    $10^{18}$ & 32 & 2.782 & 0.644 & 0.453 & 0.9 \\
    \bottomrule
  \end{tabular}
\end{table}

\begin{table}[!t]
  \centering \small
  \caption{\textbf{Parabolic re-reading of the Phase~1D val\_loss
  grid.}  For each budget we fit
  $\mathrm{val\_loss}\!=\!a(\log_{10}s)^{2}\!+\!b\log_{10}s\!+\!c$ on
  the depth-sum-diagonal cells and report the analytic minimum
  $s^{\star}$, the val\_loss there, and the projected val\_loss
  penalty at $s\!=\!2\%$.  ``flat'' marks budgets where the swept range
  is too flat ($\leq\!0.21$~nats end-to-end at $C\!=\!10^{15}$) for the
  parabola to resolve a sharp interior minimum.  Per-cell penalties
  versus the parabolic minimum are in
  \texttt{parabolic\_depth\_fit\_cells.csv} alongside the table CSV.}
  \label{tab:phase1d_parabola}
  \begin{tabular}{lcccccc}
    \toprule
    Budget & disc.\ $L_{\text{ctx}}^{\star}$ & disc.\ $s$ (\%)
           & disc.\ val\_loss
           & parab.\ $s^{\star}$ (\%) & parab.\ val\_loss
           & val\_gap at $s\!=\!2\%$ \\
    \midrule
    $10^{15}$ & 12 & 0.5 & 3.950 & flat       & flat   & flat \\
    $10^{16}$ &  6 & 3.7 & 3.439 & 1.82       & 3.446  & $\!+0.000$ \\
    $10^{17}$ &  4 & 6.1 & 3.164 & 1.80       & 3.154  & $\!+0.000$ \\
    $10^{18}$ & 32 & 0.9 & 2.782 & 0.80       & 2.786  & $\!+0.009$ \\
    \bottomrule
  \end{tabular}
\end{table}

\begin{table}[!t]
  \centering
  \small
  \caption{\textbf{Held-out \texttt{val\_loss} per (budget, contextualizer
  depth).}  ``--'' = outside swept grid; bold = best at $C\!=\!10^{18}$
  among depth-sum-$34$ cells ($^{\dagger}$/$^{\ddagger}$ = deeper
  off-diagonal extensions).}
  \label{tab:phase1d}
  \begin{tabular}{lcccc}
    \toprule
    $L_{\text{ctx}}$ & $C\!=\!10^{15}$ & $10^{16}$ & $10^{17}$ & $10^{18}$ \\
    \midrule
    1  & 4.156          & 3.556          & 3.304          & --             \\
    2  & 4.063          & 3.495          & 3.206          & --             \\
    4  & 4.064          & 3.472          & \textbf{3.164} & --             \\
    6  & 4.028          & \textbf{3.439} & 3.178          & 2.948          \\
    8  & 4.062          & 3.462          & 3.165          & 2.912          \\
    10 & 4.027          & 3.458          & 3.175          & --             \\
    11 & 3.987          & 3.457          & 3.167          & --             \\
    12 & \textbf{3.950} & 3.476          & 3.194          & 2.886          \\
    16 & --             & --             & 3.239          & 2.858          \\
    24 & --             & --             & --             & 2.823          \\
    28 & --             & --             & --             & 2.809          \\
    32 & --             & --             & --             & \textbf{2.782} \\
    40 & --             & --             & --             & 2.805$^{\dagger}$ \\
    56 & --             & --             & --             & 2.847$^{\ddagger}$ \\
    \bottomrule
  \end{tabular}
\end{table}

\paragraph{Held-out evaluation at $C\!=\!10^{18}$.}
Table~\ref{tab:phase1d_eval} reports validation metrics on the
depth-sum-$34$ diagonal.  $L\!=\!32$ wins every metric; deeper
off-diagonal extensions ($L\!=\!40,56$) do not improve held-out quality.

\begin{table}[!t]
  \centering
  \small
  \caption{\textbf{Held-out evaluation at $C\!=\!10^{18}$ (depth-sum-$34$
  diagonal and off-diagonal extensions).}}
  \label{tab:phase1d_eval}
  \begin{tabular}{rrrrrrrr}
    \toprule
    $L_c$ & val\_loss & R@1 & R@10 & NDCG@10 & MRR@10 & cov@10 \\
    \midrule
    6  & 2.948          & 0.264          & 0.619          & 0.431          & 0.372          & 0.935          \\
    8  & 2.912          & 0.267          & 0.626          & 0.435          & 0.375          & 0.932          \\
    12 & 2.886          & 0.271          & 0.630          & 0.440          & 0.380          & 0.933          \\
    16 & 2.858          & 0.272          & 0.633          & 0.442          & 0.382          & 0.936          \\
    24 & 2.823          & 0.278          & 0.639          & 0.448          & 0.388          & 0.938          \\
    28 & 2.809          & 0.280          & 0.643          & 0.451          & 0.391          & 0.938          \\
    32 & \textbf{2.782} & \textbf{0.282} & \textbf{0.644} & \textbf{0.453} & \textbf{0.393} & \textbf{0.939} \\
    \midrule
    40$^{\dagger}$ & 2.805 & 0.281 & 0.642 & 0.451 & 0.391 & 0.937 \\
    56$^{\ddagger}$ & 2.847 & 0.276 & 0.635 & 0.445 & 0.385 & 0.935 \\
    \bottomrule
  \end{tabular}
\end{table}

% =====================================================================
\appendixsection{Phase 2: Per-Metric Trajectories}
\label{app:phase2}

\begin{figure}[!t]
  \centering
  \includegraphics[width=0.92\linewidth]{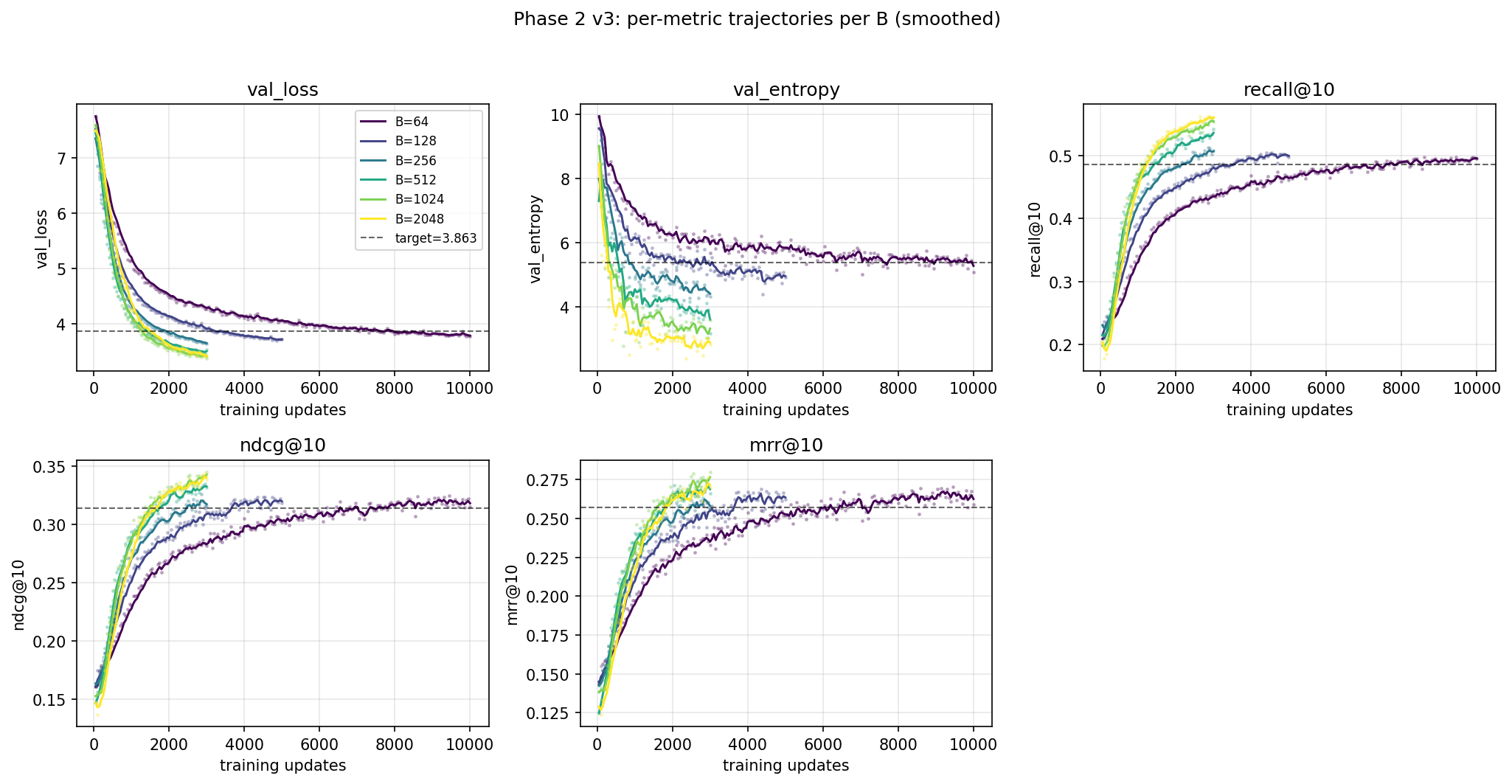}
  \caption{\textbf{Per-metric trajectories vs.\ batch size.}
  One curve per $B$; dashed line: iso-target $T_m$.}
  \label{fig:phase2_traj}
\end{figure}

Figure~\ref{fig:phase2_traj} shows the EWMA-smoothed metric trajectories
vs.\ optimizer updates for each $B\!\in\!\{64,128,256,512,1024,2048\}$ at
the Phase~1 winner ($C\!=\!10^{17}$, $s\!\approx\!2\%$;
\S\ref{sec:phase2}).  Dashed horizontal lines are the per-metric
iso-targets $T_m$; the first crossing of each curve defines
$S_m(B)$ in Tab.~\ref{tab:phase2}.  The Kaplan fits of those
updates-to-target points are Fig.~\ref{fig:phase2} in the main text.

% =====================================================================
\appendixsection{Phase 3 Train-Surrogate Allocation}
\label{app:phase3_train}

The main-text Phase~3 fits use \texttt{val\_loss} as the primary objective
(\S\ref{sec:phase3}, Table~\ref{tab:phase3_alloc}).  For completeness we
record the parallel tail-$100$ train-surrogate analysis on the same
architecture grid, the classic Chinchilla diagnostic.  Train and val
parabolic minima need not coincide: they diverge notably at $C\!=\!10^{18}$
($39.5$\,M train vs.\ $19.4$\,M val) but land near \texttt{h1152\_L16} on
both surrogates at $C\!=\!10^{19}$.  Parabolic smoothing also moves the
train-surrogate per-budget winner relative to its own discrete-cell choice
in Table~\ref{tab:phase3_alloc_train}: $0.92$\,M$\!\to\!706$\,k at
$10^{15}$, $31.9$\,M$\!\to\!39.5$\,M at $10^{18}$, $251$\,M$\!\to\!203$\,M
at $10^{19}$.

\begin{table}[!t]
  \centering
  \small
  \caption{\textbf{Train-surrogate Chinchilla allocation.}  $\Nstar$ and
  $\Dstar$ minimize the tail-$100$ training loss; $L^{\star}$ is the fitted
  irreducible-loss intercept trajectory (Eq.~\ref{eq:phase3_lstar_train}).}
  \label{tab:phase3_alloc_train}
  \begin{tabular}{lrrrr}
    \toprule
    Budget & $\Nstar$ & $\Dstar$ & $L^{\star}$ & $D/N$ \\
    \midrule
    $10^{15}$ & 0.92\,M & 62\,M     & 3.571 & 68  \\
    $10^{16}$ & 1.95\,M & 251\,M    & 3.087 & 128 \\
    $10^{17}$ & 10.9\,M & 537\,M    & 2.911 & 49  \\
    $10^{18}$ & 31.9\,M & 1.80\,B   & 2.809 & 56  \\
    $10^{19}$ & 251\,M  & 3.65\,B   & 2.585 & 15  \\
    \bottomrule
  \end{tabular}
\end{table}

Fitting $\Nstar(C)\!=\!aC^{b}$ in log-space on the five parabolic
train-surrogate minima gives
\begin{align}
  \Nstar_{\text{train}}(C) &\;=\; 4.06\!\times\!10^{-4}\, C^{0.612 \pm 0.024},
    \label{eq:phase3_nstar_train} \\
  \Dstar_{\text{train}}(C) &\;=\; 4.11\!\times\!10^{2}\, C^{0.388 \pm 0.024},
    \label{eq:phase3_dstar_train} \\
  L^{\star}_{\text{train}}(C) &\;=\; 2.485 + 1.50\!\times\!10^{3}\,C^{-0.210},
    \label{eq:phase3_lstar_train}
\end{align}
with $b\!+\!b'\!=\!1$ exact by Approach~2 construction.  For comparison,
the discrete-winner refit on the five rows of
Table~\ref{tab:phase3_alloc_train} gives slightly steeper exponents
($b_N\!=\!0.609\!\pm\!0.052$, $b_D\!=\!0.439\!\pm\!0.028$, summing to
$1.048$), the same $\sim\!5\%$ inflation that shows up on the val
surrogate in \S\ref{sec:phase3}.  The parametric three-term fit
\begin{equation}
  L_{\text{train}}(N,D) \;=\; 2.570 + \frac{2.46\!\times\!10^{3}}{N^{0.696}}
                    + \frac{8.08\!\times\!10^{2}}{D^{0.384}}
  \label{eq:phase3_Lpara_train}
\end{equation}
achieves RMSE $\!=\!0.084$ nats across the $48$ merged training points.

% =====================================================================
\appendixsection{Phase 4 Metric-Stratified Sampling Efficiency}
\label{app:phase4}

\begin{figure}[!t]
  \centering
  \includegraphics[width=0.62\linewidth]{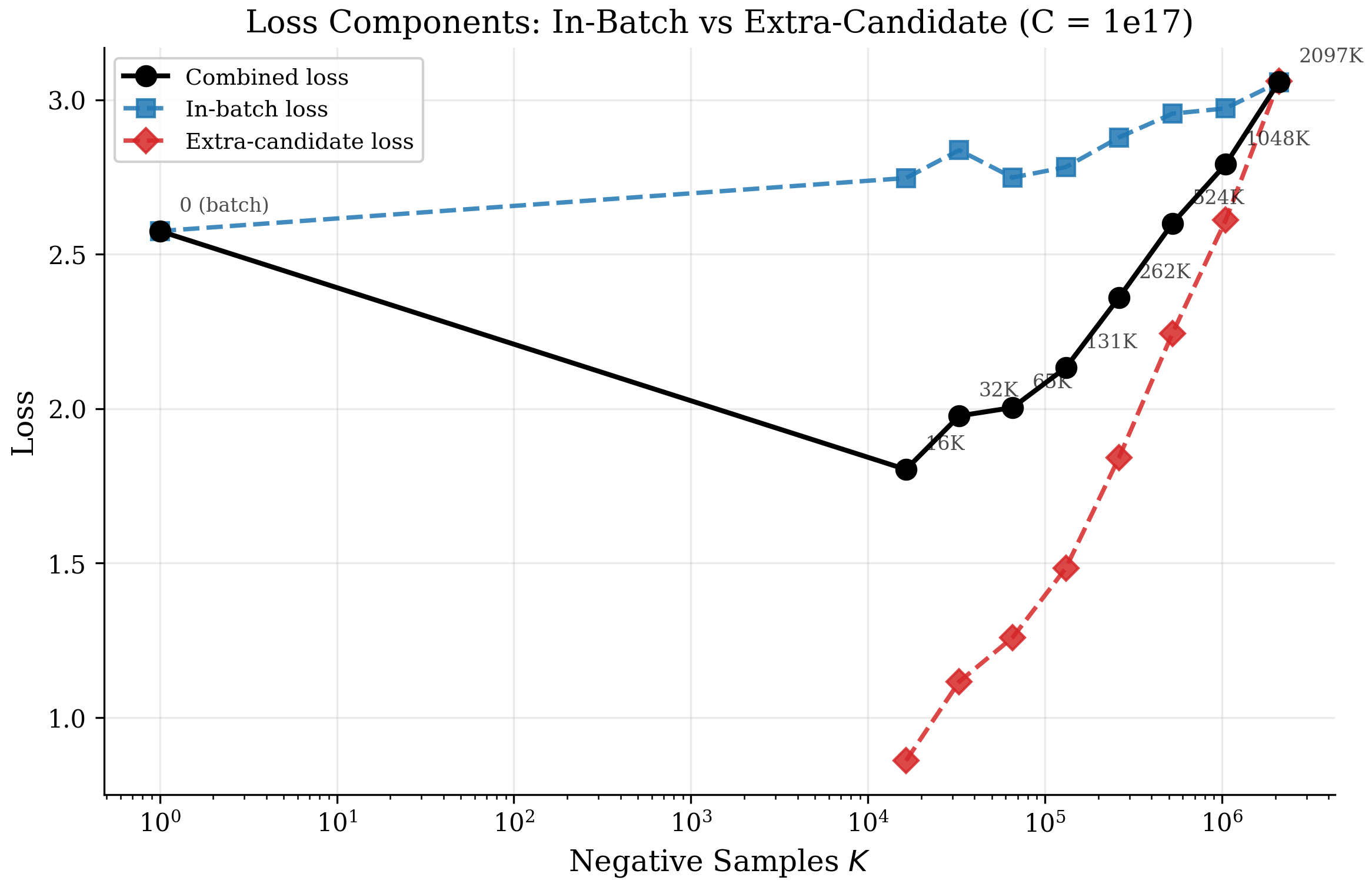}
  \caption{\textbf{Training-loss channels at $C\!=\!10^{17}$ (not
  full-catalogue evaluation).}  Blue dashed: in-batch CE; red dashed: extra-negative
  CE (only for $K\!>\!0$); black: their average, which SGD minimizes.
  The extra channel rises with $K$ by construction; the in-batch channel
  moves with the checkpoint trained at each $K$.}
  \label{fig:phase4_decomp}
\end{figure}

\paragraph{Training-loss decomposition (not comparable across $K$ as
quality).}  Figure~\ref{fig:phase4_decomp} plots the two channels the
Stage~2 trainer logs at $C\!=\!10^{17}$: in-batch CE (fixed
$|\mathcal{B}_{\text{batch}}|\!\approx\!16{,}384$ candidates) and extra CE
over the $K$ sampled catalog negatives, averaged for optimization.  These
curves are \emph{not} full-catalogue evaluation: the extra term is defined over a
candidate set that grows with $K$, so the combined training loss is not an
apples-to-apples quality metric across the sweep.  We include the panel
only to show why the implemented objective has two opposing pieces; all
$K^{\star}$ numbers in \S\ref{sec:phase4} come from full-catalogue evaluation.

\begin{figure}[!t]
  \centering
  \includegraphics[width=0.62\linewidth]{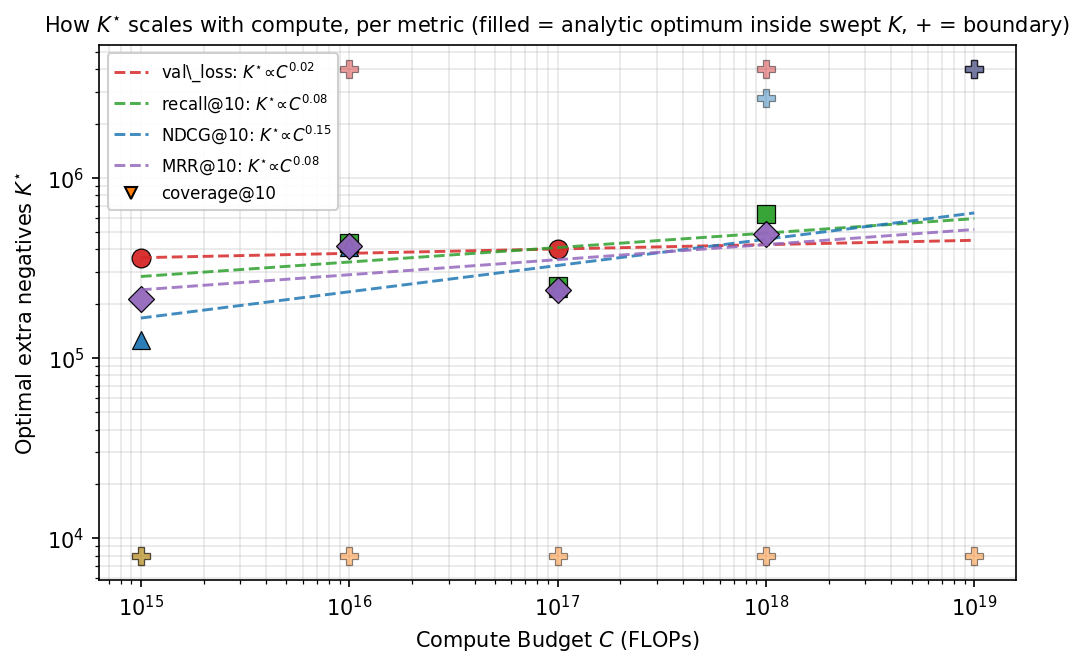}
  \caption{\textbf{Analytic $\Kstar(C)$ from \eqref{eq:LK}, per metric
  (interior-cell summary of Figure~\ref{fig:phase4_metrics}).}  Filled
  markers: interior analytic optimum from the per-(metric,$C$)
  parabola; faded $+$: boundary cases ($\Kstar$ extrapolated outside
  the swept range), excluded from the dashed power-law fit.  The fit
  is on $n\!=\!2$--$4$ interior cells per metric ($13$ of $25$ cells
  are boundary; see Table~\ref{tab:phase4}), so the dashed slopes
  $\Kstar\!\propto\!C^{0.09}\!\!-\!\!C^{0.15}$ summarize the interior
  cluster but should not be read as tightly-fit scaling exponents.
  The primary takeaway is the band, with ranking $\Kstar$ falling in
  $[125\mathrm{k},\,870\mathrm{k}]$ across $C\!\in\![10^{16},10^{18}]$, not the
  slope.}
  \label{fig:phase4_kstar_vs_C}
\end{figure}

Figure~\ref{fig:phase4_kstar_vs_C} summarizes the interior analytic
optima of Figure~\ref{fig:phase4_metrics} on a single $\Kstar$-vs-$C$
panel, and is the figure the practical recipe band of \S\ref{sec:phase4}
is read off of.  Three caveats on reading it.  \emph{First}, $13$ of
$25$ metric$\times$budget cells are boundary (Table~\ref{tab:phase4}),
so the dashed power-law slopes are fit on only $n\!=\!2$--$4$ interior
cells per metric and should be read as a within-band summary rather
than as tightly-identified scaling exponents.  This is why
finding~(ii) in \S\ref{sec:phase4} reports the band
$\Kstar\!\in\![125\mathrm{k},\,870\mathrm{k}]$ alongside the slope.
\emph{Second}, the bias exponent $\beta$ of \eqref{eq:LK} partitions
the metrics into three behaviors that are already visible in the
parabolas of Figure~\ref{fig:phase4_metrics}: bias-dominated
(\texttt{val\_loss}, \texttt{val\_entropy}; $\beta\!\sim\!1.2$ and
$\sim\!0.2$--$0.6$, still falling at $K\!=\!2$M), saturating ranking
metrics (\texttt{recall@10}, \texttt{NDCG@10}, \texttt{MRR@10};
$\beta\!\sim\!0.4$--$0.6$, interior peak at every budget
$\geq\!10^{16}$), and popularity-collapse diversity
(\texttt{coverage@10}, monotonically decreasing in $K$ at every $C$;
the fit returns $\beta\!>\!0$ but with the bias term acting as a
positive penalty, so we record \texttt{coverage@10} as a boundary case
on the larger-better side).  \emph{Third}, every metric at
$C\!=\!10^{19}$ is boundary (Table~\ref{tab:phase4}), because all
panels at that budget are still trending at $K\!=\!2$M; finding (iii)
of \S\ref{sec:phase4} attributes this to the binding constraint on
$K$ flipping from compute to memory.

\begin{figure}[!t]
  \centering
  \includegraphics[width=0.92\linewidth]{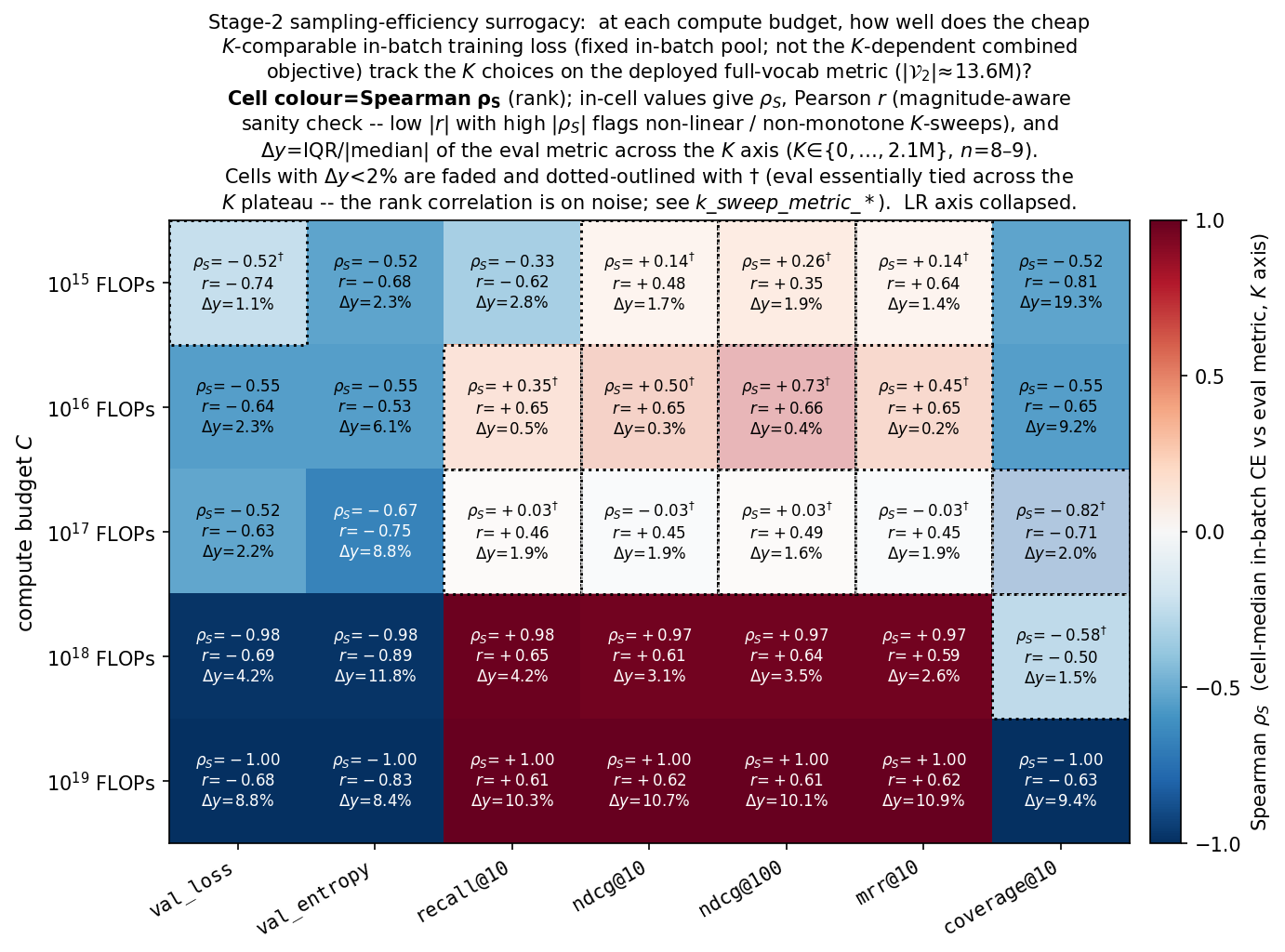}
  \caption{\textbf{Negative-sampling efficiency surrogacy.}  Cell color is
  Spearman $\rho_S$ between the $K$-comparable \emph{in-batch} training
  loss and the eval metric across the $K$ axis (the optimized combined
  objective folds in a $K$-dependent extra-negative channel and is not
  comparable across $K$; cf.\ Fig.~\ref{fig:phase4_decomp}).  Cells where
  the underlying $K$-sweep is essentially flat are faded.  At
  $C\!=\!10^{18}$--$10^{19}$ the in-batch loss is near-perfectly
  \emph{anti}-correlated with every deployed metric
  ($|\rho_S|\!\approx\!0.97$--$1.00$): the $K$ that minimizes it
  ($K\!=\!0$) is the worst $K$ for full-catalogue performance.  At smaller
  budgets the eval metrics are nearly flat across $K$ and the signal is
  weak.}
  \label{fig:p4_surrogacy}
\end{figure}

\paragraph{Surrogacy stratified by $(C,\text{metric})$.}
Figure~\ref{fig:p4_surrogacy} stratifies the Stage~2 surrogacy by
$(\text{compute budget, eval metric})$.  The training-time signal here is
the \emph{in-batch} cross-entropy---the $K$-comparable channel of the
Stage~2 objective.  We deliberately do \emph{not} use the optimized
combined loss: its extra-negative channel grows with $K$ by construction
(Fig.~\ref{fig:phase4_decomp}), so the combined loss carries a mechanical
$K$-trend and is not comparable across the sweep as a quality measure.
The high-compute rows are then unambiguous: at $C\!=\!10^{18}$ and
$10^{19}$ the in-batch loss is almost perfectly \emph{anti}-correlated
with every deployed metric---ranking $\rho_S\!\approx\!+0.97$ to $+1.00$,
and \texttt{val\_loss}, \texttt{val\_entropy}, \texttt{coverage@10}
$\rho_S\!\approx\!-0.98$ to $-1.00$---on a large underlying spread
($\Delta y\!\approx\!3$--$11\%$).  Both rows share one mechanism: the $K$
that minimizes in-batch loss is $K\!=\!0$ (with no extra negatives the
contextualizer over-fits the easy in-batch task), yet every full-catalogue
metric improves monotonically out to the largest sampled $K\!=\!2.1$M, so
the cheap in-batch signal points to exactly the wrong end of the $K$ axis.
This is in fact what we should expect from the form of the objective: the
optimized loss is the average of the batch-local and extra-candidate
cross-entropies, and as $K$ grows the extra-candidate term---scored against
an ever-larger negative pool---increasingly dominates the gradient.  The
optimizer therefore trades away in-batch fit, so the batch-local loss drifts
\emph{up} with $K$ even as the model gets better at the full-catalogue task
that the deployed metrics reward.
At $C\!\le\!10^{17}$ the in-batch loss instead carries little reliable
signal: the ranking metrics are nearly flat across $K$
($\Delta y\!\lesssim\!2\%$, faded cells) and the residual $\rho_S$ is weak
and mixed in sign.  In short, the cheap training-time loss is at best
uninformative and at worst actively misleading for choosing $K$.

% =====================================================================
\appendixsection{Maximal Update Parameterization}
\label{app:mup}

We swept four model sizes
(\textsc{tiny}/\textsc{small}/\textsc{medium}/\textsc{large}), spanning
$\sim\!10$\,M to $\sim\!500$\,M \emph{total} trainable parameters with the
embedder and contextualizer scaled together, and learning rates in
$[10^{-5},\,5\!\cdot\!10^{-2}]$ with two initialization strategies:
\textbf{Default} (our usual truncated-normal--style init, as in the rest of
this work) and
\textbf{MuP} (output and hidden weights LR-scaled by $1/\mathrm{fan\_in}$,
embeddings and biases unscaled).  We use $8$ MuP learning rates spanning
$\{10^{-4},\dots,5\!\cdot\!10^{-2}\}$ so that the MuP optimum is bracketed
from above at every size, i.e.\ verified as a local minimum rather than as
the right edge of the swept range.

\begin{table}[!t]
  \centering
  \small
  \caption{\textbf{MuP halves the LR drift but does not improve loss.}
  Training loss per (size, init) at the optimal LR; the MuP optima are
  verified local minima (bracketed from above by strictly-higher LRs that
  are strictly worse).  Our default initialization wins by $0.68$--$0.92$
  nats at every scale.}
  \label{tab:mup}
  \begin{tabular}{lcccc}
    \toprule
    Init & \textsc{tiny} & \textsc{small} & \textsc{medium} & \textsc{large} \\
    \midrule
    Default       & \textbf{4.626} & \textbf{4.210} & \textbf{3.594} & \textbf{2.904} \\
    MuP           & 5.547          & 5.103          & 4.516          & 3.584          \\
    \midrule
    Default$-$MuP & $-0.921$ & $-0.893$ & $-0.922$ & $-0.680$ \\
    \bottomrule
  \end{tabular}
\end{table}

\begin{figure}[!t]
  \centering
  \begin{subfigure}[t]{0.48\linewidth}
    \includegraphics[width=\linewidth]{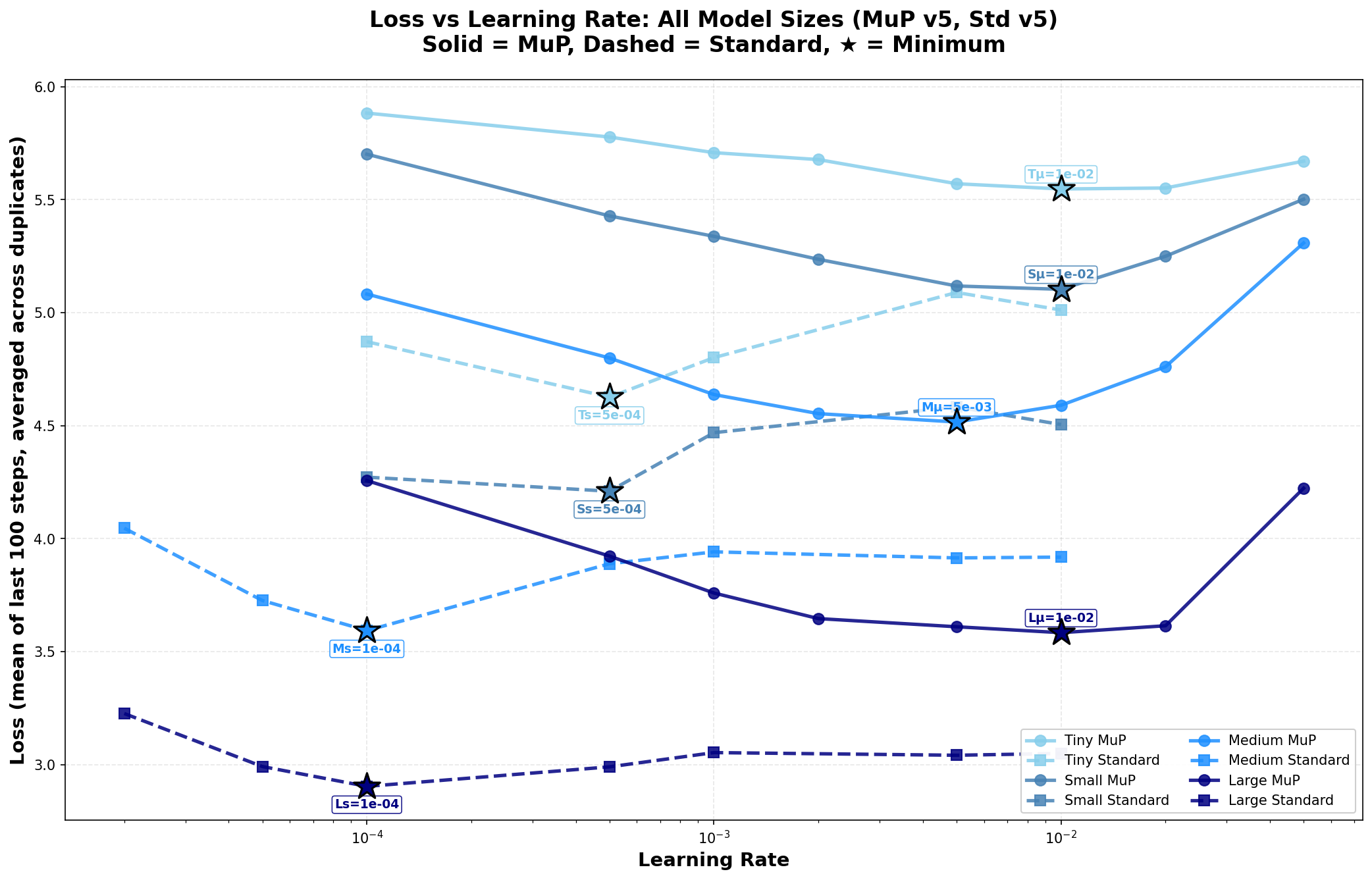}
    \caption{Loss vs.\ LR for every (size, init) cell.}
    \label{fig:mup_sweep}
  \end{subfigure}\hfill
  \begin{subfigure}[t]{0.48\linewidth}
    \includegraphics[width=\linewidth]{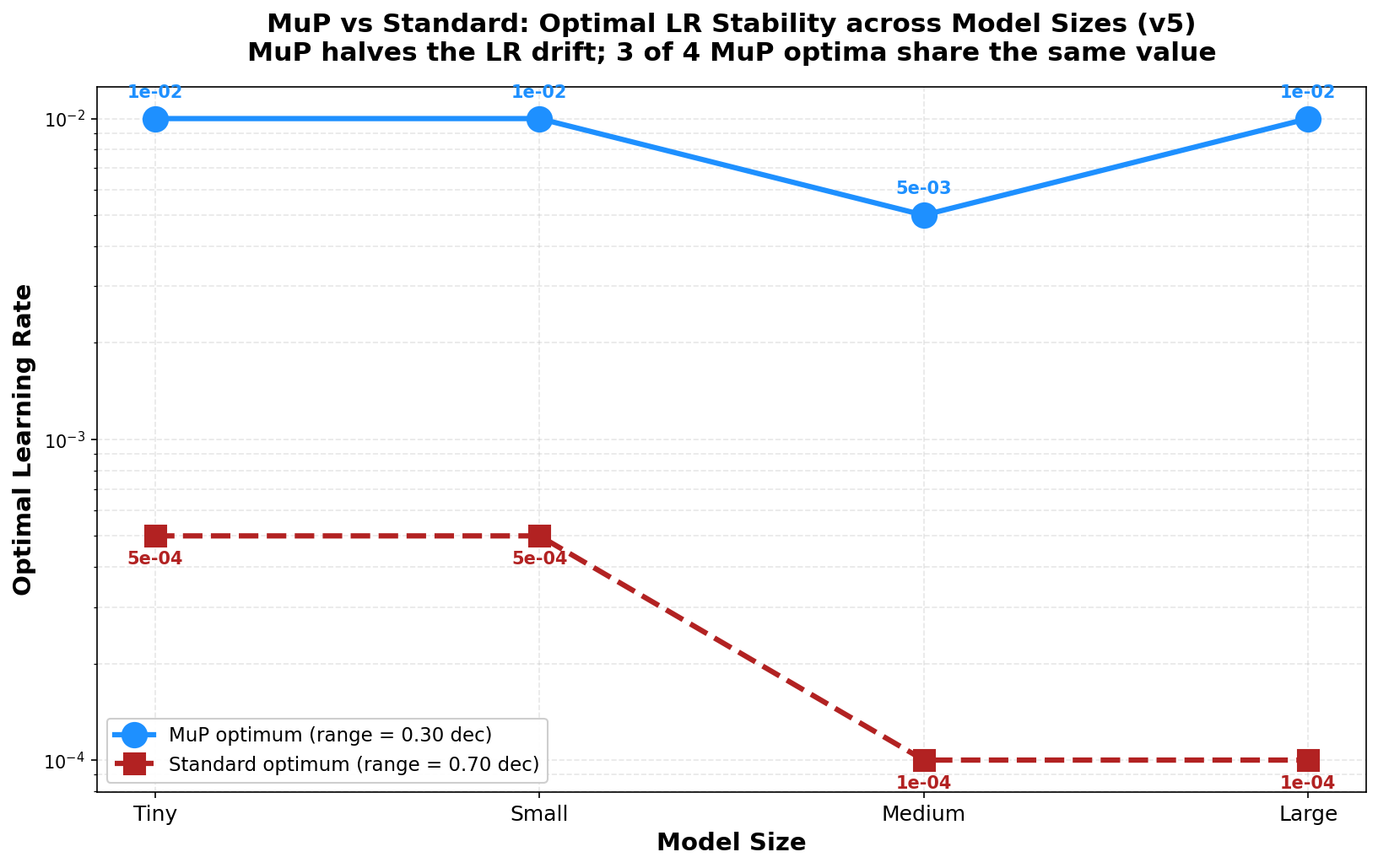}
    \caption{Optimal LR vs.\ model size.}
    \label{fig:mup_opt}
  \end{subfigure}
  \caption{\textbf{MuP halves the LR drift but does not improve loss.}
  Solid: MuP, dashed: Default.  MuP optima land at $10^{-2}$ at
  \textsc{tiny}, \textsc{small} and \textsc{large}, and at
  $5\!\cdot\!10^{-3}$ at \textsc{medium} (a $0.30$-decade band) versus
  $0.70$ decades for Default.  Default initialization sits below MuP at
  every model size.}
  \label{fig:mup}
\end{figure}

\textbf{Findings.}  MuP delivers most of its core promise: the optimum LR
is exactly $10^{-2}$ at three of the four sizes (the MuP-optimum span is
$0.30$ decades against $0.70$ for Default).  Every MuP optimum is a
verified local minimum: at every size, $2\!\cdot\!10^{-2}$ and
$5\!\cdot\!10^{-2}$ produce strictly worse loss
(Figure~\ref{fig:mup_sweep}).  But Default reaches a lower training loss at
every size by $0.68$--$0.92$ nats (Table~\ref{tab:mup}).  The same
pattern holds for every other MuP variant we tried, including a
per-layer MuP and a FLOP-budget-matched variant; we therefore retain our
default truncated-normal--style initialization (Table~\ref{tab:recipe})
and pay the modest cost of a per-phase LR sweep
instead.

% =====================================================================
\appendixsection{Cross-Metric Details}
\label{app:metrics}

\paragraph{How well does training loss track eval metrics?}
Within Stage~1 the sampled-softmax training loss correlates with the
headline eval metrics (excluding \texttt{MRR@10} and \texttt{coverage@10})
at $|\rho_S|\!\geq\!0.99$ (Table~\ref{tab:trainvseval}).
This is partly tautological: Stage~1 evaluates against a batch-local pool,
i.e.\ the same construction the optimizer sees, and at our budgets the
train/val gap is small.  Stage~2 (Phase~4) is the genuine
batch-local-vs-full-catalogue comparison; using the $K$-comparable in-batch
cross-entropy, pooled across the whole sweep it tracks the eval metrics only
moderately ($|\rho_S|\!\in\![0.51, 0.90]$).  This pooled figure is, however,
dominated by the compute axis (larger $C$ lowers in-batch loss and raises
every metric) and \emph{masks} the per-budget behavior along $K$: at fixed
high compute the in-batch loss \emph{anti}-tracks the deployed metrics almost
perfectly (Fig.~\ref{fig:p4_surrogacy}), so it is not a usable surrogate for
choosing $K$.

\begin{table}[!t]
  \centering
  \small
  \caption{\textbf{How well batch-local training loss tracks validation
  eval metrics.}  Stage~1 medians are tautologically high (same in-batch
  construction).  Stage~2 uses the $K$-comparable in-batch cross-entropy
  channel and pools all $(K,C)$ cells; this pooled value is dominated by
  the compute axis and should be read together with the per-budget
  Fig.~\ref{fig:p4_surrogacy}, which shows the in-batch loss
  \emph{anti}-tracking the deployed metrics at fixed high compute.}
  \label{tab:trainvseval}
  \begin{tabular}{lrr}
    \toprule
    Eval metric & Stage 1 median $\rho_S$ & Stage 2 $\rho_S$ ($n\!=\!44$) \\
    \midrule
    \texttt{val\_loss}      & $+0.995$ & $+0.68$ \\
    \texttt{val\_entropy}   & $+0.995$ & $+0.51$ \\
    \texttt{recall@10}      & $-0.994$ & $-0.65$ \\
    \texttt{NDCG@10}        & $-0.992$ & $-0.64$ \\
    \texttt{NDCG@100}       & $-0.994$ & $-0.64$ \\
    \texttt{MRR@10}         & $-0.83$  & $-0.64$ \\
    \texttt{coverage@10}    & $-0.97$  & $-0.90$ \\
    \bottomrule
  \end{tabular}
\end{table}

\begin{figure}[!t]
  \centering
  \includegraphics[width=0.92\linewidth]{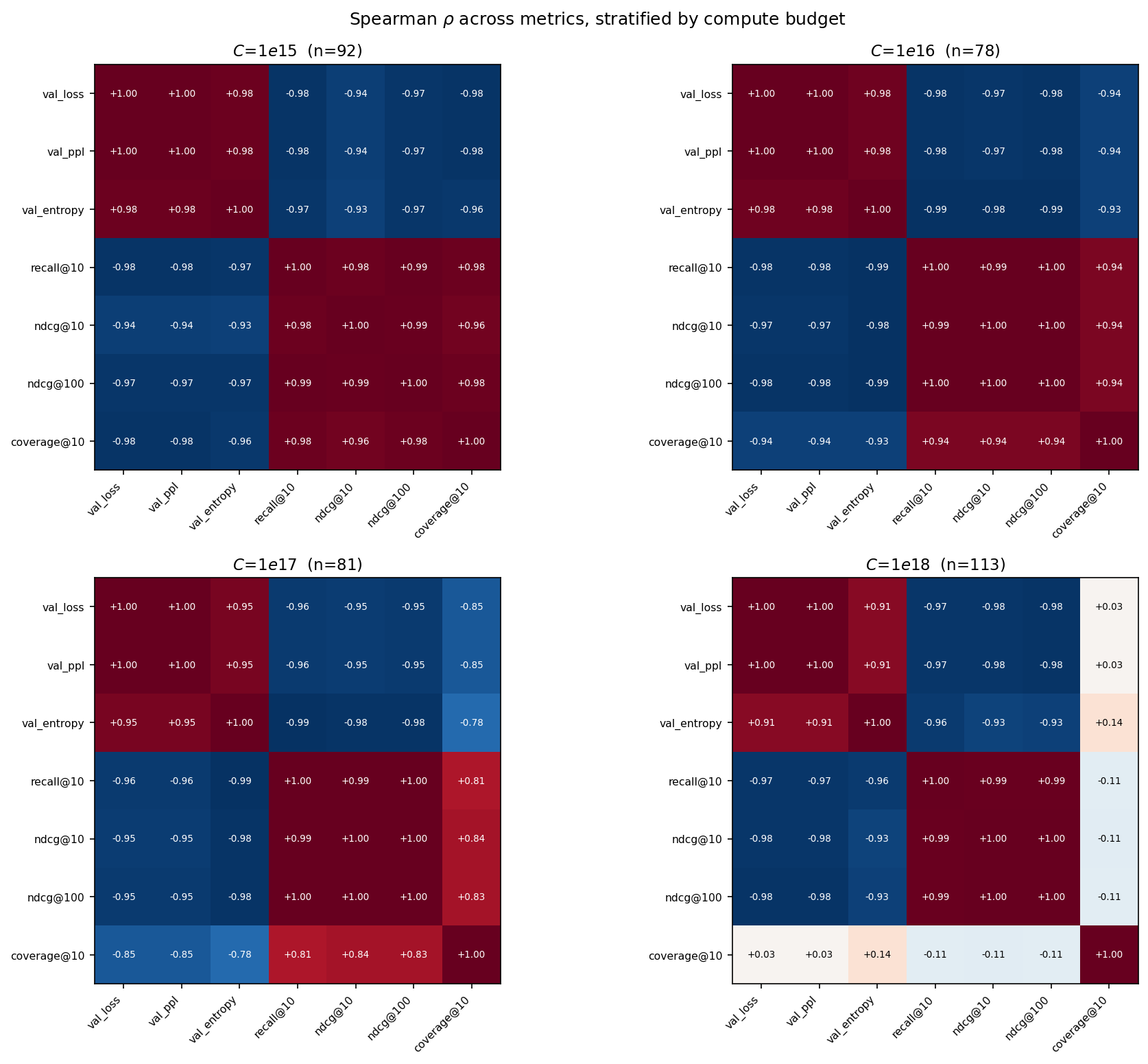}
  \caption{\textbf{Correlation matrices stratified by budget.}  The
  loss/perplexity/entropy/ranking block stays saturated at every $C$, but
  the coverage rows and columns visibly fade with scale.}
  \label{fig:corr_by_budget}
\end{figure}

\paragraph{Stratifying by compute budget.}
The loss--ranking link stays essentially flat at $\rho_S\!\approx\!-0.98$
across the full compute range, whereas the loss--coverage link weakens
monotonically and \emph{collapses} at $C\!=\!10^{18}$
(Table~\ref{tab:corr_by_budget}, Figure~\ref{fig:corr_by_budget}): once the
model is good enough that catalogue coverage saturates, which cells happen
to spread the head distribution out furthest is essentially independent of
which cells minimize loss.

\begin{table}[!t]
  \centering
  \small
  \caption{\textbf{Per-budget Spearman correlations} between headline
  metrics (Stage~1 pool).}
  \label{tab:corr_by_budget}
  \begin{tabular}{lrrrrrr}
    \toprule
    Budget & $n$ & loss--R@10 & loss--N@10 & loss--N@100 & loss--C@10 & loss--entropy \\
    \midrule
    $10^{15}$ &  99 & $-0.98$ & $-0.93$ & $-0.97$ & $-0.97$ & $+0.99$ \\
    $10^{16}$ &  85 & $-0.99$ & $-0.98$ & $-0.99$ & $-0.92$ & $+0.99$ \\
    $10^{17}$ &  86 & $-0.99$ & $-0.98$ & $-0.98$ & $-0.79$ & $+0.99$ \\
    $10^{18}$ &  86 & $-0.97$ & $-0.98$ & $-0.98$ & $-0.22$ & $+0.92$ \\
    \bottomrule
  \end{tabular}
\end{table}

\paragraph{Stage~2 per-budget correlations.}
Table~\ref{tab:corr_by_budget_stage2} is the Phase~4 counterpart to
Table~\ref{tab:corr_by_budget}: per-budget Spearman correlations on the
best-LR-per-cell $K$-sweep (8--9 $K$-cells per budget, full-catalogue evaluation).
Three contrasts with the Stage~1 table are worth flagging.  \emph{(i)}~The
loss--ranking link is \emph{not} locked: \texttt{val\_loss}--\texttt{recall@10}
swings from $+0.93$ at $C\!=\!10^{15}$ (loss and recall move in the
\emph{same} direction across $K$, so pushing $K$ up reduces both) through
near-zero at $C\!\in\!\{10^{16},10^{17}\}$ to $-1.00$ at $C\!=\!10^{18}$
(perfect alignment in the expected direction).  \emph{(ii)}~The
loss--coverage sign \emph{flips}: $-0.97$--$-0.22$ in Stage~1 (bigger
architectures get both lower loss and higher coverage) versus
$+0.60$--$+1.00$ in Stage~2 (larger $K$ sharpens predictions, which
lowers loss but \emph{narrows} coverage).  \emph{(iii)}~The Stage~2 $n$
is small (8--9 per budget) so the exact magnitudes are noisy, but the
qualitative pattern is robust: the $K$-axis decouples the four loss-like
and ranking-like metrics in a budget-dependent way.
The mechanism is the same one analysed mathematically in
\S\ref{sec:metrics} (importance-reweighting): the in-batch
contrastive loss does not learn absolute popularity, so the loss
minimizer and the full-catalogue ranking maximizer need not coincide,
and the gap shrinks as more sampled negatives push $q$ toward uniform.

\begin{table}[!t]
  \centering
  \small
  \caption{\textbf{Per-budget Spearman correlations (Stage~2, Phase~4
  $K$-sweep).}  Best-LR-per-cell; $n$ is the number of $K$-cells available
  at that budget.  \emph{Both} the ``loss'' here and every ranking, coverage
  and entropy metric are \textbf{full-catalogue evaluation} quantities,
  computed against the full $\sim\!13.6$M-item Stage-2 catalogue: ``loss''
  is the full-catalogue \texttt{val\_loss}, not the training-time loss.  These
  are therefore eval-metric vs.\ eval-metric correlations across $K$, distinct
  from the train-loss surrogacy of Fig.~\ref{fig:p4_surrogacy}.  Compare with
  the Stage~1 table (\ref{tab:corr_by_budget}): loss--ranking is no longer
  locked at $\rho_S\!\approx\!-0.98$ (it swings from $+0.93$ to $-1.00$ across
  $C$), and the loss--coverage sign flips relative to Stage~1.}
  \label{tab:corr_by_budget_stage2}
  \begin{tabular}{lrrrrrrr}
    \toprule
    Budget & $n$ & loss--R@10 & loss--N@10 & loss--N@100 & loss--MRR@10
    & loss--C@10 & loss--entropy \\
    \midrule
    $10^{15}$ & 8 & $+0.93$ & $+0.29$ & $+0.24$ & $+0.29$ & $+1.00$ & $+1.00$ \\
    $10^{16}$ & 9 & $+0.10$ & $-0.03$ & $-0.95$ & $-0.08$ & $+1.00$ & $+1.00$ \\
    $10^{17}$ & 9 & $+0.15$ & $+0.22$ & $+0.15$ & $+0.22$ & $+0.85$ & $+0.90$ \\
    $10^{18}$ & 9 & $-1.00$ & $-0.98$ & $-0.98$ & $-0.98$ & $+0.60$ & $+1.00$ \\
    $10^{19}$ & 9 & $-1.00$ & $-1.00$ & $-1.00$ & $-1.00$ & $+1.00$ & $+1.00$ \\
    \bottomrule
  \end{tabular}
\end{table}

\begin{figure}[!t]
  \centering
  \includegraphics[width=0.7\linewidth]{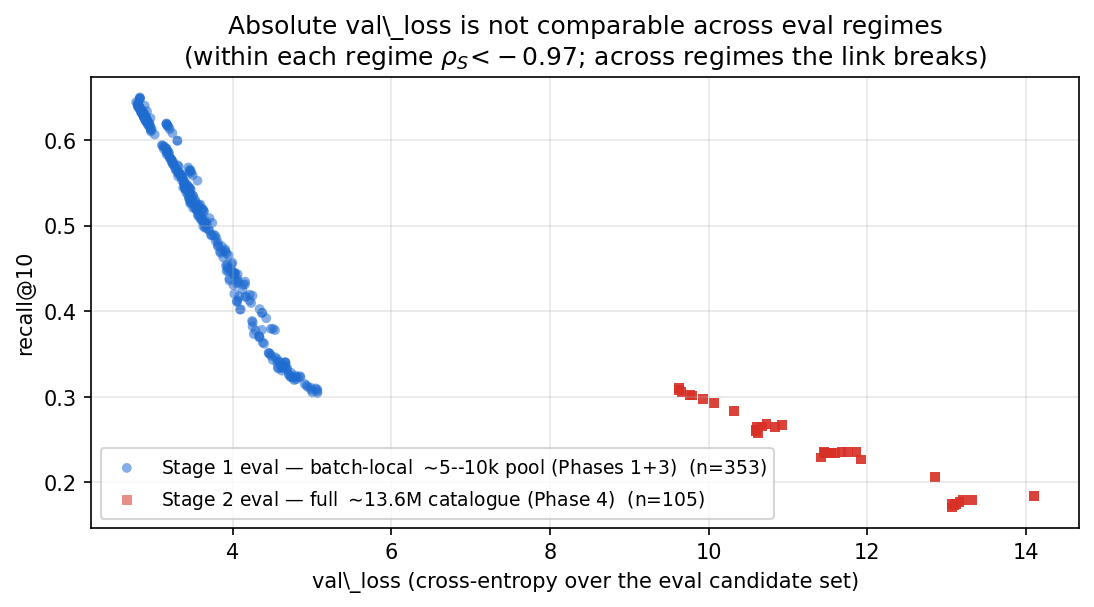}
  \caption{\textbf{Same recall@10, very different absolute val\_losses
  depending on stage.}  Blue: Stage~1 (batch-local pool); red: Stage~2
  (full catalogue).  Within either cloud $|\rho_S|\!\geq\!0.97$; across
  clouds the link breaks because the partition function is computed over a
  $\sim\!1500\!\times$ larger set.}
  \label{fig:regime_contrast}
\end{figure}

\paragraph{Why absolute loss is incomparable across stages.}
Stage~1 evaluates on a batch-local pool of $\sim\!5$--$10$k items, Stage~2
on the full deployed catalogue.  The cross-entropy of a uniform predictor
differs by several nats just from this candidate-pool denominator, which
by itself accounts for most of the observed $\sim\!7$-nat gap in
Figure~\ref{fig:regime_contrast}.

% =====================================================================
\appendixsection{Context-Length Scoring Robustness (full slice matrices)}
\label{app:context_length}

This appendix backs axis~(e) of \S\ref{sec:metrics}.  The eval
pipeline stratifies every batch by \emph{scoring} history position
and logs every headline metric at
$cl\!\in\!\{3,5,10,20,50,100\}$ events of preceding context alongside
the aggregate \texttt{val/all}.  The question is whether the cell
ranking (architectural cells for Stage~1, $K$-cells for Stage~2)
depends on which slice we score on.  All models are trained at the
full $\Lseq\!=\!256$ context length, so this measures the
scoring-position half of the context-length sensitivity, not the
training-context half (\S\ref{sec:discussion}).

\begin{figure}[!t]
  \centering
  \includegraphics[width=0.99\linewidth]{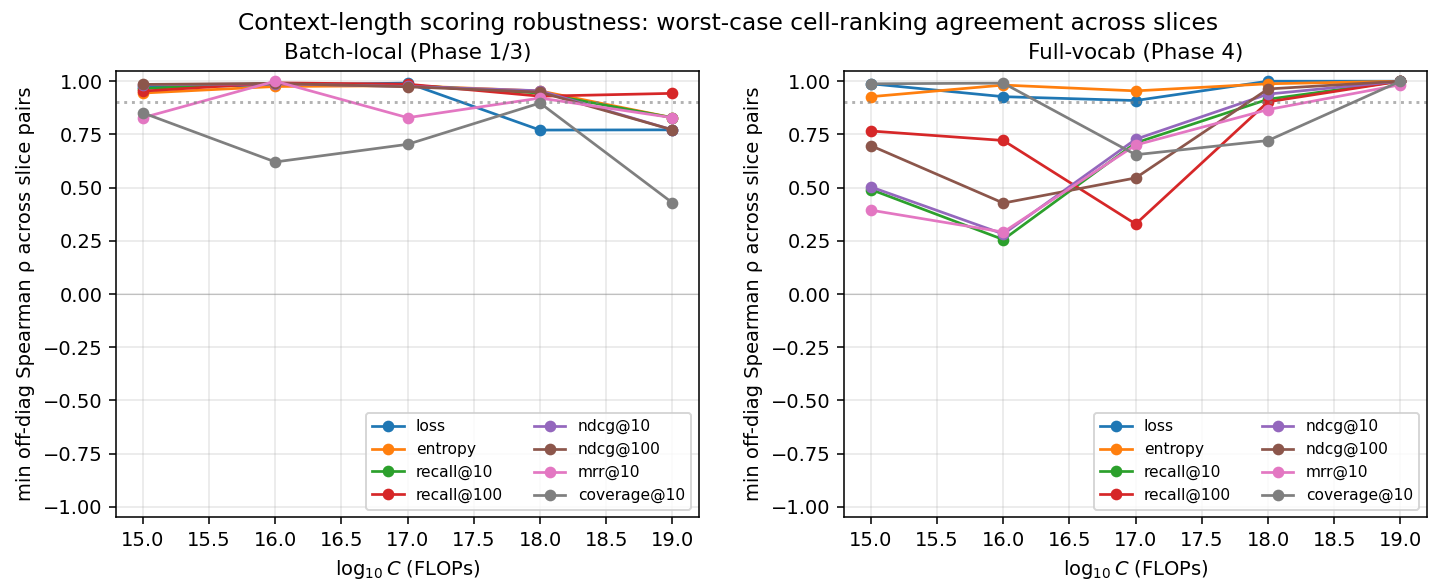}
  \caption{\textbf{Context-length scoring robustness: worst-case
  cell-ranking agreement across history-position slices, per regime.}
  Per (regime, metric, budget) we compute the Spearman~$\rho$ between
  every pair of scoring slices $\{$\texttt{ctx\_3}, \texttt{ctx\_5},
  \texttt{ctx\_10}, \texttt{ctx\_20}, \texttt{ctx\_50},
  \texttt{ctx\_100}, \texttt{val/all}$\}$ and plot the minimum
  off-diagonal pair.  \emph{Left:} batch-local Stage~1 evals
  (architecture sweep): every headline ranking metric stays
  $\rho_{\min}\!\geq\!0.93$ at $C\!\leq\!10^{18}$.  \emph{Right:}
  full-catalogue Stage~2 evals ($K$-sweep): ranking metrics
  $\rho_{\min}\!\in\![0.26,0.73]$ at $C\!\leq\!10^{17}$, realigning
  to $\rho_{\min}\!\geq\!0.87$ at $C\!\geq\!10^{18}$ on the same
  compute threshold at which (c)'s loss-vs-ranking correlation flips
  to perfectly aligned.}
  \label{fig:ctxlen_minrho}
\end{figure}

\begin{figure}[!t]
  \centering
  \includegraphics[width=0.98\linewidth]{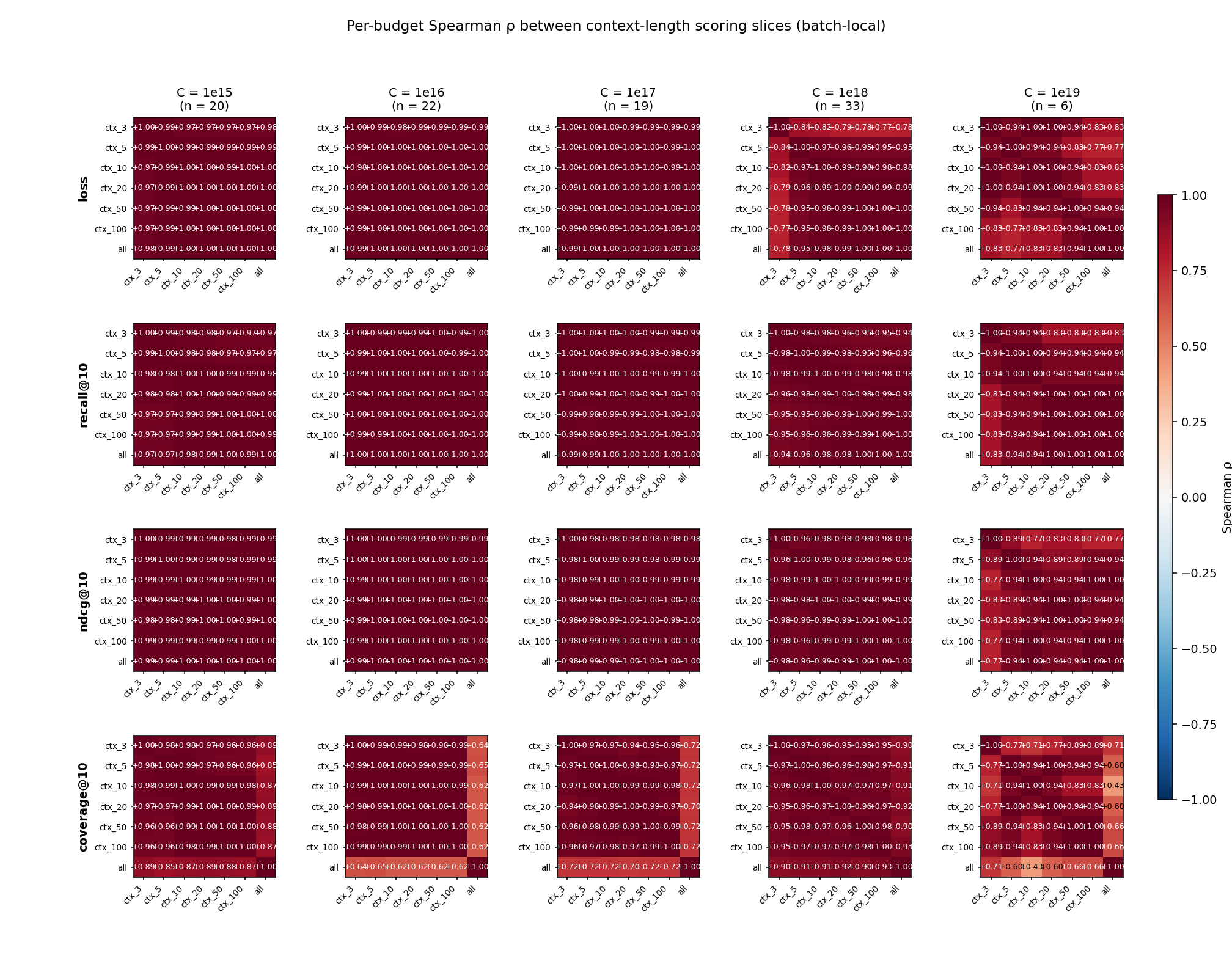}
  \caption{\textbf{Per-budget Spearman~$\rho$ between
  context-length scoring slices, batch-local evals
  (Phase~1W+1D+3 architecture sweep).}  Each row is a metric, each
  column a budget; the per-budget cell-count $n$ is in the column
  header.  Every cell is the Spearman rank correlation between two
  scoring-slice columns across the architectural cells at that
  budget.}
  \label{fig:ctxlen_heatmap_bl}
\end{figure}

\begin{figure}[!t]
  \centering
  \includegraphics[width=0.98\linewidth]{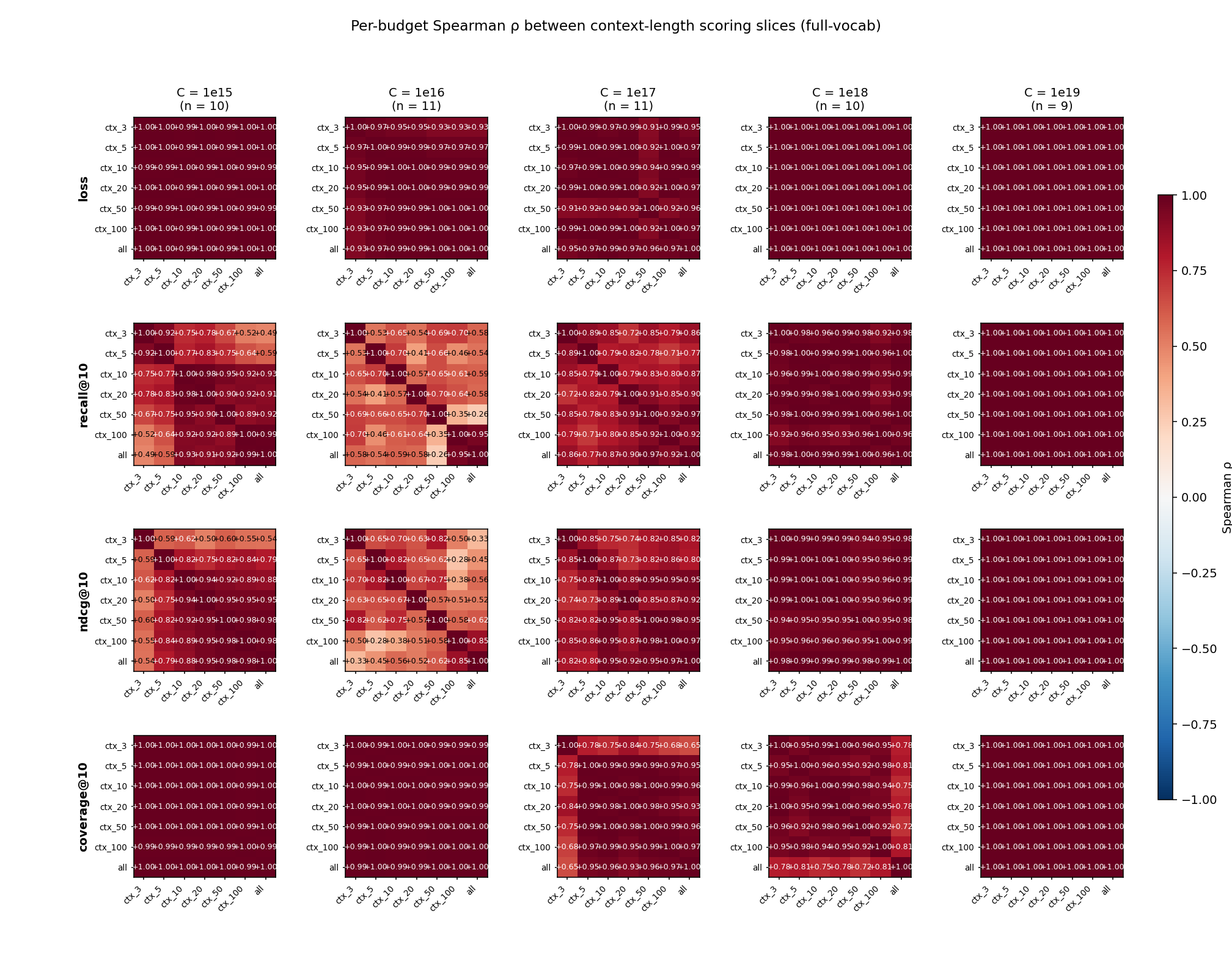}
  \caption{\textbf{Per-budget Spearman~$\rho$ between
  context-length scoring slices, full-catalogue evals (Phase~4
  $K$-sweep).}  Same axes as Figure~\ref{fig:ctxlen_heatmap_bl}.
  The \texttt{recall@10}, \texttt{NDCG@10} and \texttt{MRR@10} panels
  at $C\!\in\!\{10^{15},10^{16},10^{17}\}$ are the cells driving the
  $\rho_{\min}\!\in\![0.26,0.73]$ summary of
  Figure~\ref{fig:ctxlen_minrho}.}
  \label{fig:ctxlen_heatmap_fv}
\end{figure}

For each (regime, metric, budget) we compute the Spearman~$\rho$
between every pair of scoring-slice columns across the cells at that
budget.  Figure~\ref{fig:ctxlen_minrho} reports the worst-case
off-diagonal $\rho$ per panel;
Figures~\ref{fig:ctxlen_heatmap_bl}--\ref{fig:ctxlen_heatmap_fv}
break the same data out as full 7$\times$7 heatmaps for each
(regime, metric, budget).

\textbf{Stage~1 (batch-local; Phase~1W+1D+3 architecture sweeps,
$n\!=\!19$--$33$ cells per budget).}  The headline ranking metrics
(\texttt{recall@10}, \texttt{NDCG@10}, \texttt{NDCG@100}) all hold
$\rho_{\min}\!\geq\!0.93$ at $C\!\leq\!10^{18}$;
\texttt{val\_loss} and \texttt{val\_entropy} sit at
$\rho_{\min}\!\geq\!0.94$ for $C\!\leq\!10^{17}$ and soften to
$\rho_{\min}\!\approx\!0.77$ at $C\!=\!10^{18}\text{--}10^{19}$
where the val-loss landscape itself is flatter
(\S\ref{sec:phase1d}); \texttt{MRR@10} dips occasionally to
$\approx\!0.83$ on small-$n$ cells, and \texttt{coverage@10}
decouples earlier as in Table~\ref{tab:corr_by_budget}.  Either way,
Phase-1/3 architectural winners transfer under shorter- or
longer-history scoring; the architectural argmax is essentially
independent of scoring position.

\textbf{Stage~2 (full-catalogue; Phase~4 $K$-sweep, $n\!=\!9$--$11$ cells
per budget).}  Ranking metrics show
$\rho_{\min}\!\in\![0.26,\,0.73]$ at $C\!\leq\!10^{17}$ and only
realign to $\rho_{\min}\!\geq\!0.87$ at $C\!\geq\!10^{18}$.  The
geometry visible in Figure~\ref{fig:ctxlen_heatmap_fv} is that the
short-context slices (\texttt{ctx\_3}, \texttt{ctx\_5}) and the
long-context tail (\texttt{ctx\_100}, \texttt{val/all}) only weakly
rank-agree; the middle slices ($cl\!\in\![10,50]$) sit in between.
This is consistent with the importance-weighting mechanism of
\S\ref{sec:metrics}: short-history queries are popularity-dominated
(little informative context to condition on) so the $K$ that
minimises their conditional weights the $1/q$ sampling bias
differently from a long-history query; once $K$ is large enough to
push $q$ toward uniform the two converge, which is exactly the
$C\!\geq\!10^{18}$ realignment.  The practical corollary is that
the $K^{\star}$ band in Table~\ref{tab:phase4} should be read at the
deployed serving context length: at $C\!=\!10^{15}\text{--}10^{17}$
a model picked on \texttt{recall@10} pooled over \texttt{val/all} is
not guaranteed to be the best model for a cold-start user.
Note the two stages move in opposite directions with compute:
Stage~2 slice agreement \emph{rises} with budget (the
$C\!\geq\!10^{18}$ realignment above), whereas Stage~1 stays high
throughout with no upward trend and its
\texttt{val\_loss}/\texttt{val\_entropy} agreement even softens
slightly at the top budgets.

\end{document}